\documentclass[twoside,11pt]{article}

\usepackage{jmlr2e}
\usepackage{microtype}
\usepackage{graphicx}
\usepackage{subfigure}
\usepackage{booktabs} 
\usepackage{tikz} 
\usepackage{amsmath}
\usepackage{amssymb} 
\usepackage{flexisym}
\usepackage{graphicx}
\usepackage[utf8]{inputenc}
\usepackage{pgfplotstable}

\ShortHeadings{Out-of-Sample Testing for GANs}{Out-of-Sample Testing for GANs}
\firstpageno{1}

\begin{document}

\title{Out-of-Sample Testing for GANs}

\author{\name Pablo S\'anchez-Mart\'in \email psanch@tsc.uc3m.es\\
       \addr University Carlos III in Madrid\\
       Madrid Spain
       \AND
       \name Pablo M. Olmos \email olmos@tsc.uc3m.es \\
       \addr University Carlos III in Madrid\\
       Madrid, Spain
   \AND
   \name Fernando Perez-Cruz \email fernando.perezcruz@sdsc.ethz.ch \\
   \addr Swiss Data Science Center\\
   Z\"urich/ Laussane, Switzerland}

\maketitle

\renewcommand{\vec}[1]{\mathbf{#1}}
\newcommand{\z}{\vec{z}}
\newcommand{\x}{\vec{x}}
\newcommand{\xp}{\x\textprime}
\newcommand{\zp}{\z\textprime}
\newcommand{\noteP}[1]{\textcolor{red}{#1}}

\begin{abstract}
We propose a new method to evaluate GANs, namely EvalGAN. EvalGAN relies on a test set to directly measure the reconstruction quality in the original sample space (no auxiliary networks are necessary), and it also computes the (log)likelihood for the reconstructed samples in the test set. Further, EvalGAN is agnostic to the GAN algorithm and the dataset. We decided to test it on three state-of-the-art GANs over the well-known CIFAR-10 and CelebA datasets.
\end{abstract}


\section{Introduction}
\label{sec:intro}

Implicit generative modeling, in general, and Generative Adversarial Networks (GANs), in particular, promise to solve the universal simulator problem in an end-to-end fashion \citep{G1, G2, G3}. 
GANs have been successfully applied to a variety of tasks, such as image-to-image translation \citep{Aimimtrans}, image super-resolution \citep{SRes}, image in-painting \citep{APaint}, domain adaptation \citep{ACrossDom2}, text-to-image synthesis \citep{StackGAN}, dark matter estimation \citep{ADarkMatter}, and breaking federated learning systems \citep{Hitaj17}, among many others. 

Progress in GANs has been quite remarkable and fast in the past four years. Most of the work has concentrated on improving its training to make it more stable, robust and generalizable to numerous architectures and datasets \citep{D1, WGANGP, WGAN,  MMDGAN, SNDCGAN} to name a few. 
There has also been significant progress on theoretical aspects of GAN convergence to the underlying density \citep{T1, T2, T3, T4}, 
and on their quantitative evaluation \citep{Q1, Q2, PRD}.
This is the topic that occupies us on this paper.

Generating realistic looking natural images is a challenging unsolved problem and it has the advantage that it can be visually demonstrated (i.e. look at the pictures that I can generate), which explains why GANs research has zeroed in their generation. But, in order to evaluate quantitatively if the images generated by any GAN have the same properties than the images from our training set, we have moved to Inception-based metrics: Inception Score \citep{IS}, Fr\'echet Inception Distance \citep{FID} or Precision and Recall for Distributions \citep{PRD}, which can only be used for evaluating natural images and limits the evaluation of GANs for other problems, in which there might not be a general accepted tool like Inception \citep{inceptionv4} to evaluate the quality of the generated samples. Furthermore, for natural images, Inception-based metrics are being criticized because it seems that most GAN algorithms achieve similar performance with proper hyperparameter optimization and random restarts \citep{Q1}. 
Finally, GANs are solely validated by using iid samples from the generator network without using an out-of-sample test set because direct likelihood evaluation for that test set is not possible and, even argued, that it might not be the right metric because quality and likelihood might not be related \citep{EGM1}. 

In this paper, we argue that we should still be interested in the likelihood of test samples even when it is not correlated with image quality, because it will inform us if the samples cannot be generated at all (i.e. mode dropping). We propose a procedure to directly evaluate GANs, namely EvalGAN, using a test set, as it is customary in most machine learning algorithms, and without relying on Inception \citep{inceptionv4} or any other auxiliary network. 
EvalGAN measures two different and relevant metrics for understanding the quality of a trained GAN: reconstruction quality and marginal likelihood for the reconstructed test sample. 

First, we measure how good we can reconstruct any given sample. Since GANs typically map a lower dimensional random input to higher dimensional space, there might be some reconstruction error that we want to account for, e.g. not every image might be reconstructed equally well or at all. Second, and irrespectively of the sample quality, we measure the marginal likelihood of each reconstructed sample, because it provides us with an indication of the regions in the sample space that we are over-repressing or fully ignoring. 
One key aspect of EvalGAN is the need to define a metric in the sample space that captures the complexity of each problem and that we can rely on to define quality and marginal likelihood for any sample. 

In this paper, we are agnostic about what GAN to use. Our evaluation method is demonstrated using Wasserstein GANs \citep{WGAN}, WGAN with gradient penalties (WGAN-GP) \citep{WGANGP}, and Spectral-normalized DCGANs \citep{SNDCGAN} trained over both CIFAR10 and CELEBA datasets. Our code can be accessed at \url{https://github.com/psanch21/EvalGAN} and can be used over any GAN. 
%

\section{Literature Review}
\label{sec:review}

Measuring GAN performance and quality is proving to be elusive, because, in high dimensional spaces, there are many ways in which the generated samples are different from true samples. When we compare samples in the original sample space those differences are more significant than the striking similarities \citep{ LR3,LR4}.

Given that GAN advances are driven by natural image generation and that we have a general tool for classifying them, i.e. Inception, we have settled for comparing images with it. 
The well-known IS \citep{IS} and FID \citep{FID} are the prime example for this evaluation trend. 
Recently, to improve on FID, \citep{PRD} proposes two metrics that resemble precision and recall for understanding how good the generated samples cover the training samples and vice versa, allowing to understand the different failure modes of GANs. Also, in \citep{LR1}, the authors have proposed a goodness-of-fit that inform us in linear time about the regions in which each GAN might perform best. Even when both of these procedures are explained in general terms, they are tested on features from the last pooling layer from Inception, as  for FID. The main criticism for these metrics is the need for Inception, as it is unclear how such a solution can be extended to GANs for other samples spaces. 

EvalGAN first computes the noise input that generates the GAN sample with the lowest distortion w.r.t. the original image, leading to a direct comparison between the test image and its best GAN reconstruction.   
This reconstruction has been previously applied to explore the visual manifold of GANs in \citep{zhu2016} 
and briefly introduced in the experimental section of \citep{Metz17} 
 for illustrating their GAN performance for a few training examples. However, those authors do not advocate for this error measure to be used as the main tool for evaluating GANs. On the contrary, we see this measure as the central measure to understand the quality of the samples being generated by the GAN.

Finally, \citep{LR2} proposes to used Annealed importance sampling to compute a lower bound to log-likelihood of a test set and showed it was two-orders of magnitude better than KDE. The authors only use low dimensional noise input and test with MNIST. They assume the reconstruction error does not affect the likelihood of the generated samples and they do not noticed that for more challenging datasets and higher dimensional input spaces, the generated test samples would lie outside the typical set for the given input noise distribution. Hence, their estimated likelihood would be biased by the sample's reconstruction quality. In this paper we measure both of them independently.


\section{EvalGAN}
\label{sec:method}

\begin{figure}[ht!]
	\vskip 0.2in
	\begin{center}
		\centerline{\includegraphics[width=14cm,height=7cm]{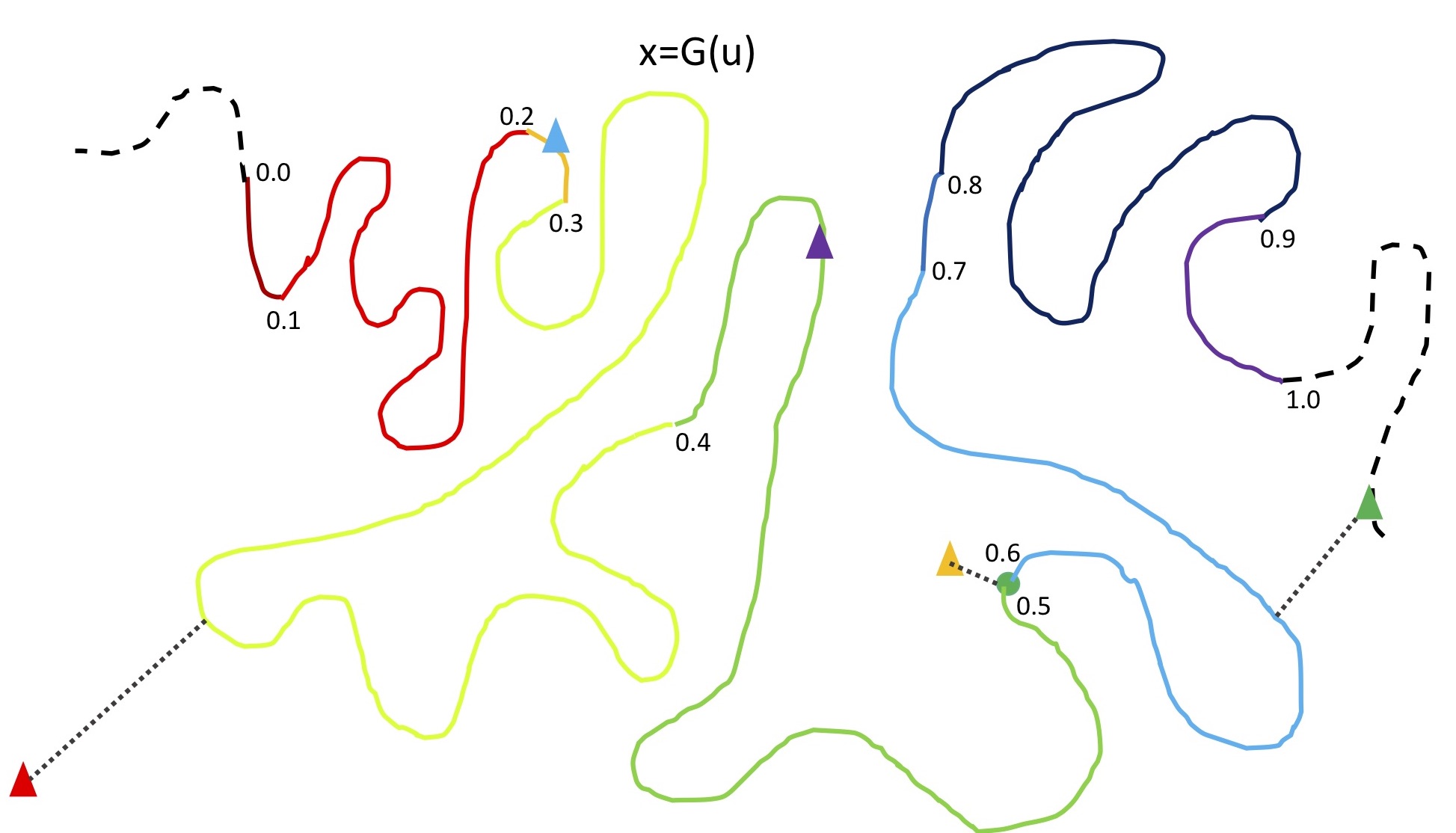}}
		\caption{In this figure we show a cartoon that illustrates the need to evaluate GANs in two dimensions: quality and probability of being sampled. Details about the image meaning are described in the motivation of Section \ref{sec:method}.}
		\label{fig1}
	\end{center}
	\vskip -0.2in
\end{figure}

To illustrate the two different types of evaluations that we want to address with EvalGAN and why they are both different and relevant, we show a cartoon representation in Figure \ref{fig1}. For this cartoon, we assume the input to the GAN is a one-dimensional uniform distribution between 0 and 1 and the output is a two-dimensional vector.  In this example and throughout the paper, we take $\z$ to be input noise to the generative deep neural network $G(\cdot)$ and $\x$ denotes the output space. 

The five triangles in the plot represent five test samples and the continuous line represents the manifold of all the points in the 2D space that the GAN can produce. This line is divided in 10 segments (note that one of them, the green dot, has a point mass of 0.1) and each one of them has equal probability. If we assume a Euclidean metric is valid for the 2D space, we can easily see that the points in the longer segments are less probable than those in the shorter segments. 

Note that the cyan and purple test sample are reconstructed with very low error, and the cyan triangle has higher probability than the purple triangle, because it lies on a shorter segment. The orange triangle is generated with some non-negligible error (represented by the dotted line), but its reconstruction is generated 10\% of the time. The red triangle represents a sample that it is reconstructed poorly and with low probability. 

Finally, we have extended the manifold for values less than 0 and greater than 1 with dashed lines. During training we are not generating samples from this part of the manifold, so we are not controlling what points on the manifold they express but nevertheless we could generate those samples by changing the input distribution. The green triangle shows a sample that can be reconstructed with low error if we do not limit the input $z$ to be between 0 and 1, but presents a high error otherwise. Even though this might look like a fringe example, our results demonstrate that we see this case repeatedly in practice.


\subsection{EvalGAN: reconstruction quality}\label{quality}
 
 
Given a test set sample, $\mathbf{x}_{\text{test}}$, we find the best approximation the GAN can generate by solving the following optimization problem:
\begin{equation}\label{eqz}
\z^*=\arg\min_\z d\left(\mathbf{x}_{\text{test}},G(\z)\right),
\end{equation} 
where $\z^*$ represents the input noise to the GAN to generate $\x^*=G(\z^*)$ as the sample that it is closest to $\mathbf{x}_{\text{test}}$, as defined by the suitable metric $d(\cdot,\cdot)$. 
The solution to this problem can be easily found by standard back-propagation, as it is done for generating adversarial training examples \citep{AdvSam1, AdvSam2}.

We have found that when solving \eqref{eqz} the values of $\z^*$  end up being far from the examples that can be generated by the input distribution\footnote{This issue was not reported in \citep{zhu2016,Metz17}, where this optimization was previously proposed.}. 
For example, if $\z$ is uniformly distributed the values of $\z^*$ found after solving \eqref{eqz} are outside the valid range.
 If $\z$ is a zero-mean unit-covariance Gaussian, the squared norm of $\z$ tends to be much larger than the dimension of $\z$, i.e. the values of $\z^*$ are (far) outside the typical set for a (high-dimensional) Gaussian \citep{CoverThomas}. Furthermore, these deviations are more significant as the dimension of $\z$ increases. 
Hence, we also propose solving the following constraint optimization problem:
\begin{equation}\label{eqz2}
\z_c^*=\arg\min_\z d\left(\mathbf{x}_{\text{test}},G(\z)\right)\ \ \ \text{s.t.}\  ||\z||^2\leq \dim(\z)+\delta,
\end{equation} 
when $\z\sim\mathcal{N}(\mathbf{0},\mathbf{I})$ and we denote $\x_c^*=G(\z_c^*)$. In our experiments, we set $\delta$ to zero because most $\z_c^*$ tend to be in the upper bound ($||\z||^2= \dim(\z)$) and for high-dimensional input spaces it should not matter, as the norm of any randomly generated sample $\z\sim \mathcal{N}(\mathbf{0},\mathbf{I})$ concentrates around $\sqrt{\dim(\z)}$ \citep{CoverThomas}. For uniformly distributed $\z$, the necessary constraints are straightforward. In the experimental section, we show examples when the optimization is carried out with and without constraints and for some GANs and some samples the difference are quite significant.

\subsection{EvalGAN: marginal likelihood}\label{evallik}

The likelihood of the test samples can be computed as follows:
\begin{equation}\label{like}
p(\x_{\text{test}})=\int p(\x_{\text{test}}|\z)p(\z)d\z,
\end{equation} 
In \citep{LR2}, the authors proposed an isotropic Gaussian likelihood for GANs, i.e:
\begin{equation}\label{aprox_LR2}
p(\x_{\text{test}}|\z)\approx\frac{1}{(2\pi\sigma^2)^{\dim(\z)}}\exp\left(\frac{||\x_{\text{test}}-G(\z)||^2}{2\sigma^2}\right).
\end{equation}
They solved the integral in \eqref{like} by annealed importance sampling. This is a fine choice if all samples in the test set could be matched to a $\z$ (i.e. there exist a $\z_{\text{test}}$ for which $\x_{\text{test}}=G(\z_{\text{test}})$) or the reconstruction error is similar (and small) for all test samples. But when the reconstruction can be uneven, best reconstructed images would seem more likely, which does not need to be the case, and setting the value of $\sigma$ would be extremely hard. 

This effect can be easily appreciated in the cartoon example in Figure \ref{fig1}, as a small $\sigma$ would lead to the orange and red triangles presenting negligible likelihoods compared to the cyan and purple triangles, while a large $\sigma$ would boost the likelihood of the orange triangle, because it is close to highly probable $z$. The value of $\sigma$ would significantly affect the measured likelihood of the samples in ways that does not illustrate the quality or likelihood of any GAN.

In the previous subsection, we advocated for computing the quality of the reconstruction independently on how likely they could be generated. In this section, we now compute the likelihood of this reconstruction by counting all the $\z$ that can generate the same reconstruction with a negligible error:
\begin{equation}\label{approx0}
p(\x_{\text{test}})\approx \int_{\small{d(\x^*_c,G(\z))<T}}\!\!\!\!\!\!\!\!\!\!\!\!\!\!\!\!p(\z)d\z \approx \frac{1}{N}\sum_i \mathbb{I}_{[d(\x^*_c,G(\z_i))<T]}
\end{equation} 
where $T$ is a threshold to ensure that $G(\z_i)$ is close enough to $\x_c^*$, $\z_i$ are iid samples from $p(\z)$, and $\mathbb{I}_{[d(\x^*_c,G(\z_i))<T]}$ is an indicator function that it is one if the condition holds and zero otherwise. We can (and should) set $T$ to be significantly smaller than $d(\x_{\text{test}},\x_c^*)$, which is the error of the best reconstruction of the test sample \footnote{For a Euclidean metric our approximation is equivalent to changing $\x_{\text{test}}$ by $\x_c^*$ in the righthand side of \eqref{aprox_LR2}.}. In this case, $\z_c^*$ generates $\x_c^*$ and we have decoupled measuring the reconstruction quality and how likely the generated sample can be.  

We could also use $\x^*=G(\z^*)$ instead, where $\z^*$ is the solution to \eqref{eqz}, but we show in the experimental section that those samples would not be generated when sampling from $p(\z)$. The likelihood of $\x^*$ would be negligible compared to the likelihood of $\x_c^*$. When $\x^*$ is a better reconstruction than $\x_c^*$ emphasizes the need for separating both measures (quality and likelihood), because even if we could recover $\x^*$ by backpropagation, it would never be generated by sampling from $p(\x)$. This also remarks that setting $\sigma$ in \eqref{aprox_LR2} would be challenging, while setting $T$ in our case is fairly straightforward.  

Of course, for typical GANs, in which the dimension of $\z$ is the hundreds, the approximation in \eqref{approx0} is impractical at best. We now present three approximations that can be easily computed. We advocate for the last one, as it is the most computationally efficient and accurate of the three.

\paragraph{Isotropic samples.} We can approximate the log likelihood as follows:
\begin{equation}\label{L2}
\log p(\x_{\text{test}})\approx \dim(\z)\log \bar{\sigma}_\epsilon-\log Z,
\end{equation}
where
\begin{equation}\label{sigmamax}
\bar{\sigma}_\epsilon=\arg\max_{\sigma_\epsilon}\left(\frac{1}{N}\sum_{i=1}^N d(\x_c^*,\x_i^*)\right)\leq T,
\end{equation}
$\x_i^*=G(\z_c^*+\epsilon_i)$, and $\epsilon_i\sim\mathcal{N}(0,\sigma_\epsilon^2\mathbf{I})$. The partition function $Z$ only depends on $p(\z)$ and it is independent of the GAN, because by construction all $p(\z_c^*)$ have the same probability. 


If the curvature of $G(\cdot)$ changes considerably in different dimensions of $\z$ the previous measure benefits those samples that are in a more isotropic region of $G(\cdot)$, because it underestimates the probability of those sample in which $G(\cdot)$ changes significantly in different directions.  

\paragraph{Non-isotropic samples.} We can adapt the previous measure to account for differences in the curvature of $G(\cdot)$, by instead computing:
\begin{equation}\label{L3}
p(\x_{\text{test}})\propto N_c^*/N,
\end{equation}
where
\begin{equation}\label{Threshold2}
N_c^*=\sum_{i} \mathbb{I}_{[d(\x^*_c,\x_i^*)<T]},
\end{equation}
and $N_c^*$ counts how many $\x_i^*$ samples  are sufficiently close to $\x_c^*$, when $\sigma_\epsilon$ is small and fixed.

Selecting a good $\sigma_\epsilon$ to ensure that $N_c^*$ is nonzero for a given $N$ and for all the test samples can be hard (and require a very large $N$), if the marginal likelihood for all the test samples vary substantially (which they do).

\paragraph{Proposed measure.} Finally, by combining the previous two approximations we get:
\begin{equation}\label{L4}
\log p(\x_{\text{test}})\approx \log\frac{N_c^*}{N} + \dim(\z)\log \sigma_\epsilon-\log Z.
\end{equation}
This approximation becomes more accurate as we increase $\sigma_\epsilon$, because we are able to capture all the directions in $\z$-space in which the samples $\x_i^*$ are close enough to $\x_c^*$. This approximation can be computed accurately by gradually increasing $\sigma_\epsilon$ and $N$. In our simulations, we set the maximum $N$ to 10,000 and we stop increasing $\sigma_\epsilon$ when $N_c^*$ drops below 100.

\subsection{EvalGAN: metric} 

One of the aspects that we have not investigated in this paper is the selection of the ideal metric, i.e. $d(\cdot,\cdot)$. Defining this metric correctly is crucial for EvalGAN to succeed at evaluating GANs and it should be carefully selected by each different problem. The different communities using GANs for creating universal simulators, should coalesce around the relevant metric for evaluating their GANs with EvalGAN.

In this paper, we illustrate three different GANs by generating natural images (CIFAR-10 and CelebA) and we have used the well-know Peak Signal-to-Noise Ratio (PSNR) typically used in image compression:
\begin{align*}
\text{PSNR}(\x_i,\x_j ) = 10 \log_{10}\left( \frac{M^2}{\text{MSE}(\x_i, \x_j)} \right),
\end{align*}
where $M$ is the maximum possible pixel value of the images, i.e. 255 for 8-bit color images. The higher the PSNR (in dB) leads to higher image quality. The Mean Squared Error (MSE) of color images can be computed as follows:
\begin{align}
\label{eq:MSE}
	\text{MSE}(\x_i,\x_j) = \frac{||\x_i - \x_j ||_ 2^2}{3K},
\end{align}
where $K$ is the number of pixels in the images. 

In this paper, we have opted for a simple metric. We understand that other metrics for images in which smoothness or other properties of the generated images are captured might be more relevant. We are not specially advocating for PSNR, except that it relates to image quality and it is easy to understand and compute.

\section{EvalGAN in practice}
\label{sec:experiments}


\subsection{Experimental Setup}

Three different state-of-the-art GANs have been considered: Wasserstein GAN (WGAN) \citep{WGAN}, Improved WGAN with gradient penalty (WGAN-GP) \citep{WGANGP}  and deep convolutional GAN with spectral normalization (SN-DCGAN) \citep{SNDCGAN}.  Tensorflow implementation for the three of them are publicly available. To facilitate reproducibility of our results, in the Appendix we provide an exhaustive description of the parameters selected to construct both the generator and discriminator networks and those regarding the training process. To train all models, we consider as input a Gaussian noise model: $\mathbf{z}\sim \mathcal{N}(\mathbf{0},\mathbf{I})$, with $\text{dim}(\z)\in [16, 32, 64, 128, 256]$. To solve the optimization problem in \eqref{eqz} and \eqref{eqz2}, we use Adam algorithm \citep{ADAM} with parameters $\alpha=0.005$ (learning rate), $\beta_1=0.9$ and $\beta_2=0.999$  and a stopping tolerance of $0.1$ in 3000 iterations. For solving \eqref{eqz2}, we project the norm of $\z$ to the unit hypersphere if the norm of $\z$ is larger than $\sqrt{\text{dim}(\z)}$.

CIFAR10 is taken as the main running example in this section to illustrate our discussion and the quality metrics proposed. CIFAR10 contains 50,000  images for training and 10,000 images for test. Further experiments using the CelebA dataset are mainly included in the Appendix. In CelebA, 2,000 face images are used for test and 200,000 images for training. The results in this section refers to the SN-DCGAN algorithm, while WGAN and WGAN-GP are reported in the Appendix.

\subsection{Assessing reconstruction quality in EvalGAN} \label{reconstruction}

\begin{figure}[h!]
	\vskip 0.2in
	\begin{center}
		\centerline{\includegraphics[scale=1.0]{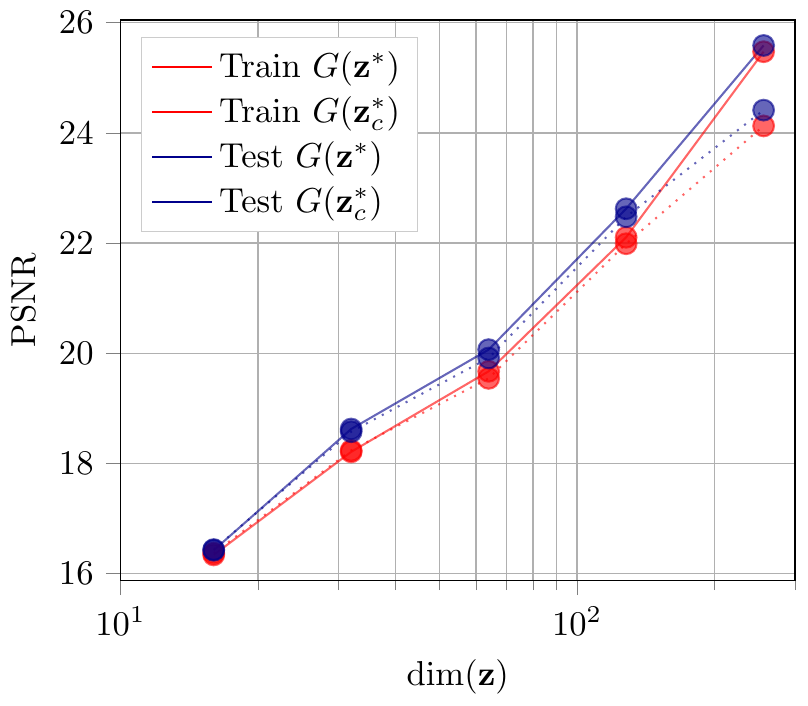}}
		\caption{Evolution of the average PSNR between the original image and its reconstruction with the dimension of the latent space for SN-DCGAN trained using CIFAR10. Solid lines correspond to the reconstruction PSNR using the unconstrained projection found solving \eqref{eqz}. Dashed lines correspond to the reconstruction PSNR using the constrained optimization in \eqref{eqz2} with $\delta=0$. }
		\label{fig:MSE_dim}
	\end{center}
\end{figure}

We first analyze the influence of the generator input-dimension on the GAN reconstruction quality. 
We compute $\z^*$ for the images in the test set using the solution to the  unconstraint  problem in \eqref{eqz}, once the GAN has been trained. The solid lines in Figure \ref{fig:MSE_dim} show the evolution of the average PSNR  with respect to $\text{dim}(\z)$, as expected the image quality improves with $\text{dim}(\z)$. Also, it is remarkable that the reconstruction quality of test samples is as good as those in the training set.

\begin{figure}[h!]
	\centering
	\subfigure[]{\hspace{-1.25cm}\includegraphics[scale=1.0]{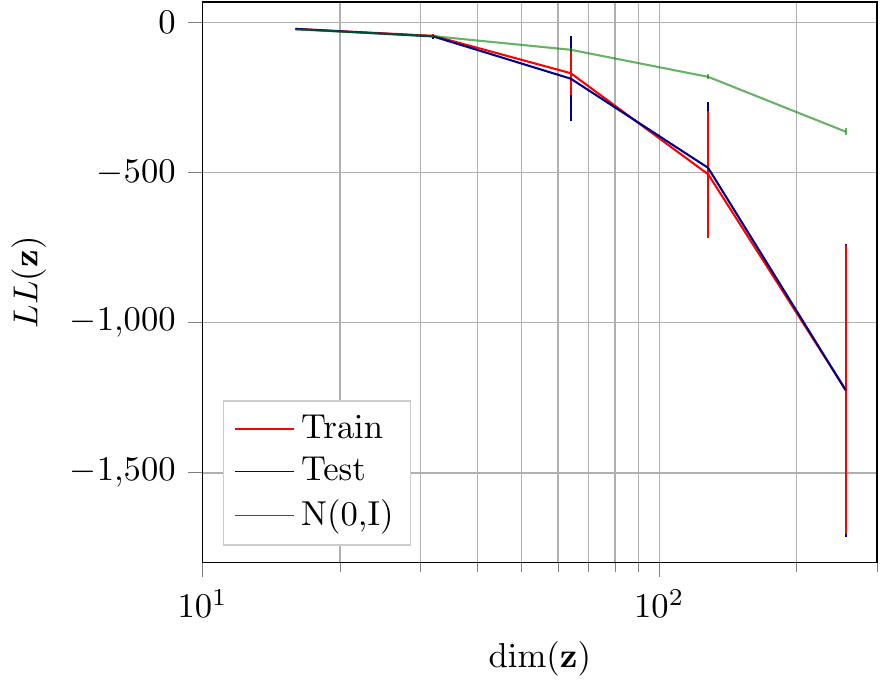}}%
	\subfigure[]{\includegraphics[scale=1.0]{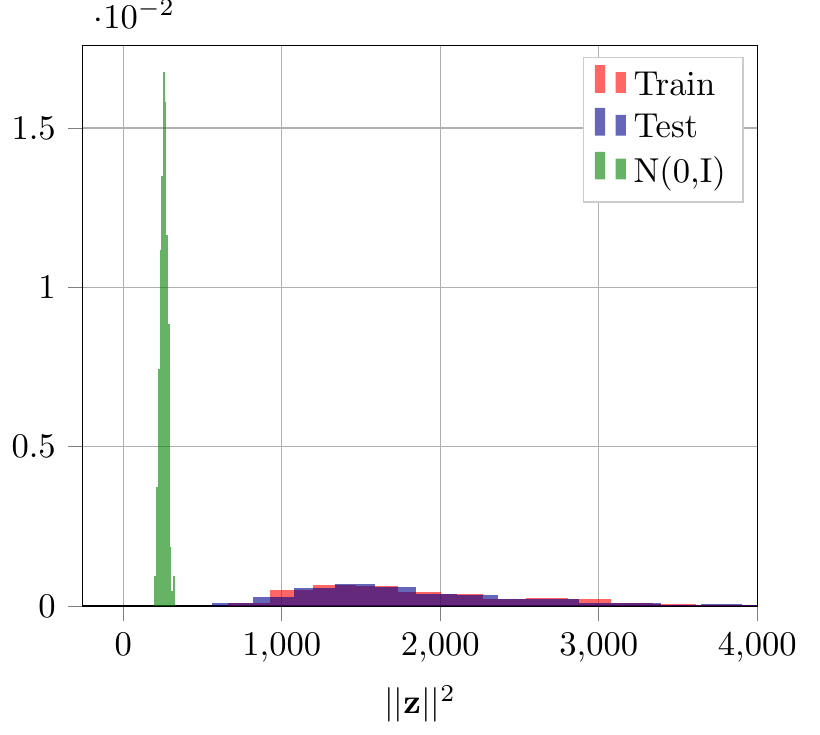}}%
	\caption{In (a), evolution of the average log-likelihood of the $\z^*$ solutions of the unconstrained problem in \eqref{eqz} computed over both the training and the test set for the SN-DCGAN trained with the CIFAR10 dataset. In (b), we show  the histogram of $||\z^*||^2$ for the case  $\text{dim}(\z)=256$. }%
	\label{fig4}
\end{figure}

We also found that (almost) all $\z^*$ samples lie outside the typical set and hence the found images would never be generated when sampling from $p(\z)$. This effect has not been previously reported in the literature and it shows that during the optimization of the GAN we are not controlling accurately the mapping from $\z$ to $\x$. This issue is illustrated in Figure \ref{fig4}(a), where we show the average $\log p(\z)$ for the test and training samples and we compare it with the $\log p(\z)$ of the samples from the typical set. In Figure \ref{fig4}(b), we show the histogram for $||\z^*||^2$ from the training and test samples, as well as the histogram of samples from $p(\z)$ for $\text{dim}(\z)=256$. It is fairly obvious the values of $\z^*$ would never be sampled in practice.

\begin{figure*}[t]
\hspace{-0.75cm}
		\begin{tabular}{cc}
			\hspace{-0.1cm} Image \;\hspace{-0.05cm}$G(\z^*)$ \;\hspace{-0.1cm}$G(\z_c^*)$ Image  $G(\z^*)$ $G(\z_c^*)$ & \hspace{-0.1cm} Image \;\hspace{-0.05cm}$G(\z^*)$ \;\hspace{-0.1cm}$G(\z_c^*)$ Image  $G(\z^*)$ $G(\z_c^*)$   \\
			\includegraphics[scale=0.7]{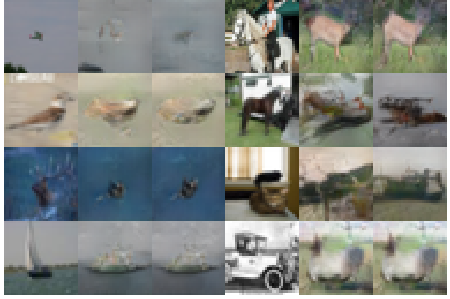} & \includegraphics[scale=0.7]{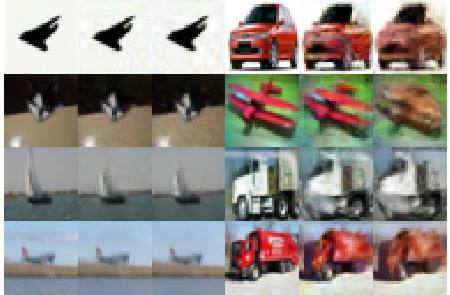}\\
			(a) $\text{dim}(\z)=16$ & (b) $\text{dim}(\z)=256$
		\end{tabular}
	\caption{Each figure contains two groups with 3 columns each. The left group represents the test samples with largest PSNR, while the right group contain samples with the lowest PSNR values.}%
	\label{fig:PSNR_images}
	\vskip -0.05in
\end{figure*}

As advanced in Section \ref{quality}, we also advocate to constraint $\z^*$ to be in the typical set of $p(\z)$. The dashed lines in Figure \ref{fig:MSE_dim} represent the PSNR of the original image w.r.t. $G(\z_c^*)$, where $\z^*_c$ is found by solving \eqref{eqz2}. There is a noticeable degradation for high-dimension inputs in both train and test sets.  In Figure \ref{fig:PSNR_images} we show some test set examples reconstructed with $\z^*$ and $\z^*_c$. In (a) we use $\dim(\z)=256$ and in (b) $\dim(\z)=16$. In the lefthand side of each subplot, we report the images with largest PSNR and, in the righthand side, we show the images with the lowest PSNR values. For the high quality reconstructions, there is little visual difference between the constraint and unconstraint optimization and the input dimension does not seems to affect the reconstruction that much.
For the lower quality reconstructions and $\text{dim}(\z)=256$, the differences are quite significant between the three images, but still the objects are recognizable in both reconstructions. For $\dim(\z)=16$ neither reconstruction is meaningful, showing larger dimensions for $\z$ are really needed.

\begin{figure}[h!]
	\centering
	\subfigure[]{\includegraphics[scale=0.8]{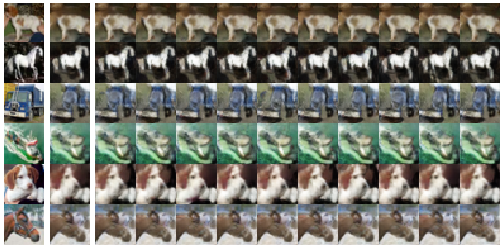}}
	\subfigure[]{\includegraphics[scale=0.8]{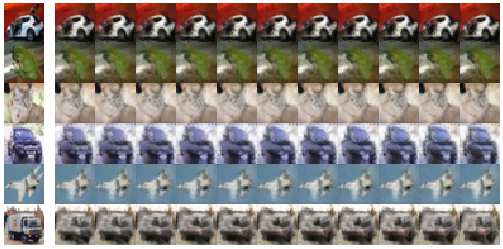}}%
	\caption{In (a) from left to right: real image, reconstruction using the latent mean $\sum_m \z^*_{c,m}/10$, and reconstruction using the solution to \eqref{eqz2} for 10 different initializations. For this experiment we consider $\text{dim}(\z)=256$. In (b), the leftmost image is the real sample. The rest are the reconstructions from linearly interpolated $\z$ values using \eqref{eqz2} for two different initializations. }%
	\label{fig:SIM_recons_z1_z2}
\end{figure}

To obtain the results above, we also checked if different initializations for $\z$ in \eqref{eqz2} lead to the same $\x^*_c=G(\z_c^*)$. In Figure \ref{fig:SIM_recons_z1_z2} (a), we show 10 different reconstruction for the same image from 10 different initializations, as well as the sample from the mean input noise sample, i.e.  $\z^*_{c,p}=\sum_m \z^*_{c,m}/10$, where $\z^*_{c,m}$ are each one of the 10 solutions to \eqref{eqz2} with the same test image. The first column is the original image, the second column represents the image coming from $G(\z^*_{c,p})$ and the last 10 columns shows each one of the individual reconstructions $G(\z^*_{c,m})$. 
We also took the two $\z^*_{c,m}$ that were further apart and linearly interpolate their values to generate the images in between. These images are shown in Figure \ref{fig:SIM_recons_z1_z2} (b) with similar behavior as the previous experiment. Similar conclusions can be drawn when we perform polar interpolation instead of linear interpolation. In short, even if the optimization problems are not convex and uni-modality is not enforced by GAN training, we did not find issues with either. 

\subsection{EvalGAN marginal likelihood}

\begin{figure}[h]
	\centering
	\subfigure[]{\includegraphics[scale=1.0]{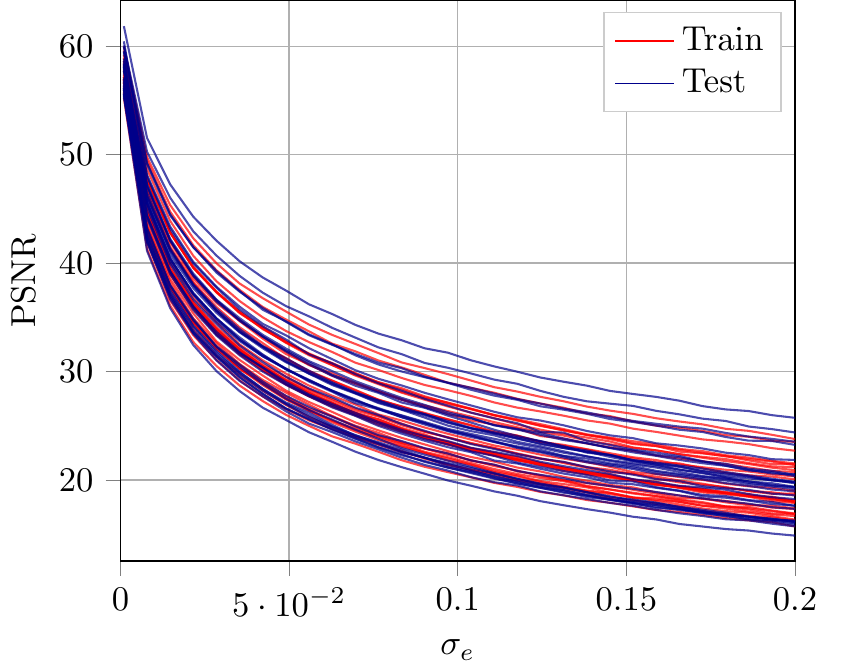}}
	\subfigure[]{\includegraphics[scale=1.0]{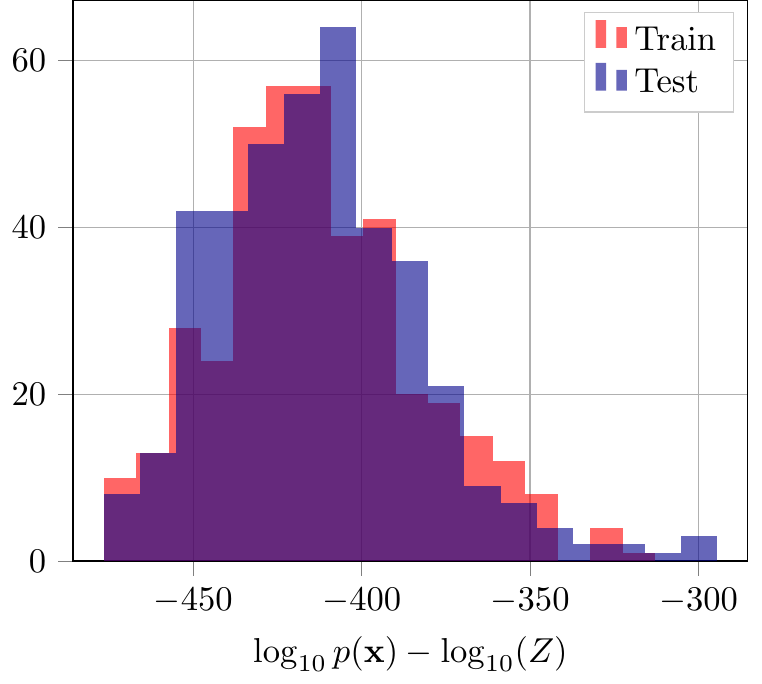}}%
	\caption{In (a), we show the evolution of the SN-DCGAN average PSNR for 20 CIFAR10 images between $G(\z_c^*)$ and $G(\z_c^*+\epsilon_i)$ as a function of $\sigma_\epsilon$, where $\epsilon_i \sim \mathcal{N}(\mathbf{0},\sigma_\epsilon^2\mathbf{I})$. In (b) we show the unnormalized marginal likelihood histogram for the SN-DCGAN using  \eqref{L4} for 400 CIFAR10 images and a 40 dB PSNR threshold.}%
	\label{fig:z_region_PSNR_norm_some_logpx}
\end{figure}

We now concentrate in evaluating the likelihood of the reconstructed images, independent of their reconstruction quality. First, in Figure \ref{fig:z_region_PSNR_norm_some_logpx} (a) we show the evolution of $ d(\x_c^*,\x_i^*)$ as a function of $\sigma_\epsilon$ for 20 train and 20 test samples. We can see that the degradation of the samples varies considerably. For example, if we set the threshold for the PSNR at 40dB (much larger than the 25dB reconstruction error reported in Figure \ref{fig:MSE_dim} the image with the largest $\sigma_\epsilon$, for which this mean reconstruction quality is achieved, is above 0.04. For the image with lowest $\sigma_\epsilon$, before the quality threshold is met, is below 0.01. This means that the most probable image in the set is at least $(0.04/0.01)^{256}\approx10^{154}$ more probable than the least likely image and we are only comparing 20 random samples in this plot.

%
%

\begin{figure}[h!]
	\centering
	\hspace{-0.5cm}\includegraphics[scale=0.35]{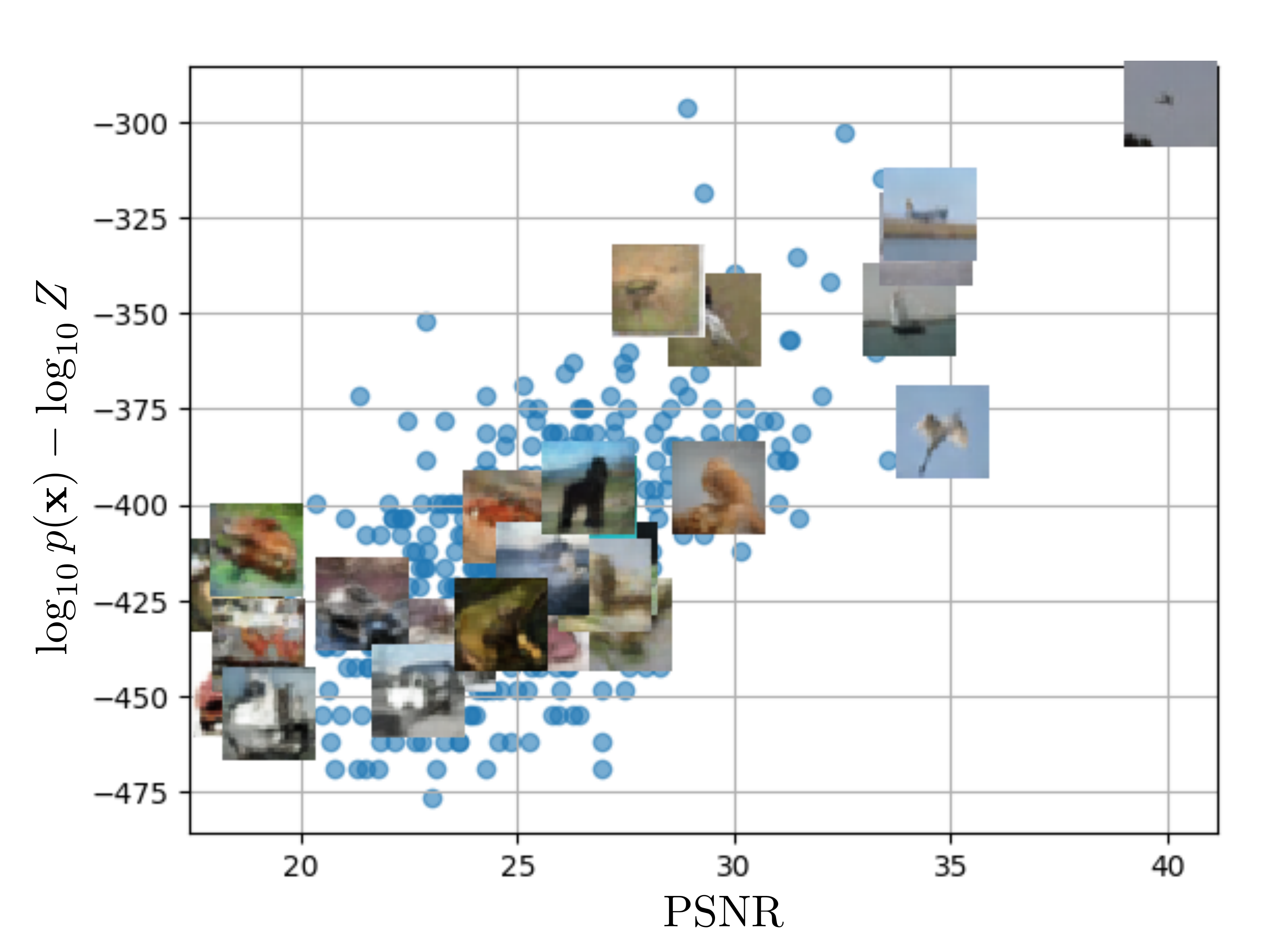}
	\caption{Scatter plot to compare the log unnormalized marginal likelihood in \eqref{L4} with the PSNR between the real image and $G(\z_c^*)$ for 400 test CIFAR10 images using SN-DCGAN with $\text{dim}(\z)=256$.}\label{scatter}
	\vskip -0.2in
\end{figure}

We now turn to computing the log-likelihood for 400 images using the approximation in \eqref{L4}, in Figure \ref{fig:z_region_PSNR_norm_some_logpx} (b) we show the histogram of $\log_{10} p(\x) -\log_{10} Z$. We use a threshold  $T$ in \eqref{Threshold2} corresponding to a PSNR w.r.t. to $G(\z_c^*)$ of 40 dB.  Note that the few images in the right-most tale of the histogram are $10^{125}$ times more probable of being generated than those in the mode of the histogram, and are $10^{175}$ times more likely than those in the left tail of the histogram. Hence, at a sample level, we are able to point exactly where overrepresentation and mode dropping occurs. The log-likelihood distribution is similar for the training and test sets, it does not seem to be an over-representation of the samples in the training set. 


\begin{figure}[h!]
	\centering
	\subfigure[CIFAR10]{\includegraphics[scale=0.15]{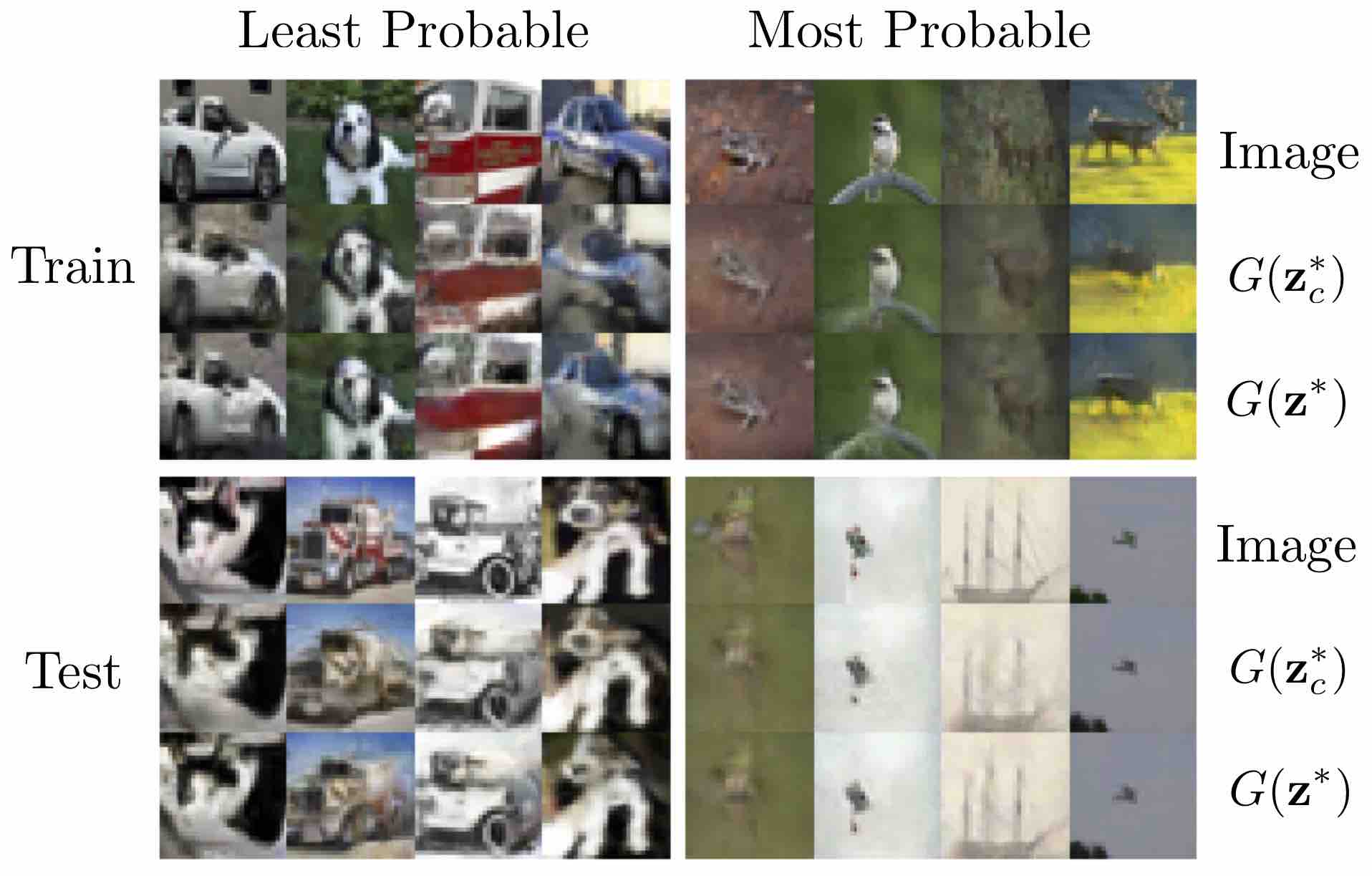}}
	\subfigure[CelebA]{\includegraphics[scale=0.15]{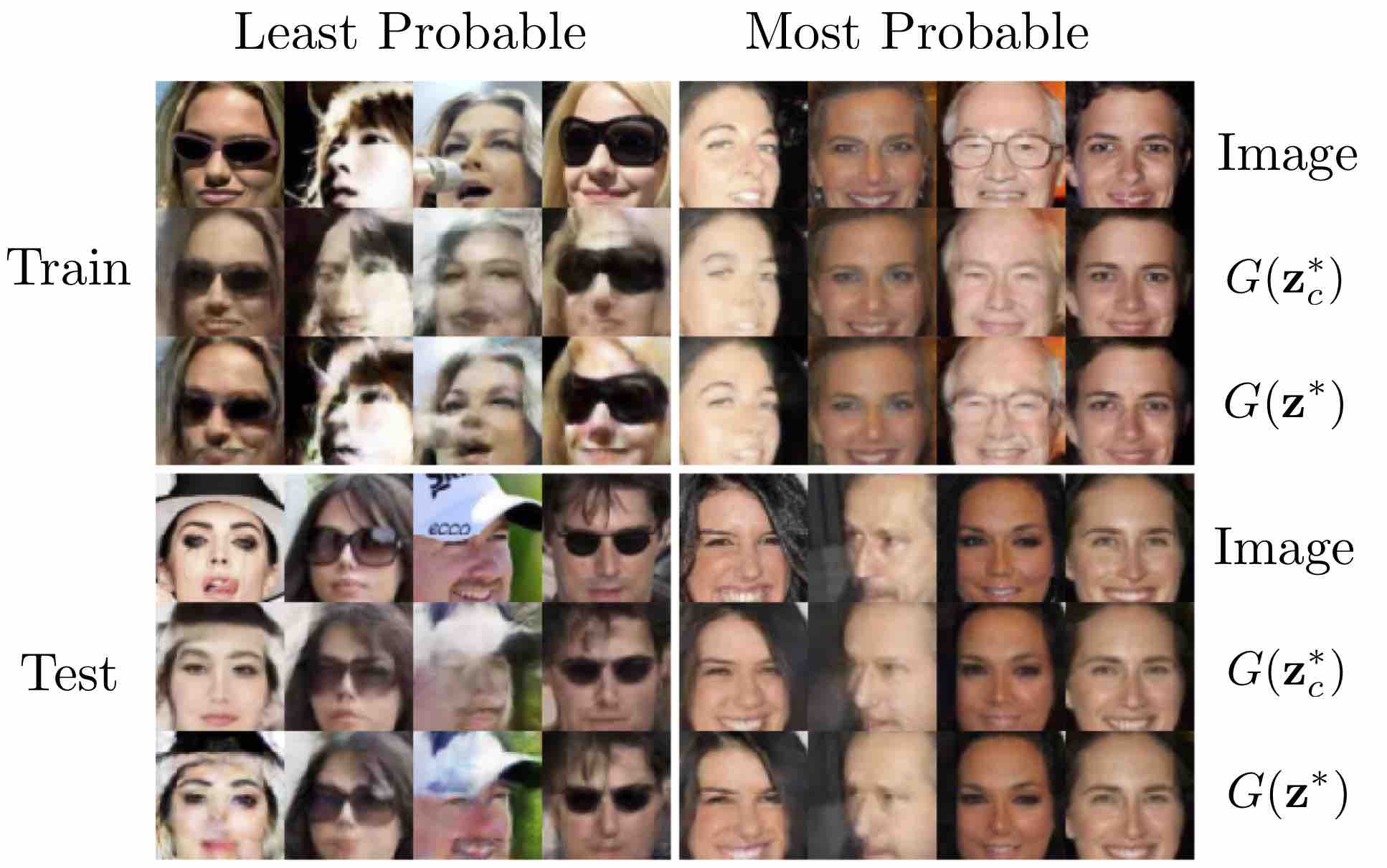}}%
	\caption{In (a), we plot the most and least probable images for SN-DCGAN and CIFAR10 according to \eqref{L4}. In (b), we repeat the experiment for CelebA. In both cases $\text{dim}(\z)=256$.}%
	\label{figprobs}
\end{figure}

In Figure \ref{scatter}, we compare the log unnormalized marginal likelihood with the reconstruction PSNR between the real image and $G(\z_c^*)$ for 400 test CIFAR10 images using SN-DCGAN. First, we can notice that the dynamic range of both likelihood and PSNR is quite large, especially the former. We can also observe that images with simpler textures and large uniform backgrounds are not only reconstructed with better quality, but also they are being overrepresented by the generator network. In Figure \ref{figprobs}, we compare the reconstruction of some of the most likely and least likely images with the original image and we can easily see this effect too. In the plots, we have added the reconstruction with the unconstraint optimization problem for completeness.

We also include the results CelebA, in which the most likely images seem to contain plain faces with soft smiling gestures, while least likely samples in the set are associated to people that either have a weird posture or they are wearing glasses or hats. It is interesting to note that in CelebA, reconstructed images using the solution to \eqref{eqz2}, i.e. $\z_c^*$. tend to simplify the original image including features common in the set of most probable images, e.g. inserting soft smiles instead of more complicated gestures, or even removing objects like glasses, hats, or even a microphone. The solution to the unconstrained problem in \eqref{eqz}, i.e. $\z^*$, tend to partially keep those features. 

Finally, in the Appendix we reproduce the previous experiments using WGANs, WGAN-GP and SN-DCGAN with CIFAR10 and CelebA datasets.

\section{Discussion}
\label{sec:discussion}

The two measures that we have put forward in this paper, are very relevant when evaluating GANs and they have not been systematically used in the past. The reproduction quality tells us if a sample can be generated by the GAN and how good it matches the test sample \footnote{This measure had been proposed previously in \cite{zhu2016, Metz17}, but has not been advocated for systematically evaluating GANs.}. The estimation of the log likelihood of the reconstruction (not the test sample) tell us how likely are we to see that reconstruction, which is the only image that the GAN can produce (This is a new metric proposed in this paper). Estimating the likelihood of the test sample directly is much harder and it mixes these two relevant metrics in one, making it useless to evaluate GANs, as already point it out in \cite{EGM1}.

The results in log likelihood estimation shows that training and test samples suffer significant over and under-representation issues that needs to be corrected when training GANs. We can use the mean log-likelihood to compare GANs, but we should also try to equalize the log-likelihoods for the training (and test) samples when training GANs. Because a difference in marginal likelihood of more than $10^{10}$ seems a bit extreme, in our opinion, and these differences happen for most pair of images (the largest difference are larger than $10^{150}$). 

We have also noticed that the samples that are more visually complex lead to lower reconstruction error and lower marginal likelihoods. For example, we can argue that the samples that present lower marginal likelihood can be over-sampled when training GANs, as we should not expect that harder to generate samples need to be seen an equal number of times that those that are easier to generate. This will also improve the reconstruction quality of these samples. 

In this paper, we have left open what the right metric for the different GANs would be. Is PSNR adequate or should we consider other distances for images? Also, what should be the right metric for generating text or speech? In general, for each problem, in which we want to evaluate GANs, we would need to design the right metric. 

Finally, we have not been able to apply EvalGAN to Variational Autoencoders (VAE), as we had wished for. EvalGAN can be used to evaluate the decoding network of VAEs the same way we proposed to evaluate the generative networks of GANs. Additionally, EvalGAN, given a test data set, can help compare the $\z_c^*$ given by \eqref{eqz2} with the $\z$ that is obtained from the encoding VAE network. Understanding if these two distributions are similar would tell us about how well the encoder and decoder have been trained and open a different way to further optimizing them. This has been left as further work. 

\subsection{The need for constraint optimization for evaluating the test samples}

One of the main results from using EvalGAN is an ancillary result that we were not expecting when we embarked on this project. The values of $\z^*$ in \eqref{eqz} are well off the typical set of that would be generated from $p(\z)$. When we constraint the result to be in the typical set the image quality degrades slightly, but still it does degrade and it is more apparent as $\text{dim}(\z)$ grows \footnote{In the Appendix we show that this effect is less pronounced for WGAN-GP, but the samples are still outside the typical set.}. 


Expecting that the distribution of $\z^*$ matches that of $p(\z)$ might be too much to ask for, because of biases in the available sets and the training of GANs and its architecture. But we should expect that $\z^*$ for both training and test samples should lie on the typical set of $p(\z)$ without needing to constrain it, because otherwise we would not be controlling the samples that GANs will be generating as well as we could. We believe that GAN training should be modify to account for this problem. This is probably the most important conclusion of this study. We have not figure out a way forward (yet).

\newpage
\acks{The work of Pablo M. Olmos and Pablo S\'anchez-Mart\'in is supported by   Spanish   government   MEC   under   grant   TEC2016-78434-C3-3-R,  by  Comunidad de   Madrid   under   grants   IND2017/TIC-7618, IND2018/TIC-9649, and Y2018/TCS-4705,    and   by   the European  Union  (FEDER).   We  also  gratefully  acknowledge the support of NVIDIA Corporation with the donation of the Titan X Pascal GPU used for this research.

 }

\vskip 0.2in
\bibliography{bibliography}

\begin{thebibliography}{36}
\providecommand{\natexlab}[1]{#1}
\providecommand{\url}[1]{\texttt{#1}}
\expandafter\ifx\csname urlstyle\endcsname\relax
  \providecommand{\doi}[1]{doi: #1}\else
  \providecommand{\doi}{doi: \begingroup \urlstyle{rm}\Url}\fi

\bibitem[Arjovsky et~al.(2017)Arjovsky, Chintala, and Bottou]{WGAN}
Martin Arjovsky, Soumith Chintala, and L{\'e}on Bottou.
\newblock {Wasserstein Generative Adversarial Networks}.
\newblock In \emph{Proceedings of the 34th International Conference on Machine
  Learning}, pages 214--223, International Convention Centre, Sydney,
  Australia, 2017.

\bibitem[Arora et~al.(2017)Arora, Ge, Liang, Ma, and Zhang]{T3}
Sanjeev Arora, Rong Ge, Yingyu Liang, Tengyu Ma, and Yi~Zhang.
\newblock {Generalization and Equilibrium in Generative Adversarial Nets
  ({GAN}s)}.
\newblock In \emph{Proceedings of the 34th International Conference on Machine
  Learning}, volume~70, pages 224--232, 2017.

\bibitem[Borji(2018)]{Q2}
Ali Borji.
\newblock {Pros and Cons of GAN Evaluation Measures}.
\newblock \emph{arXiv preprint arXiv:1802.03446}, 2018.

\bibitem[Cover and Thomas(1991)]{CoverThomas}
Thomas~M. Cover and Joy~A. Thomas.
\newblock \emph{Elements of Information Theory}.
\newblock Wiley-Interscience, New York, NY, USA, 1991.
\newblock ISBN 0-471-06259-6.

\bibitem[Goodfellow et~al.(2014{\natexlab{a}})Goodfellow, Pouget-Abadie, Mirza,
  Xu, Warde-Farley, Ozair, Courville, and Bengio]{G1}
Ian Goodfellow, Jean Pouget-Abadie, Mehdi Mirza, Bing Xu, David Warde-Farley,
  Sherjil Ozair, Aaron Courville, and Yoshua Bengio.
\newblock {Generative Adversarial Nets}.
\newblock In \emph{Advances in neural information processing systems}, pages
  2672--2680, 2014{\natexlab{a}}.

\bibitem[Goodfellow et~al.(2014{\natexlab{b}})Goodfellow, Shlens, and
  Szegedy]{AdvSam1}
Ian~J. Goodfellow, Jonathon Shlens, and Christian Szegedy.
\newblock Explaining and harnessing adversarial examples.
\newblock In \emph{International Conference on Learning Representations
  (ICLR)}, 2014{\natexlab{b}}.

\bibitem[Gulrajani et~al.(2017)Gulrajani, Ahmed, Arjovsky, Dumoulin, and
  Courville]{WGANGP}
Ishaan Gulrajani, Faruk Ahmed, Martin Arjovsky, Vincent Dumoulin, and Aaron~C
  Courville.
\newblock {Improved Training of Wasserstein GANs}.
\newblock In \emph{Advances in Neural Information Processing Systems}, pages
  5767--5777. 2017.

\bibitem[Heusel et~al.(2017)Heusel, Ramsauer, Unterthiner, Nessler, and
  Hochreiter]{FID}
Martin Heusel, Hubert Ramsauer, Thomas Unterthiner, Bernhard Nessler, and Sepp
  Hochreiter.
\newblock {GANs Trained by a Two Time-Scale Update Rule Converge to a Local
  Nash Equilibrium}.
\newblock In \emph{Advances in Neural Information Processing Systems}, pages
  6626--6637. 2017.

\bibitem[Hitaj et~al.(2017)Hitaj, Ateniese, and Perez-Cruz]{Hitaj17}
Briland Hitaj, Giuseppe Ateniese, and Fernando Perez-Cruz.
\newblock {Deep models under the GAN: information leakage from collaborative
  deep learning}.
\newblock In \emph{Proceedings of the 2017 ACM SIGSAC Conference on Computer
  and Communications Security}, 2017.

\bibitem[{Im} et~al.(2018){Im}, {Ma}, {Taylor}, and {Branson}]{LR4}
Daniel~J. {Im}, He~{Ma}, Graham {Taylor}, and Kristin {Branson}.
\newblock {Quantitatively Evaluating GANs With Divergences Proposed for
  Training}.
\newblock \emph{International Conference on Learning Representations (ICLR)},
  2018.

\bibitem[Isola et~al.(2017)Isola, Zhu, Zhou, and Efros]{Aimimtrans}
Phillip Isola, Jun-Yan Zhu, Tinghui Zhou, and Alexei~A. Efros.
\newblock {Image-to-Image Translation with Conditional Adversarial Networks}.
\newblock \emph{IEEE Conference on Computer Vision and Pattern Recognition
  (CVPR)}, pages 5967--5976, 2017.

\bibitem[Jitkrittum et~al.(2018)Jitkrittum, Kanagawa, Sangkloy, Hays,
  Sch\"{o}lkopf, and Gretton]{LR1}
Wittawat Jitkrittum, Heishiro Kanagawa, Patsorn Sangkloy, James Hays, Bernhard
  Sch\"{o}lkopf, and Arthur Gretton.
\newblock {Informative Features for Model Comparison}.
\newblock In \emph{Advances in Neural Information Processing Systems}, pages
  816--827. 2018.

\bibitem[Kingma and Ba(2014)]{ADAM}
Diederik Kingma and Jimmy Ba.
\newblock {Adam: A Method for Stochastic Optimization}.
\newblock \emph{International Conference on Learning Representations (ICLR)},
  2014.

\bibitem[Kingma and Welling(2014)]{G2}
Diederik~P Kingma and Max Welling.
\newblock Auto-encoding variational bayes.
\newblock In \emph{International Conference on Learning Representations
  (ICLR)}, 2014.

\bibitem[Ledig et~al.(2017)Ledig, Theis, Huszar, Caballero, Cunningham, Acosta,
  Aitken, Tejani, Totz, Wang, and Shi]{SRes}
Christian Ledig, Lucas Theis, Ferenc Huszar, Jose Caballero, Andrew Cunningham,
  Alejandro Acosta, Andrew Aitken, Alykhan Tejani, Johannes Totz, Zehan Wang,
  and Wenzhe Shi.
\newblock {Photo-Realistic Single Image Super-Resolution Using a Generative
  Adversarial Network}.
\newblock In \emph{IEEE Conference on Computer Vision and Pattern Recognition
  (CVPR)}, pages 105--114, 2017.

\bibitem[Li et~al.(2017)Li, Chang, Cheng, Yang, and P{\'o}czos]{MMDGAN}
Chun-Liang Li, Wei-Cheng Chang, Yu~Cheng, Yiming Yang, and Barnab{\'a}s
  P{\'o}czos.
\newblock {MMD GAN: Towards deeper understanding of moment matching network}.
\newblock In \emph{Advances in Neural Information Processing Systems}, pages
  2203--2213, 2017.

\bibitem[Liu et~al.(2017)Liu, Bousquet, and Chaudhuri]{T4}
Shuang Liu, Olivier Bousquet, and Kamalika Chaudhuri.
\newblock {Approximation and Convergence Properties of Generative Adversarial
  Learning}.
\newblock In \emph{Advances in Neural Information Processing Systems}, pages
  5545--5553. 2017.

\bibitem[Lopez-Paz and Oquab(2017)]{LR3}
David Lopez-Paz and Maxime Oquab.
\newblock {Revisiting Classifier Two-Sample Tests}.
\newblock \emph{International Conference on Learning Representations (ICLR)},
  2017.

\bibitem[Lucic et~al.(2018)Lucic, Kurach, Michalski, Gelly, and Bousquet]{Q1}
Mario Lucic, Karol Kurach, Marcin Michalski, Sylvain Gelly, and Olivier
  Bousquet.
\newblock {Are GANs Created Equal? A Large-Scale Study}.
\newblock In \emph{Advances in Neural Information Processing Systems}, pages
  698--707. 2018.

\bibitem[Mescheder et~al.(2017)Mescheder, Nowozin, and Geiger]{T1}
Lars Mescheder, Sebastian Nowozin, and Andreas Geiger.
\newblock {The Numerics of GANs}.
\newblock In \emph{Advances in Neural Information Processing Systems}, pages
  1825--1835. 2017.

\bibitem[Metz et~al.(2017)Metz, Poole, Pfau, and Sohl{-}Dickstein]{Metz17}
Luke Metz, Ben Poole, David Pfau, and Jascha Sohl{-}Dickstein.
\newblock {Unrolled Generative Adversarial Networks}.
\newblock In \emph{International Conference on Learning Representations
  (ICLR)}, 2017.

\bibitem[Miyato et~al.(2018)Miyato, Kataoka, Koyama, and Yoshida]{SNDCGAN}
Takeru Miyato, Toshiki Kataoka, Masanori Koyama, and Yuichi Yoshida.
\newblock {Spectral normalization for Generative Adversarial Networks}.
\newblock \emph{International Conference on Learning Representations (ICLR)},
  2018.

\bibitem[Mohamed and Lakshminarayanan(2016)]{G3}
Shakir Mohamed and Balaji Lakshminarayanan.
\newblock {Learning in Implicit Generative Models}.
\newblock \emph{arXiv preprint arXiv:1610.03483}, 2016.

\bibitem[Nowozin et~al.(2016)Nowozin, Cseke, and Tomioka]{D1}
Sebastian Nowozin, Botond Cseke, and Ryota Tomioka.
\newblock {f-GAN: Training Generative Neural Samplers using Variational
  Divergence Minimization}.
\newblock In \emph{Advances in Neural Information Processing Systems}, pages
  271--279. 2016.

\bibitem[Pathak et~al.(2016)Pathak, Krahenbuhl, Donahue, Darrell, and
  Efros]{APaint}
Deepak Pathak, Philipp Krahenbuhl, Jeff Donahue, Trevor Darrell, and Alexei~A.
  Efros.
\newblock {Context Encoders: Feature Learning by Inpainting}.
\newblock In \emph{The IEEE Conference on Computer Vision and Pattern
  Recognition (CVPR)}, June 2016.

\bibitem[Rodriguez et~al.(2018)Rodriguez, Kacprzak, Lucchi, Amara, Sgier,
  Fluri, Hofmann, and R{\'e}fr{\'e}gier]{ADarkMatter}
Andres~C Rodriguez, Tomasz Kacprzak, Aurelien Lucchi, Adam Amara, Raphael
  Sgier, Janis Fluri, Thomas Hofmann, and Alexandre R{\'e}fr{\'e}gier.
\newblock {Fast Cosmic Web Simulations with Generative Adversarial Networks}.
\newblock \emph{arXiv preprint arXiv:1801.09070}, 2018.

\bibitem[Sajjadi et~al.(2018)Sajjadi, Bachem, Lucic, Bousquet, and Gelly]{PRD}
Mehdi S.~M. Sajjadi, Olivier Bachem, Mario Lucic, Olivier Bousquet, and Sylvain
  Gelly.
\newblock {Assessing Generative Models via Precision and Recall}.
\newblock In \emph{Advances in Neural Information Processing Systems}, pages
  5234--5243. 2018.

\bibitem[Salimans et~al.(2016)Salimans, Goodfellow, Zaremba, Cheung, Radford,
  Chen, and Chen]{IS}
Tim Salimans, Ian Goodfellow, Wojciech Zaremba, Vicki Cheung, Alec Radford,
  Xi~Chen, and Xi~Chen.
\newblock {Improved Techniques for Training GANs}.
\newblock In \emph{Advances in Neural Information Processing Systems}, pages
  2234--2242. 2016.

\bibitem[Szegedy et~al.(2015)Szegedy, Liu, Jia, Sermanet, Reed, Anguelov,
  Erhan, Vanhoucke, and Rabinovich]{AdvSam2}
Christian Szegedy, Wei Liu, Yangqing Jia, Pierre Sermanet, Scott Reed, Dragomir
  Anguelov, Dumitru Erhan, Vincent Vanhoucke, and Andrew Rabinovich.
\newblock Going deeper with convolutions.
\newblock In \emph{Proceedings of the IEEE conference on computer vision and
  pattern recognition}, pages 1--9, 2015.

\bibitem[Szegedy et~al.(2017)Szegedy, Ioffe, and Vanhoucke]{inceptionv4}
Christian Szegedy, Sergey Ioffe, and Vincent Vanhoucke.
\newblock {Inception-v4, Inception-ResNet and the Impact of Residual
  Connections on Learning}.
\newblock In \emph{AAAI}, 2017.

\bibitem[Theis et~al.(2016)Theis, Oord, and Bethge]{EGM1}
Lucas Theis, A{\"a}ron van~den Oord, and Matthias Bethge.
\newblock A note on the evaluation of generative models.
\newblock \emph{International Conference on Learning Representations (ICLR)},
  2016.

\bibitem[Tolstikhin et~al.(2017)Tolstikhin, Gelly, Bousquet, Simon-Gabriel, and
  Sch\"{o}lkopf]{T2}
Ilya~O Tolstikhin, Sylvain Gelly, Olivier Bousquet, Carl-Johann Simon-Gabriel,
  and Bernhard Sch\"{o}lkopf.
\newblock {AdaGAN: Boosting Generative Models}.
\newblock In \emph{Advances in Neural Information Processing Systems}, pages
  5424--5433. 2017.

\bibitem[Wu et~al.(2017)Wu, Burda, Salakhutdinov, and Grosse]{LR2}
Yuhuai Wu, Yuri Burda, Ruslan Salakhutdinov, and Roger Grosse.
\newblock {On the Quantitative analysis of Decoder-Based Generative Models}.
\newblock \emph{International Conference on Learning Representations (ICLR)},
  2017.

\bibitem[Zhang et~al.(2017)Zhang, Xu, Li, Zhang, Wang, Huang, and
  Metaxas]{StackGAN}
Han Zhang, Tao Xu, Hongsheng Li, Shaoting Zhang, Xiaogang Wang, Xiaolei Huang,
  and Dimitris~N. Metaxas.
\newblock {StackGAN: Text to Photo-Realistic Image Synthesis With Stacked
  Generative Adversarial Networks}.
\newblock In \emph{The IEEE International Conference on Computer Vision
  (ICCV)}, 2017.

\bibitem[Zhu et~al.(2016)Zhu, Kr{\"a}henb{\"u}hl, Shechtman, and
  Efros]{zhu2016}
Jun-Yan Zhu, Philipp Kr{\"a}henb{\"u}hl, Eli Shechtman, and Alexei~A. Efros.
\newblock {Generative Visual Manipulation on the Natural Image Manifold}.
\newblock In \emph{Proceedings of European Conference on Computer Vision
  (ECCV)}, pages 597--613, 2016.

\bibitem[Zhu et~al.(2017)Zhu, Park, Isola, and Efros]{ACrossDom2}
Jun-Yan Zhu, Taesung Park, Phillip Isola, and Alexei~A Efros.
\newblock {Unpaired Image-to-Image Translation using Cycle-Consistent
  Adversarial Networks}.
\newblock In \emph{The IEEE International Conference on Computer Vision
  (ICCV)}, 2017.

\end{thebibliography}

\appendix

\section*{Appendix A: Architecture of GANs}
\label{app:GAN_architecture}

The structural parameters of both the discriminator and generator networks used to train the different GANs in our study (WGAN, WGAN-GP and SN-DCGAN) are as follows.

\textbf{SNDCGAN}: The discriminator is a  7 layer deep CNN with [64, 128, 128, 256, 256, 512, 512] filters each followed by a fully connected layer. We use Leaky ReLU as activation function of the intermediate layers. The generator starts with a fully connected layer followed by 4 deconvolutional layers with depths [512, 256, 128, 64]. We use batch normalization between the hidden layers and  ReLU as activation function. This model is trained with the Adam optimizer with learning rate of $0.0001$ and parameters $\beta_1=0.5$ and $\beta_2=0.999$.

\textbf{WGANGP}: For the discriminator, we use a CNN with 4  layers with [64, 128, 256, 512] filters each for CelebA and 3 layers with depths [128, 256, 512] for CIFAR10, followed by a single fully connected layer in both cases. We use Leaky ReLU as the activation function of the hidden layers. The generator starts with a fully connected layer and continues with  a 4 layers CNN for CelebA  and a 3 layer CNN for CIFAR10 with depths [512, 256, 128, 64] and [512, 256, 128] respectively. We use batch normalization in the hidden layers and ReLU as the activation function. We have used the Adam optimizer with learning rate of $0.0001$ and parameters $\beta_1=0.5$ and $\beta_2=0.9$.

\textbf{WGAN}: The discriminator is a 4 layer CNN with depths [32, 64, 128, 256]  followed by a fully connected layer. All convolutional layers use Leaky ReLU as activation function and  batch normalization.The generator contains a fully connected layer followed by 4 convolutional layers with depths [256, 128, 64, 32]. We use batch normalization between the hidden layers and Leaky ReLU activation function. For training we use the RMSProp optimizer  with learning rate of $0.0001$.

\begin{figure}[h!]
\begin{center}
	\subfigure[WGANGP]{\includegraphics[width=5cm]{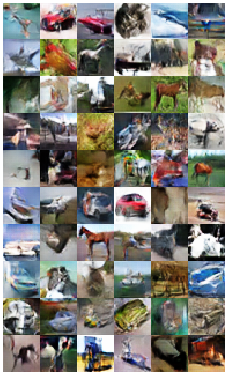}}%
	\subfigure[WGAN]{\includegraphics[width=5cm]{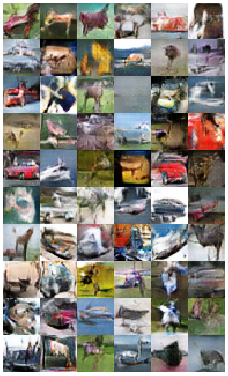}}%
	\subfigure[SNDCGAN]{\includegraphics[width=5cm]{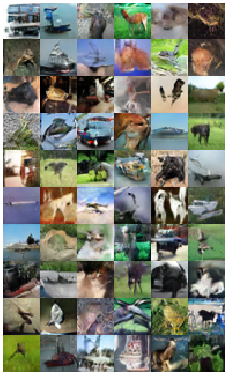}}%
	\\
	\subfigure[WGAN-GP]{\includegraphics[width=5cm]{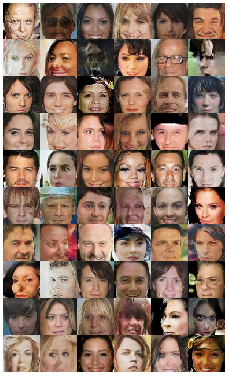}}%
	\subfigure[SNDCGAN]{\includegraphics[width=5cm]{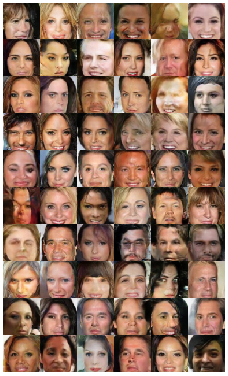}}%
	\end{center}
	\caption{Samples drawn from WGAN, WGAN-GP and SN-DCGAN when trained over CIFAR and CelebA dataset with $\text{dim}(\z)=256$.}%
	\label{fig:gener_imgs}
\end{figure}

In Figure \ref{fig:gener_imgs} we show samples of the three GANs when trained over CIFAR10 and CelebA dataset with $\text{dim}(\z)=256$.

\section*{Appendix B:  Data Reconstruction}
\label{app:data_recons}

\begin{figure*}[h!]
\hspace{-2.5cm}{
	\subfigure[WGANGP CIFAR10]{\includegraphics[width=4cm]{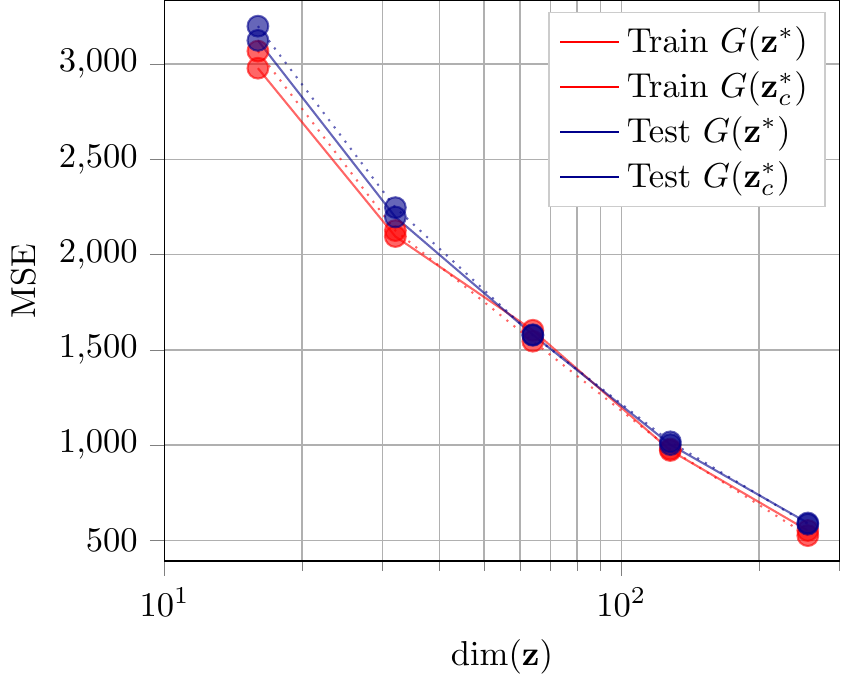}}%
	\subfigure[WGAN CIFAR10]{\includegraphics[width=4cm]{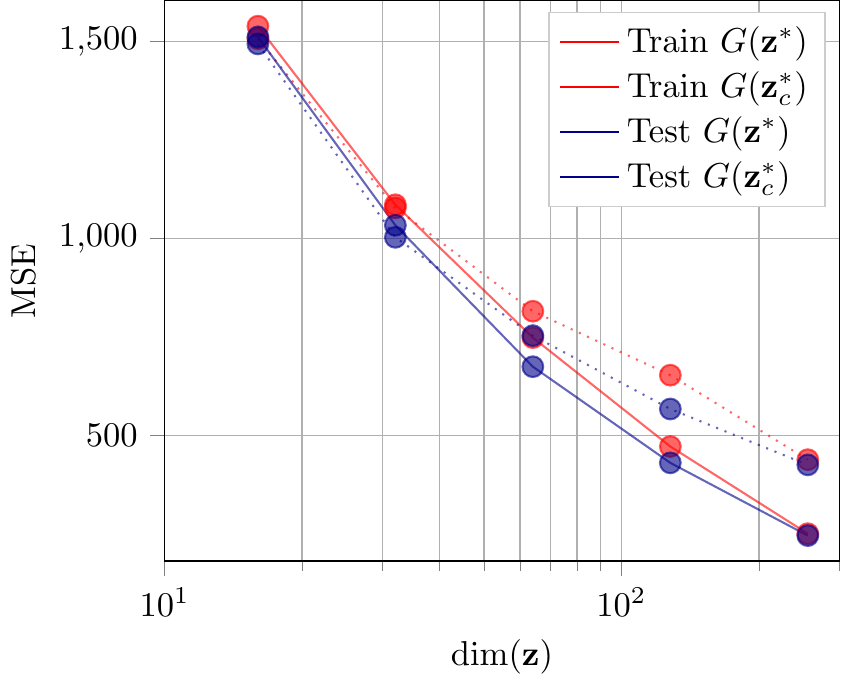}}%
	\subfigure[SN-DCGAN C10]{\includegraphics[width=4cm]{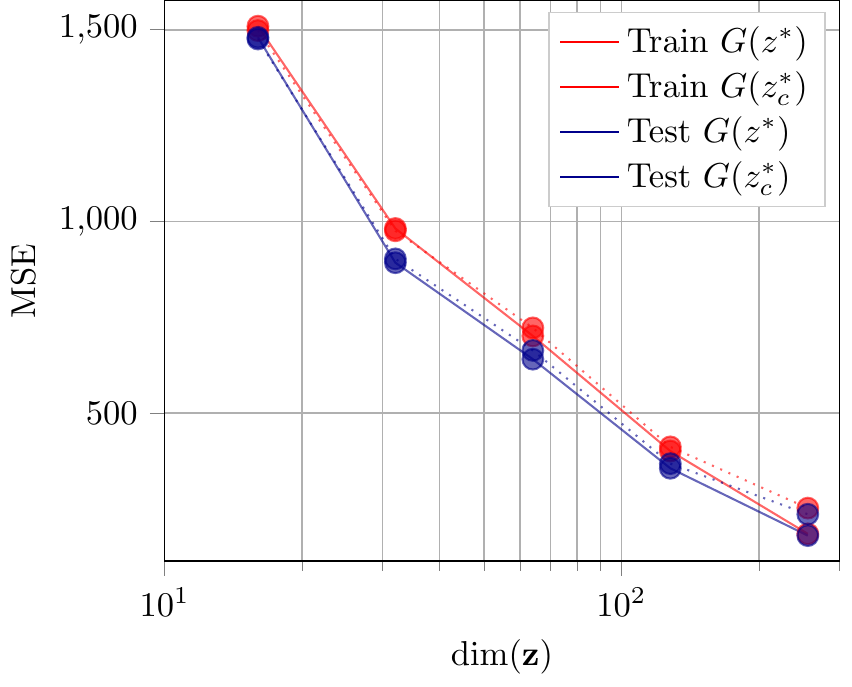}}%
	\subfigure[WGAN-GP celebA]{\includegraphics[width=4cm]{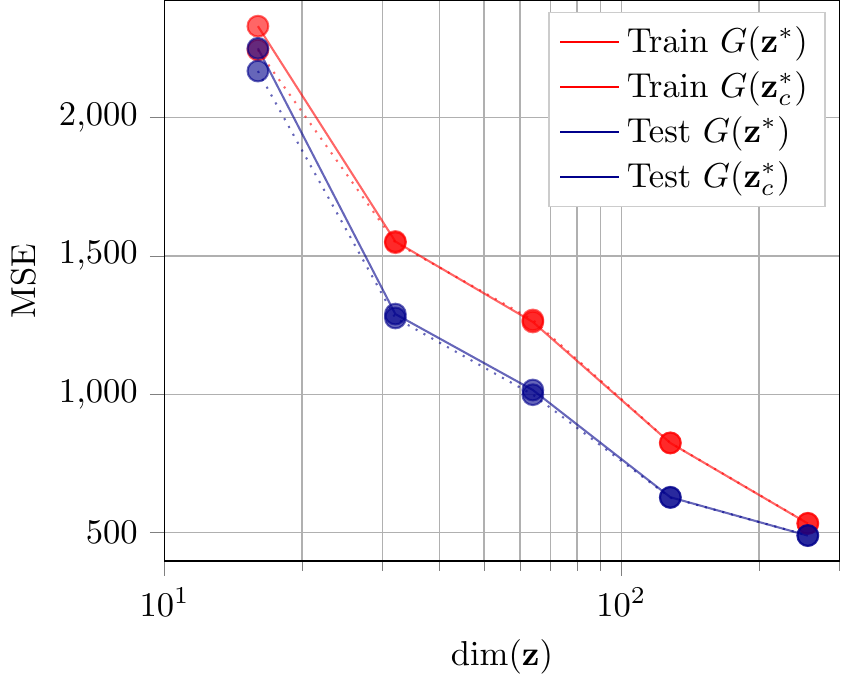}}%
	\subfigure[SN-DCGAN celebA]{\includegraphics[width=4cm]{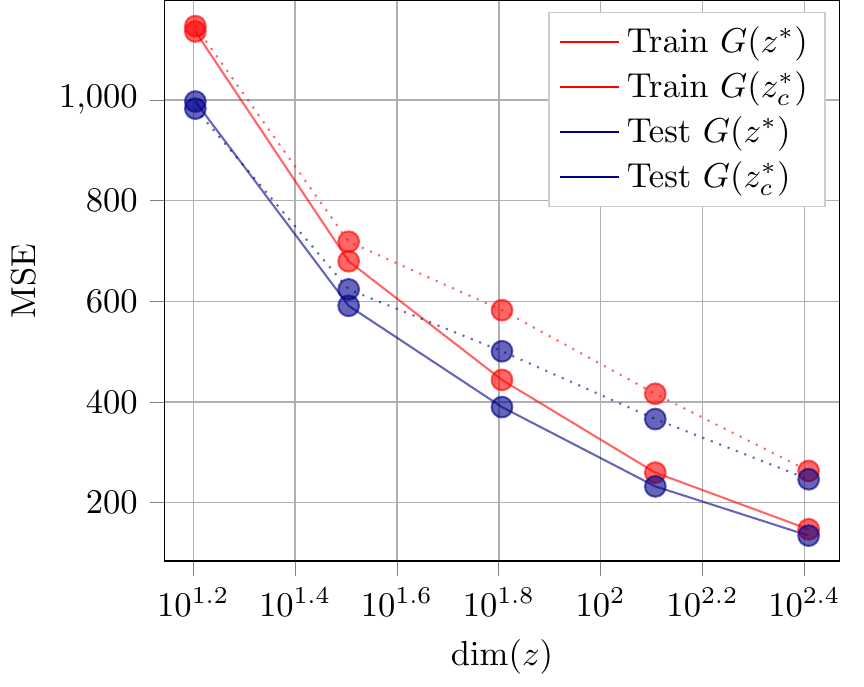}}%
	}
	\\
	
	\hspace{-2.5cm}{
	\subfigure[WGANGP CIFAR10]{\includegraphics[width=4cm]{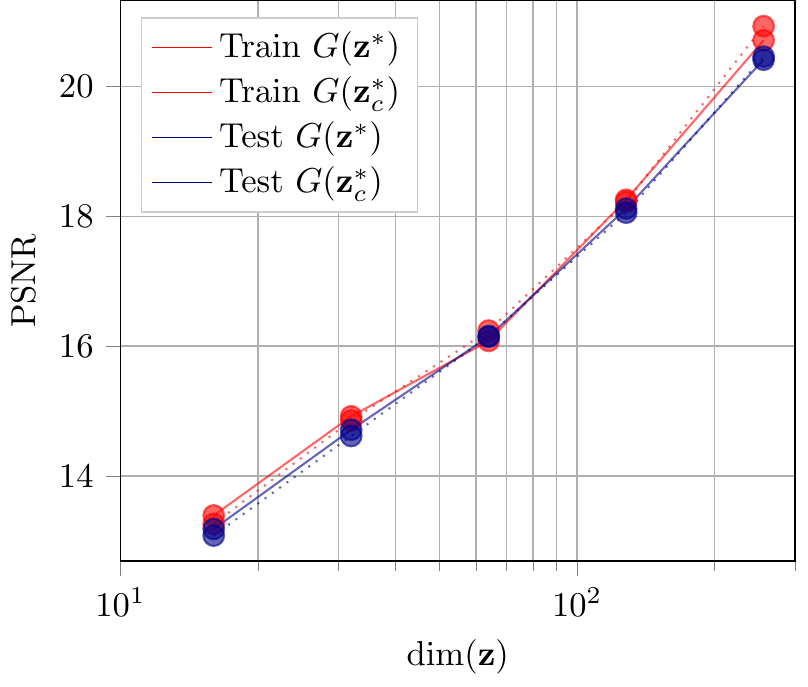}}%
	\subfigure[WGAN CIFAR10]{\includegraphics[width=4cm]{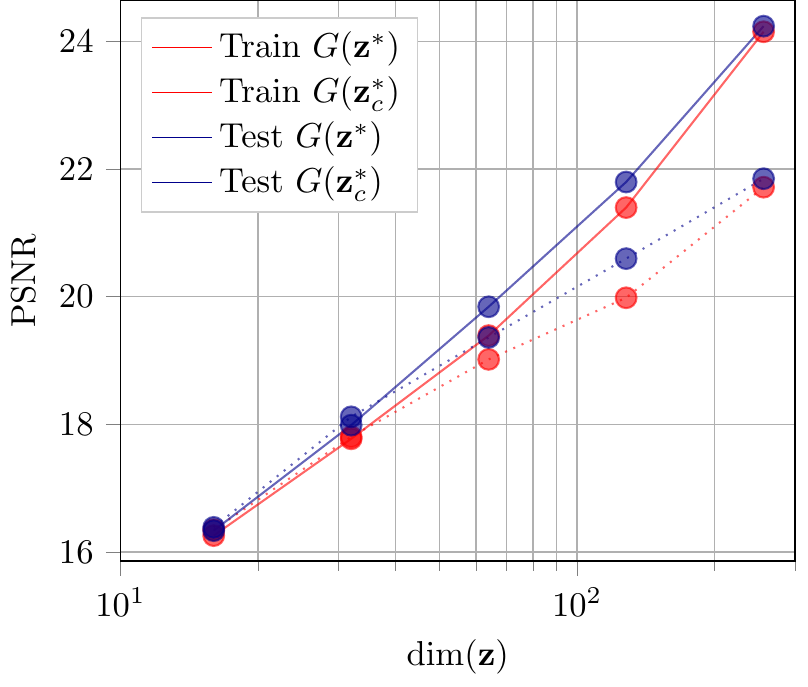}}%
	\subfigure[SN-DCGAN C10]{\includegraphics[width=4cm]{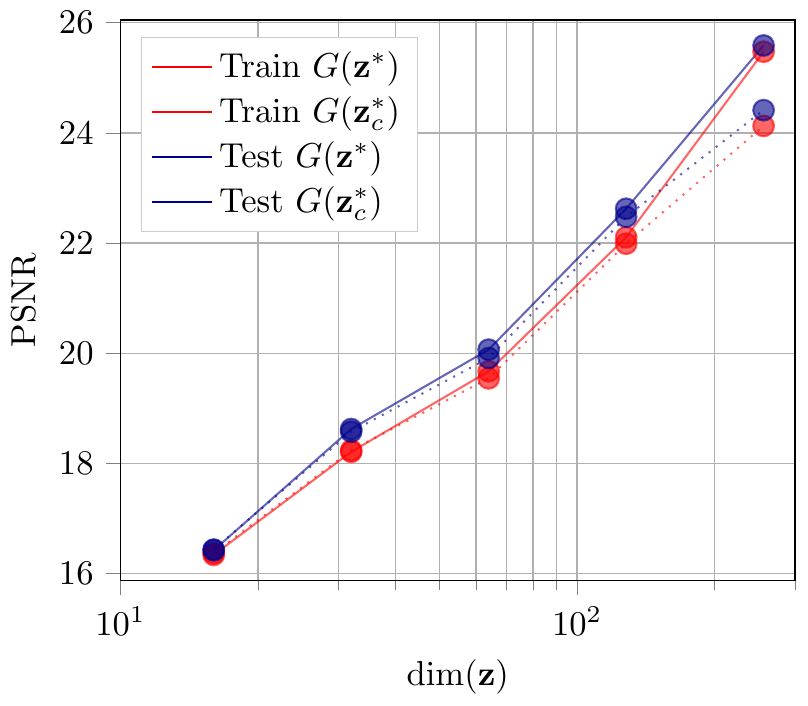}}%
	\subfigure[WGAN-GP celebA]{\includegraphics[width=4cm]{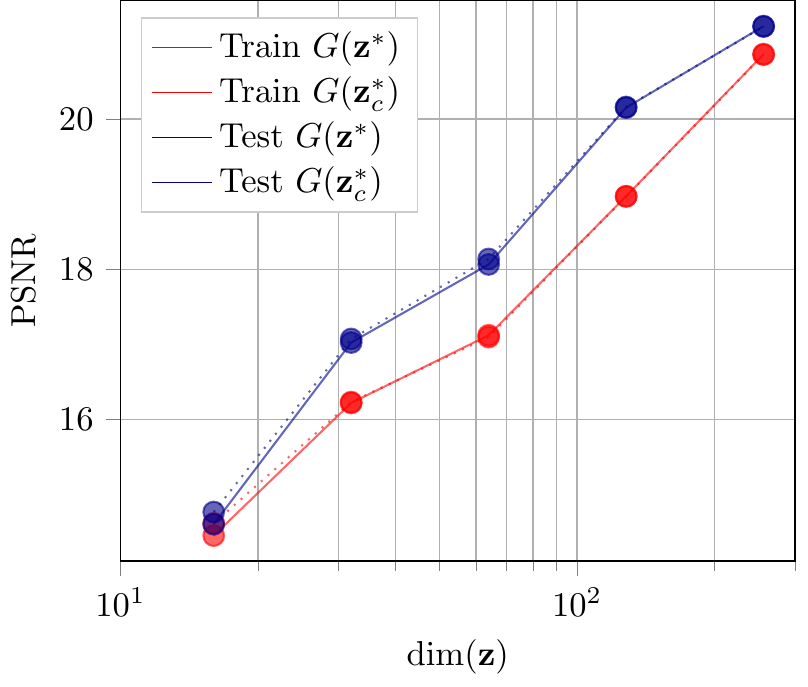}}%
	\subfigure[SN-DCGAN celebA]{\includegraphics[width=4cm]{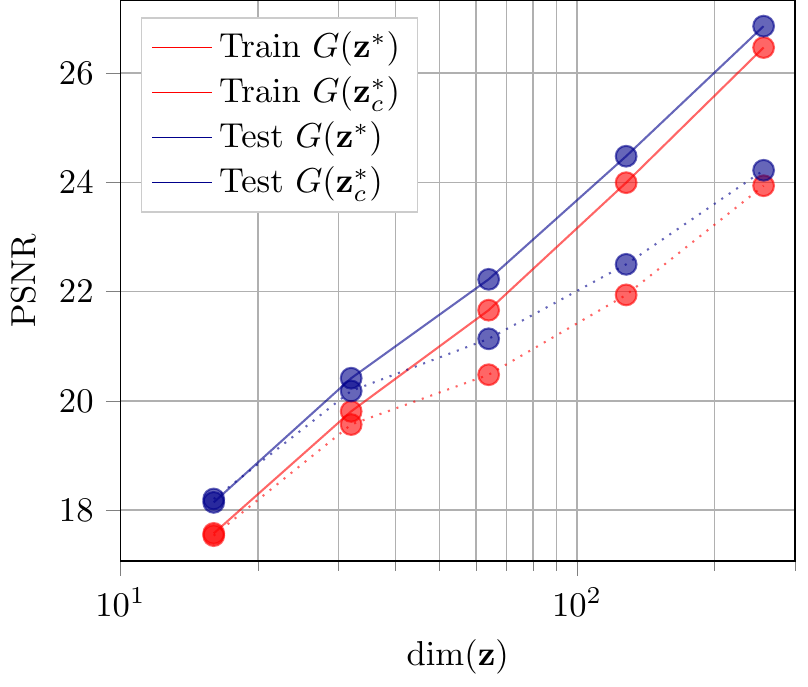}}%
	
	}
	\caption{Average MSE and PSRN between real test/training images and their reconstruction using $\z^*$ in (1) or $\z_c^*$ in (2), as $\text{dim}(\z)$ grows.}%
	\label{fig:MSE_dim_app}
\end{figure*}

\begin{figure}[h!]
\hspace{-2.5cm}{
	\subfigure[WGANGP CIFAR10]{\includegraphics[width=4cm]{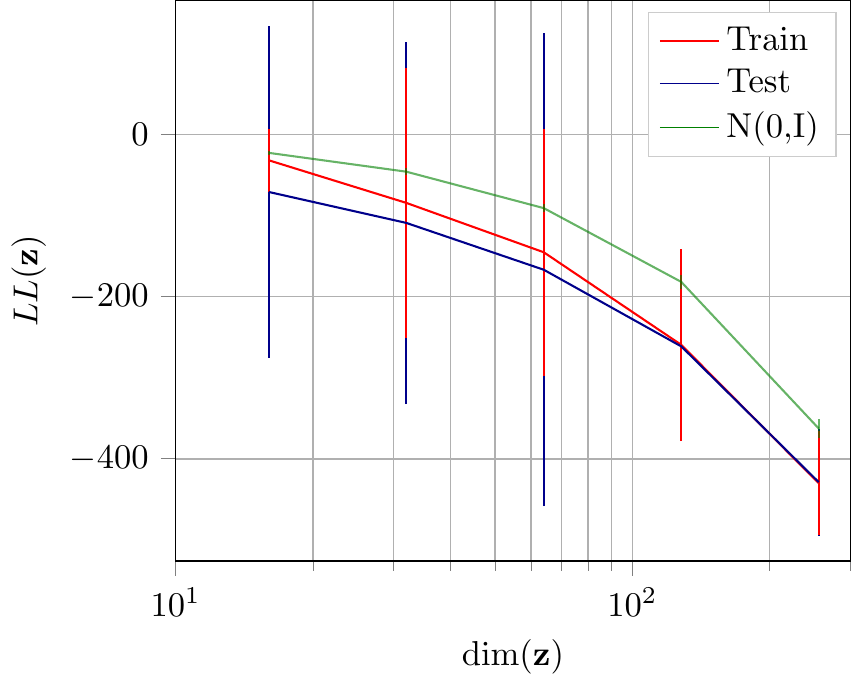}}%
	\subfigure[WGAN CIFAR10]{\includegraphics[width=4cm]{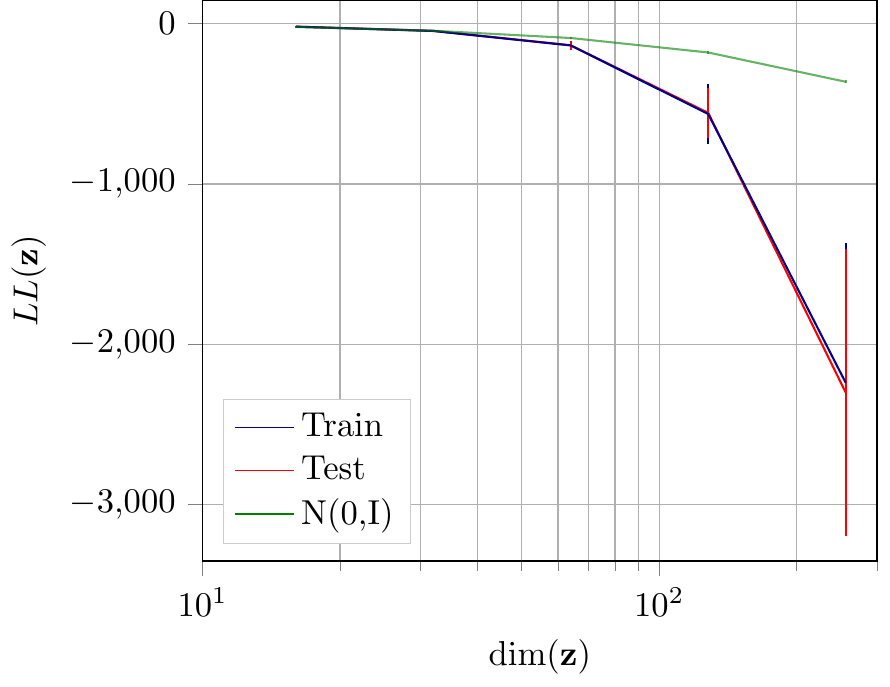}}%
	\subfigure[SN-DCGAN C10]{\includegraphics[width=4cm]{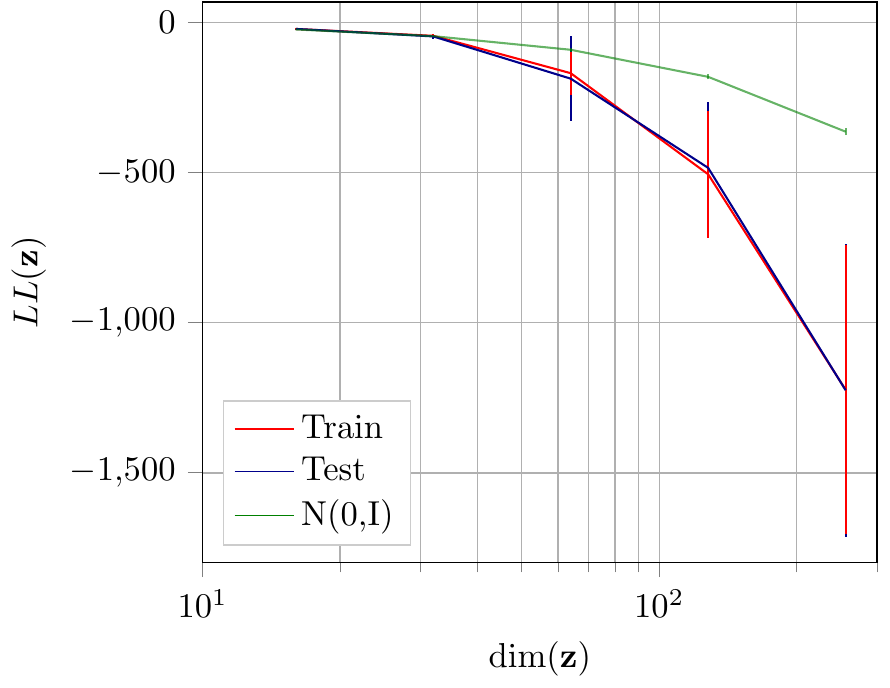}}%
	\subfigure[WGAN-GP celebA]{\includegraphics[width=4cm]{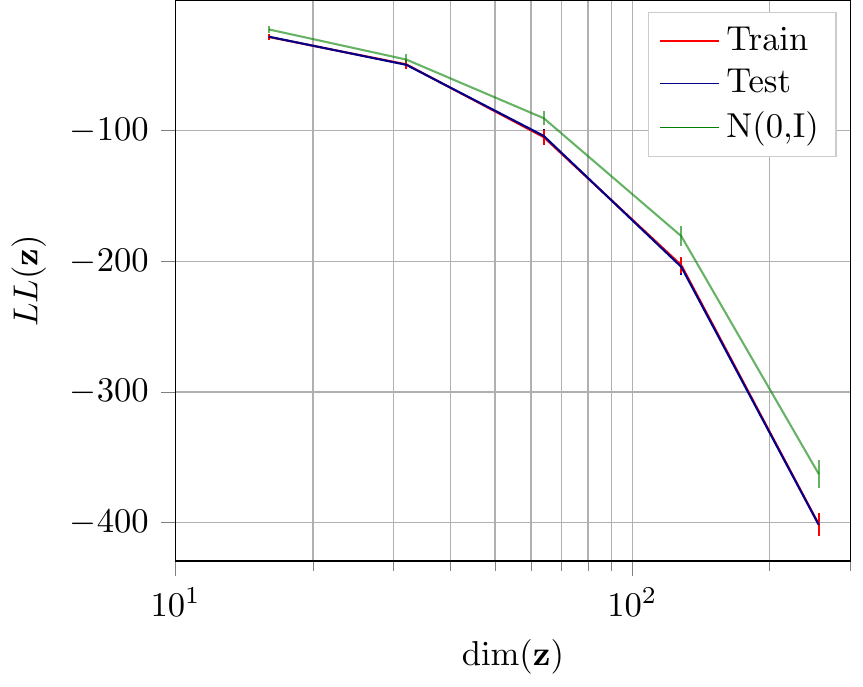}}%
	\subfigure[SN-DCGAN celebA]{\includegraphics[width=4cm]{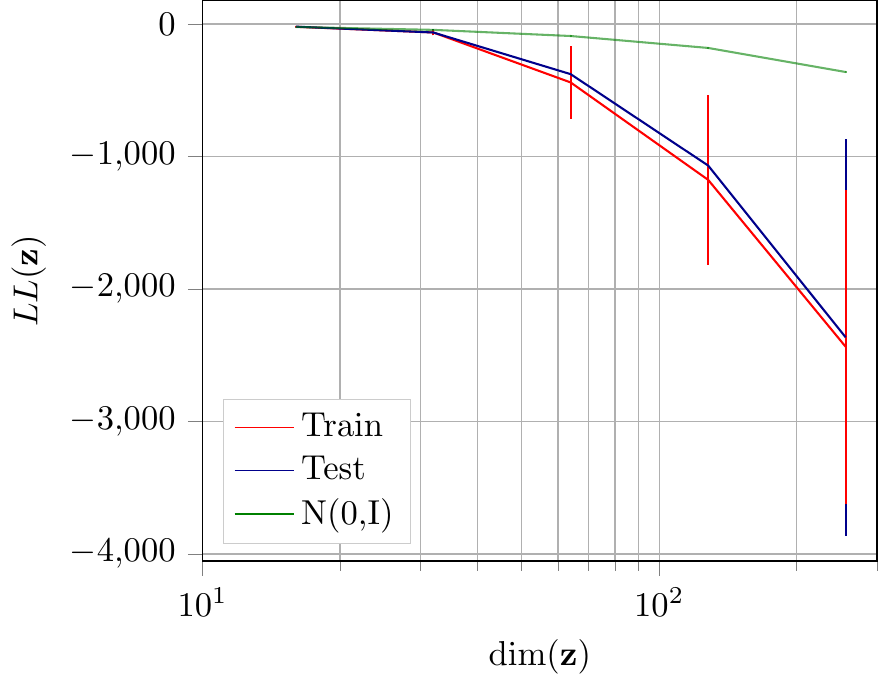}}%
	}
	\\
	
	\hspace{-2.5cm}{
	\subfigure[WGANGP CIFAR10]{\includegraphics[width=4cm]{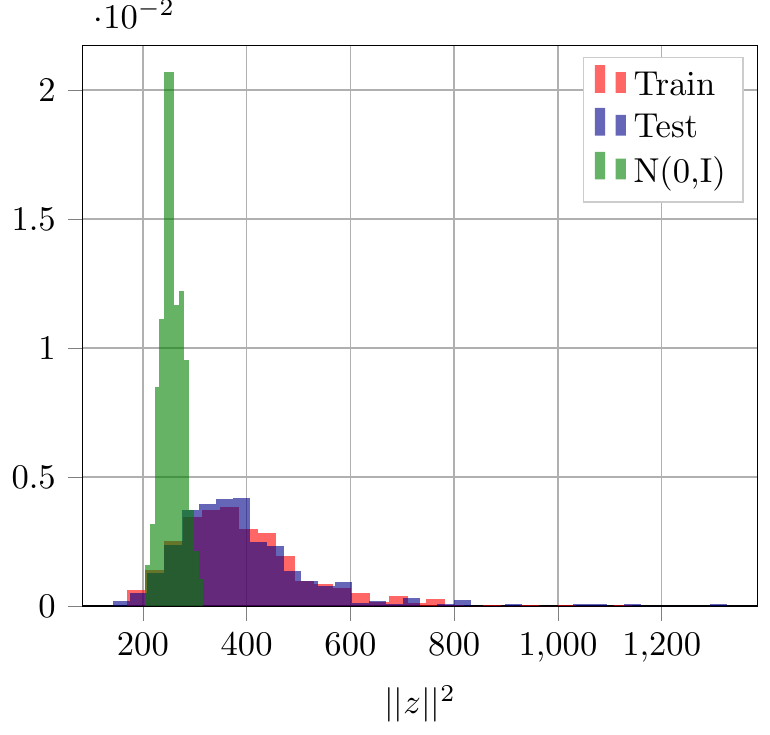}}%
	\subfigure[WGAN CIFAR10]{\includegraphics[width=4cm]{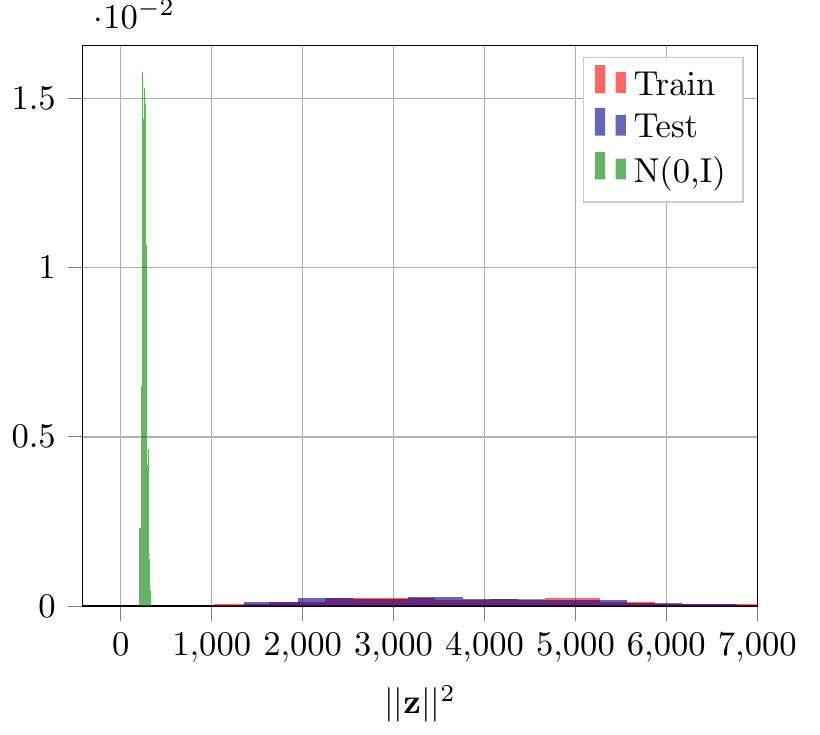}}%
	\subfigure[SN-DCGAN C10]{\includegraphics[width=4cm]{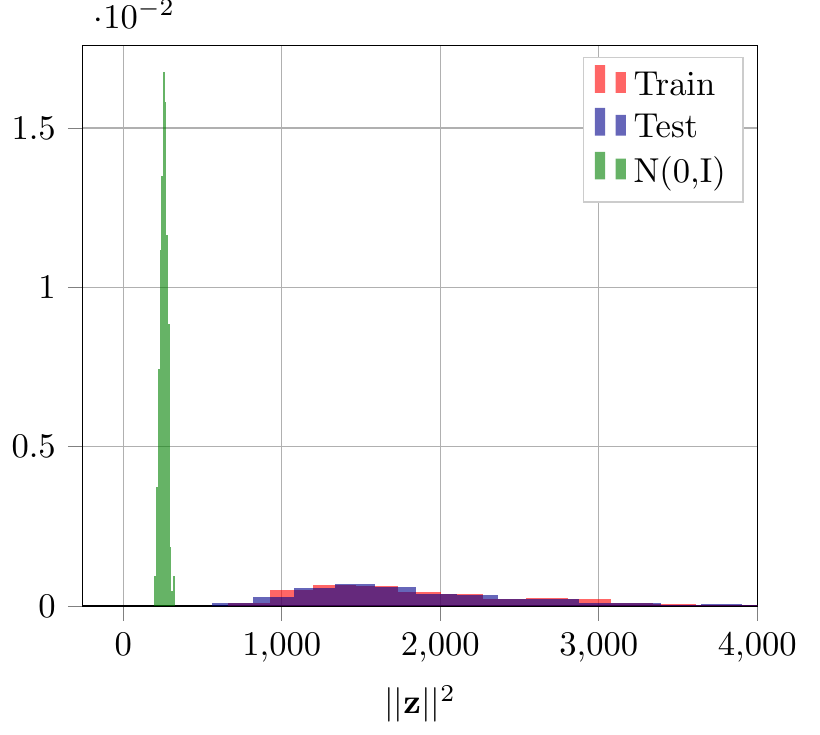}}%
	\subfigure[WGAN-GP celebA]{\includegraphics[width=4cm]{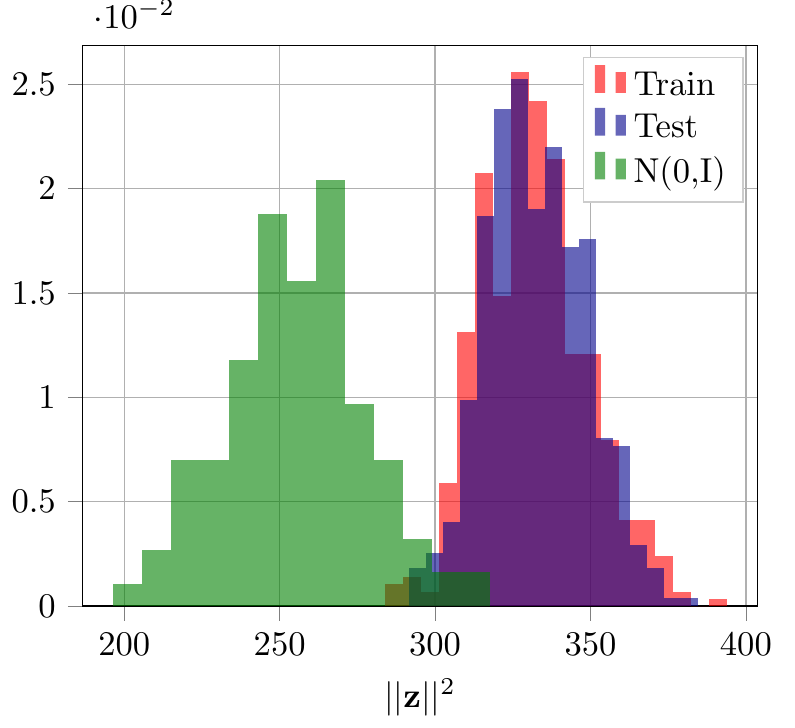}}%
	\subfigure[SN-DCGAN celebA]{\includegraphics[width=4cm]{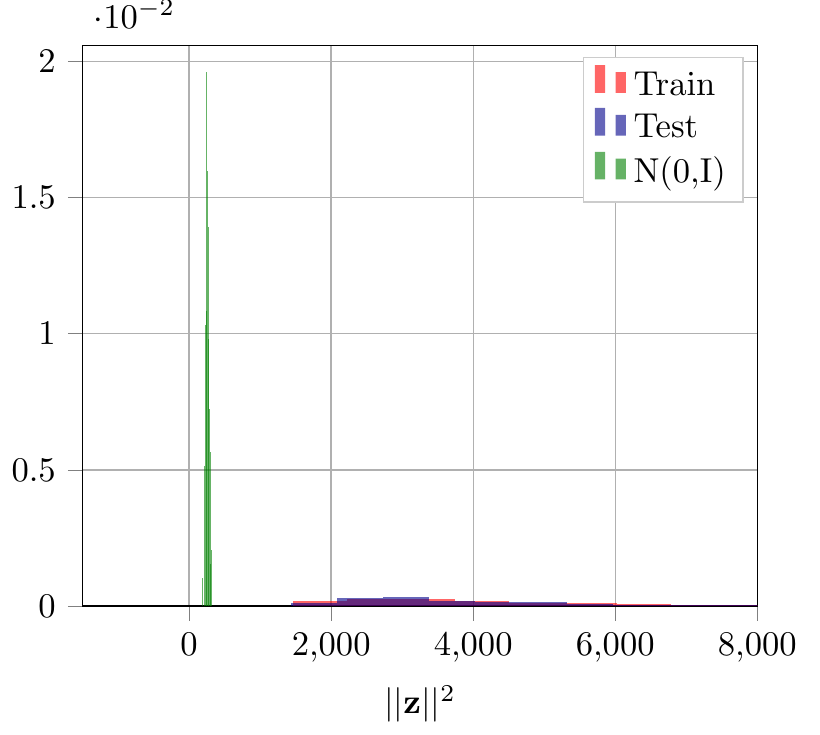}}%
	}
	\caption{In the top row we show the average log-likelihood $LL(\z^*)$ as a function of $\text{dim}(\z)$. In the bottom row, we show the histogram of $||\z^*||^2$ for $\text{dim}(\z)=256$.}%
	\label{fig:LL_dim_app}
\end{figure}

Figure \ref{fig:MSE_dim_app} shows the average MSE between real test/training images and their reconstruction using $\z^*$ in (1) or $\z_c^*$ in (2), as $\text{dim}(\z)$ grows. SN-DCGAN stands out in terms of reconstruction error, achieving PSNR values above 26 dB for $\text{dim}(\z)=256$. In the top row of Figure \ref{fig:LL_dim_app} we show the average log-likelihood $LL(\z^*)$ as a function of $\text{dim}(\z)$.  For high dimensions, in all cases it is significantly smaller than the typical LL of samples from the input distribution $p(\z)$, indicating that the sampling from the input distribution so that the best reconstructed image is obtained is extremely unlikely. In the bottom row, we show the histogram of $||\z^*||^2$ for $\text{dim}(\z)=256$.

\begin{figure}[hb!]
\hspace{-2.4cm}{
	\subfigure[WGANGP CIFAR10]{\includegraphics[width=4cm]{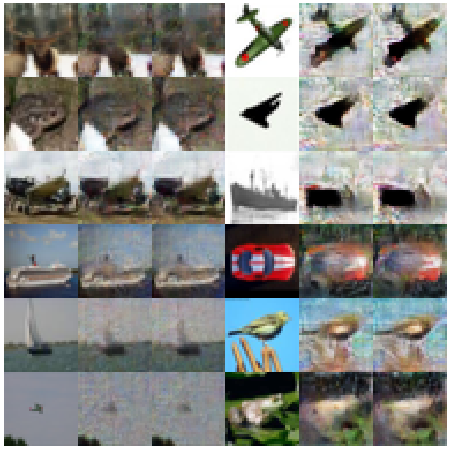}}%
	\subfigure[WGAN CIFAR10]{\includegraphics[width=4cm]{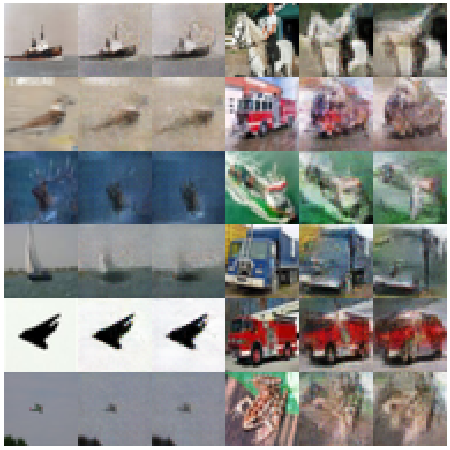}}%
	\subfigure[SNDCGAN CIFAR10]{\includegraphics[width=4cm]{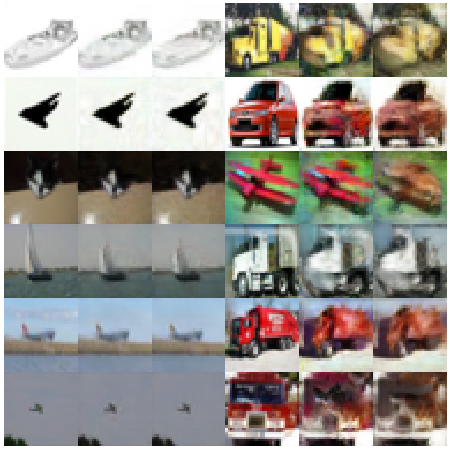}}%
	\subfigure[WGAN-GP celebA]{\includegraphics[width=4cm]{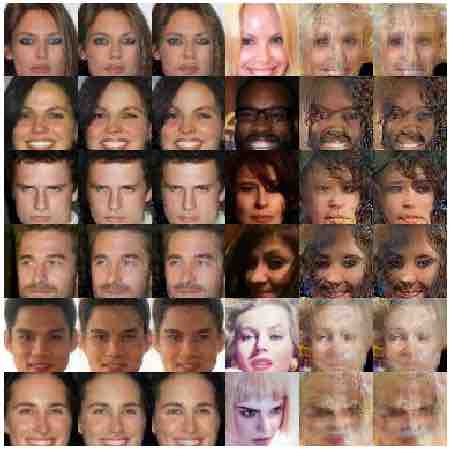}}%
	\subfigure[SNDCGAN celebA]{\includegraphics[width=4cm]{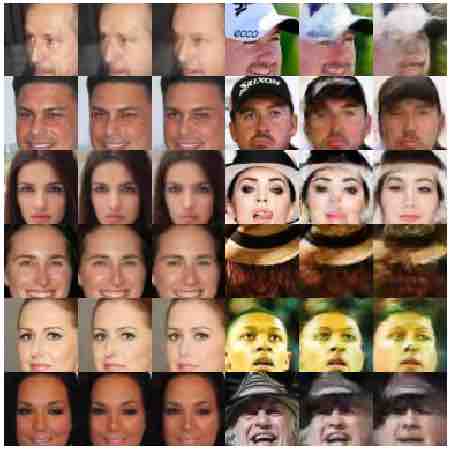}}
}
	\\
	
	\hspace{-2.4cm}{
	\subfigure[WGANGP CIFAR10]{\includegraphics[width=4cm]{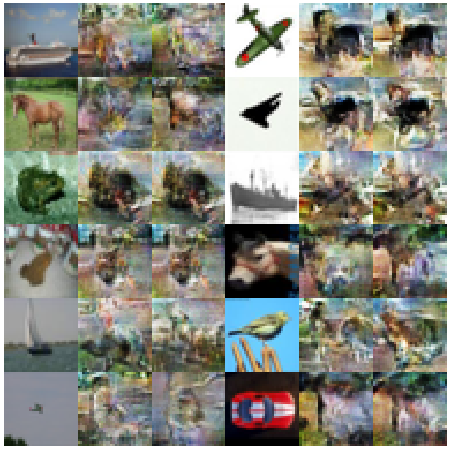}}%
	\subfigure[WGAN CIFAR10]{\includegraphics[width=4cm]{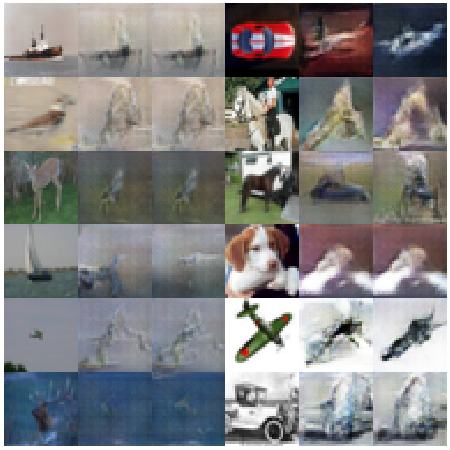}}%
	\subfigure[SNDCGAN CIFAR10]{\includegraphics[width=4cm]{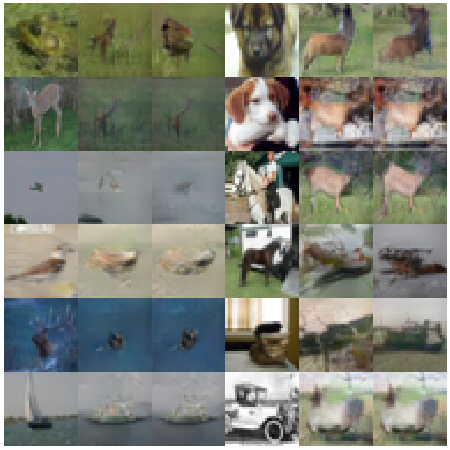}}%
	\subfigure[WGAN-GP celebA]{\includegraphics[width=4cm]{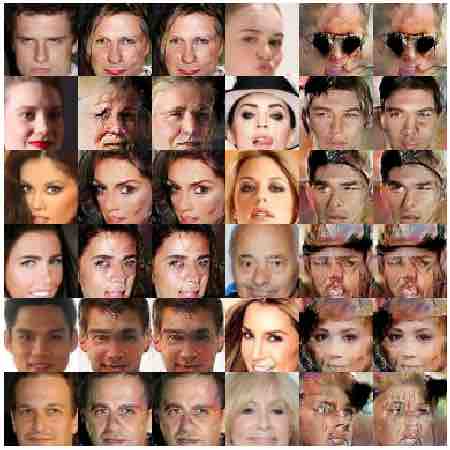}}%
	\subfigure[SNDCGAN celebA]{\includegraphics[width=4cm]{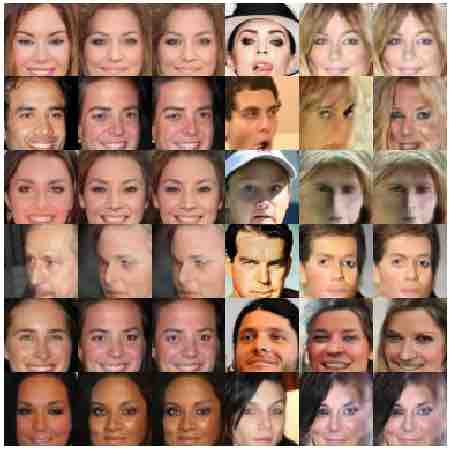}}%
}
	\caption{We compare test images (first column) with $G(\z^*)$ (central column) and $G(\z^*)$ (right column). The left group of images represent the test samples with largest $\text{PSNR}(\x, G(\z^*_c))$  while the right group contains the samples with the worst PSNR values. The top row corresponds to $\text{dim}(\z)=256$, and the bottom row to $\text{dim}(\z)=16$.}%
	\label{fig:PSNR_256}
\end{figure}

In Figure \ref{fig:PSNR_256} we compare test images (first column) with $G(\z^*)$ (central column) and $G(\z^*)$ (right column). The left group of images represents the test samples with largest $\text{PSNR}(\x, G(\z^*_c))$  while the right group contains the samples with the worst PSNR values. The top row corresponds to $\text{dim}(\z)=256$, and the bottom row to $\text{dim}(\z)=16$. While for $\text{dim}(\z)=16$ the reconstruction error is in general large for all images, for the high quality reconstructions in the case $\text{dim}(\z)=256$ there is little difference between the constraint and unconstraint optimizations, while for the lower quality reconstructions the differences are quite significant.

\begin{figure*}[h!]
	\centering
	\subfigure[WGANGP CIFAR10]{\includegraphics[width=6.5cm]{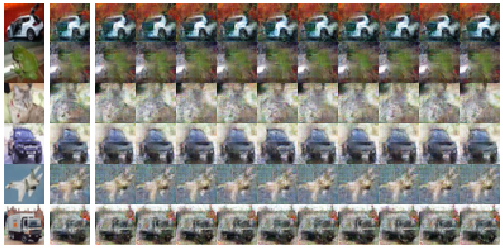}}\\%
	\subfigure[WGAN CIFAR10]{~~\includegraphics[width=6.5cm]{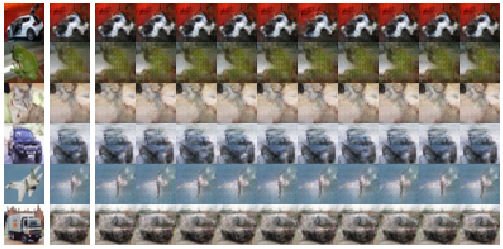}}%
	\subfigure[SNDCGAN CIFAR10]{~~\includegraphics[width=6.5cm]{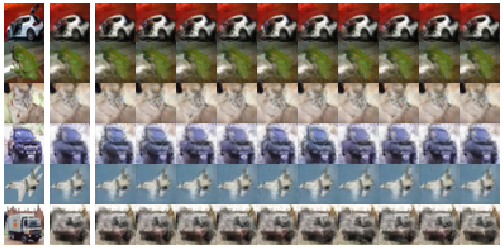}}\\%
	\subfigure[WGAN-GP celebA]{\includegraphics[width=6.5cm]{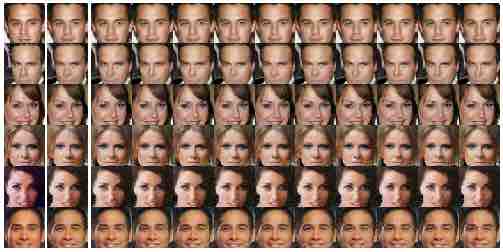}}%
	\subfigure[SNDCGAN celebA]{~~\includegraphics[width=6.5cm]{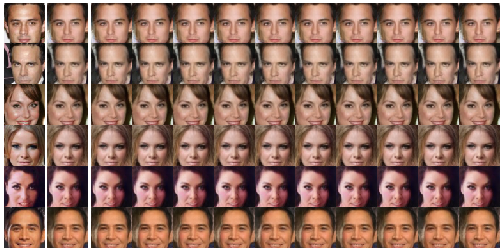}}%
	\caption{Reconstructed image $G(\z_c^*)$ for 5 different test images using 10 different initalizations. Left most column is the original image. In the second column we also show the reconstruction mean input noise sample, i.e.  $\z^*_{c,p}=\sum_m \z^*_{c,m}/10$, where $ \z^*_{c,m}$, $m=1,\ldots,10$ are each one of the 10 solutions.}%
	\label{fig:SIM_recons}
\end{figure*}

\begin{figure*}[hb!]
	\centering
	\subfigure[WGANGP CIFAR10]{\includegraphics[width=6.5cm]{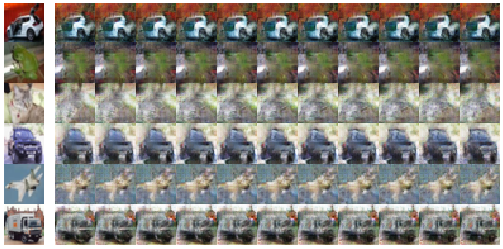}}
	\subfigure[WGAN CIFAR10]{~~\includegraphics[width=6.5cm]{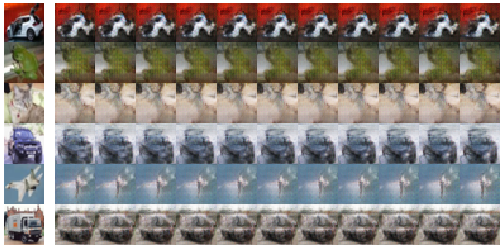}}%
	\subfigure[SNDCGAN CIFAR10]{~~\includegraphics[width=6.5cm]{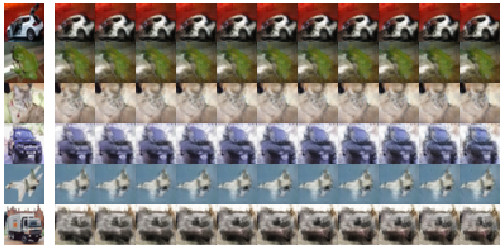}}\\%
	\subfigure[WGAN-GP celebA]{\includegraphics[width=6.5cm]{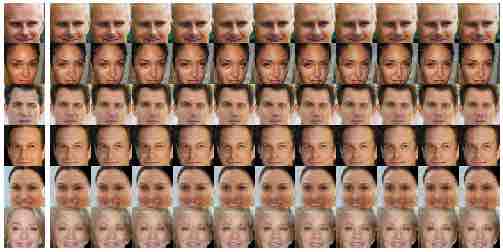}}%
	\subfigure[SNDCGAN celebA]{~~\includegraphics[width=6.5cm]{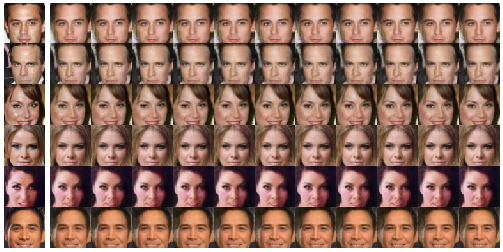}}%
	\caption{Reconstruction from linearly interpolated noise samples using the two noise samples $\z^*_{c,m}$ that are further apart among those found for 10 different initalizations of the constrained problem in (2). The left most column is the original image.}%
	\label{fig:SIM_z1_z2}
\end{figure*}

In Figure \ref{fig:SIM_recons} we show the reconstructed image $G(\z_c^*)$ for 5 different test images using 10 different initializations. We also show the reconstruction mean input noise sample, i.e.  $\z^*_{c,p}=\sum_m \z^*_{c,m}/10$, where $ \z^*_{c,m}$, $m=1,\ldots,10$ are each one of the 10 solutions. In Figure \ref{fig:SIM_z1_z2} we also took the two $\z^*_{c,m}$ that were further apart and linearly interpolate their values to generate the images in between. These images are shown in  with similar behavior as the experiment in Figure \ref{fig:SIM_recons}. Similar conclusions can be drawn when we perform polar interpolation instead of linear interpolation. In short, even if the optimization problems are not convex and uni-modality is not enforced by GAN training, we did not find issues with either. 

\section*{Appendix C:  EvalGAN sample marginal likelihood}
\label{app:data_logpx}

\begin{figure*}[hb!]
\hspace{-2.5cm}{
	\subfigure[WGANGP CIFAR10]{\includegraphics[width=4cm]{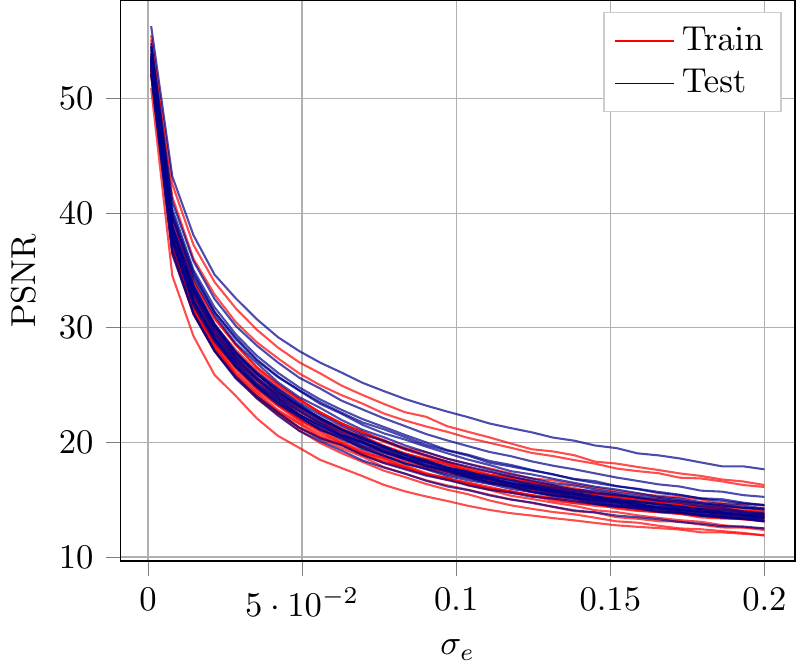}}%
	\subfigure[WGAN CIFAR10]{\includegraphics[width=4cm]{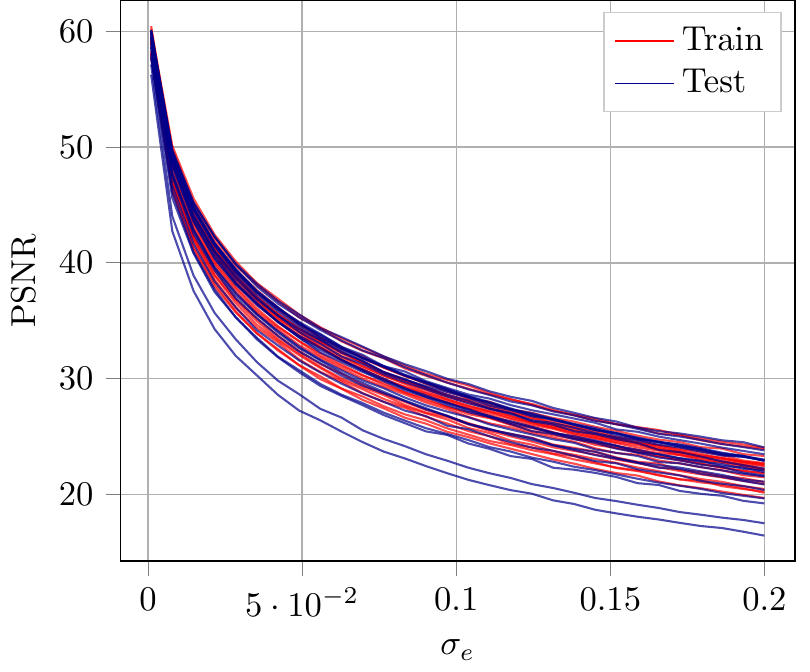}}%
	\subfigure[SNDCGAN CIFAR10]{\includegraphics[width=4cm]{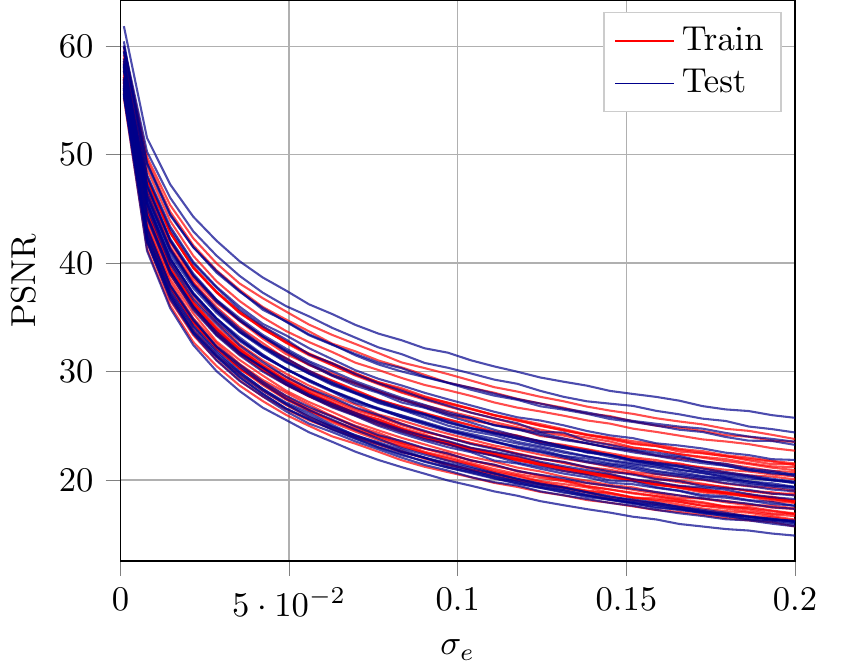}}%
	\subfigure[WGAN-GP celebA]{\includegraphics[width=4cm]{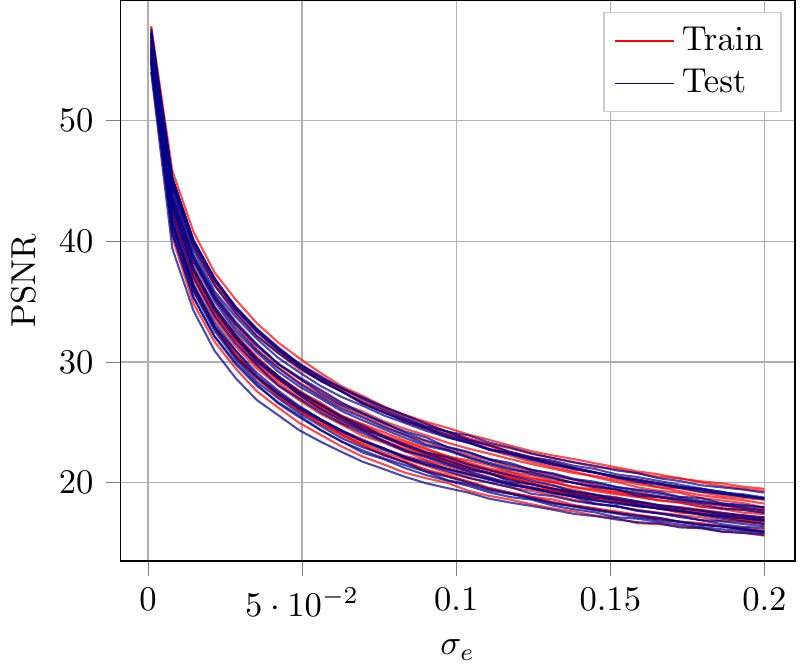}}%
	\subfigure[SNDCGAN celebA]{\includegraphics[width=4cm]{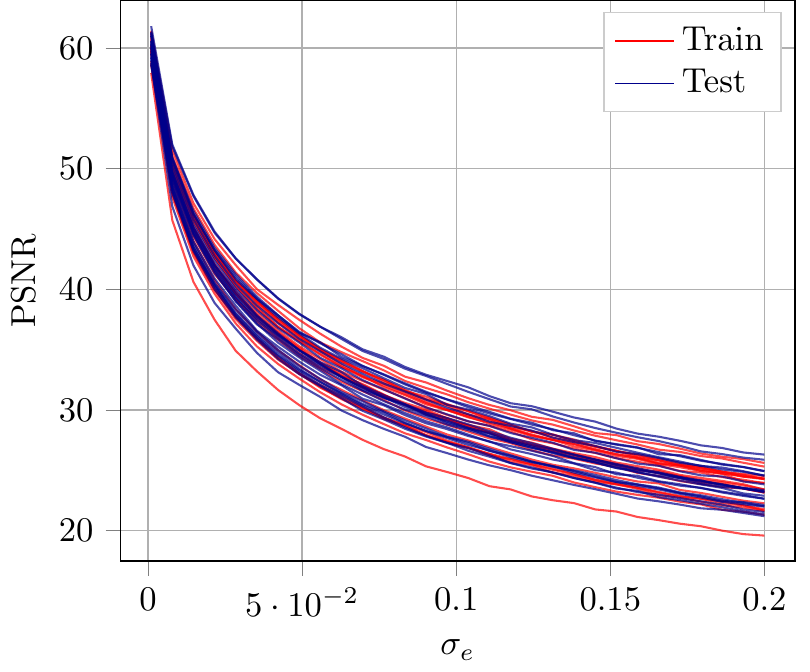}}%
}
	\caption{Evolution of the average PSNR between $G(\z_c^*)$ and $G(\z_c^*+\epsilon_i)$  as $\sigma_\epsilon$ grows for 20 test and 20 train images, where $\epsilon_i\sim\mathcal{N}(\mathbf{0},\sigma_\epsilon^2\mathbf{I})$}.%
	\label{fig:z_region_PSNR_norm_some}
\end{figure*}

\begin{figure*}[hb!]
\hspace{-2.5cm}{
	\subfigure[WGANGP CIFAR10]{\includegraphics[width=4cm]{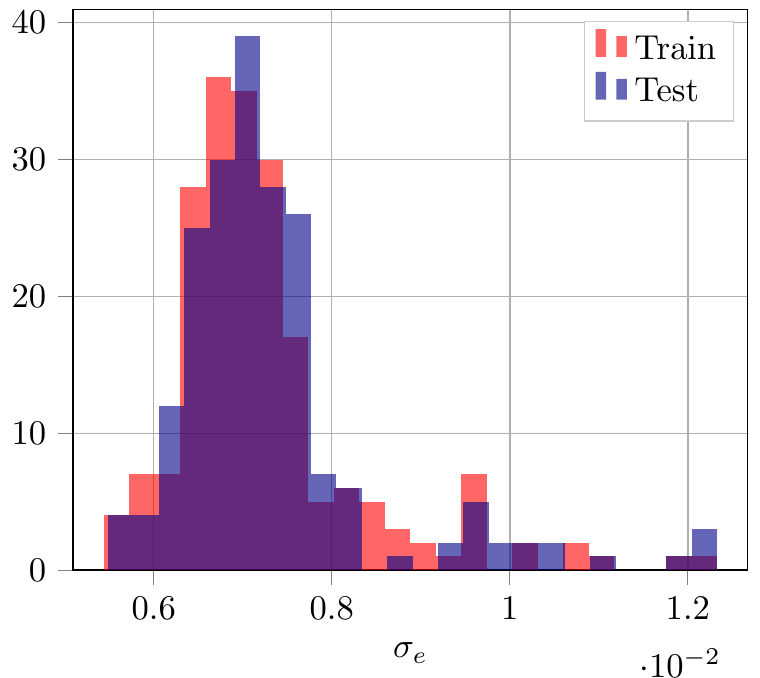}}%
	\subfigure[WGAN CIFAR10]{\includegraphics[width=4cm]{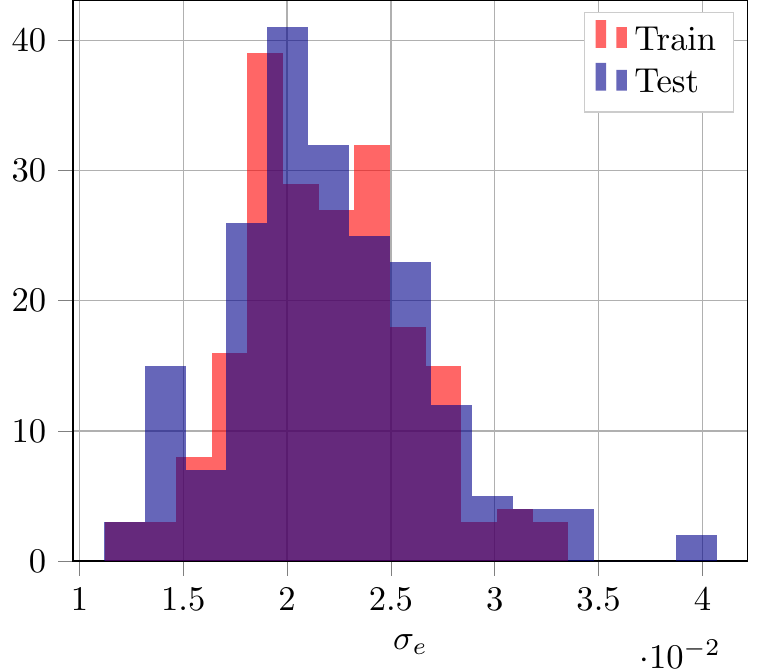}}%
	\subfigure[SNDCGAN CIFAR10]{\includegraphics[width=4cm]{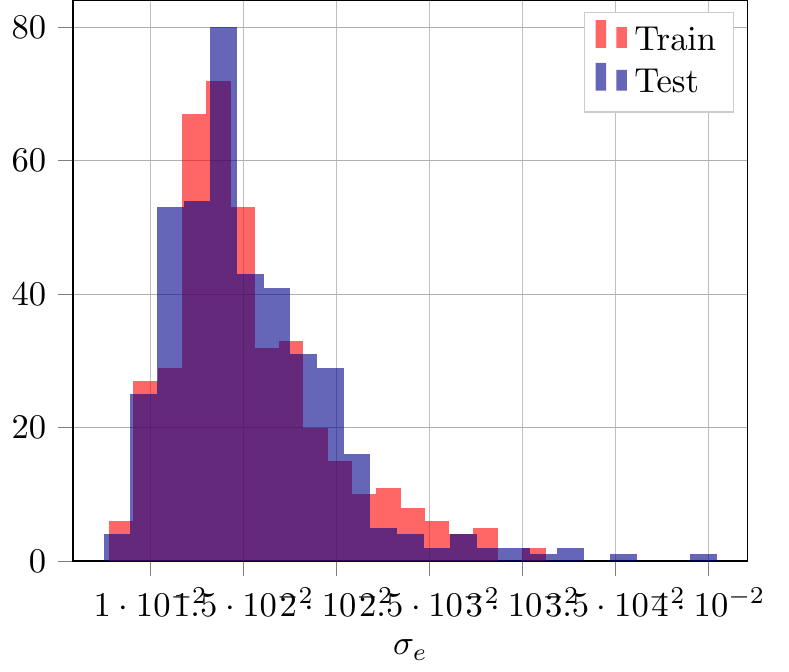}}%
	\subfigure[WGAN-GP celebA]{\includegraphics[width=4cm]{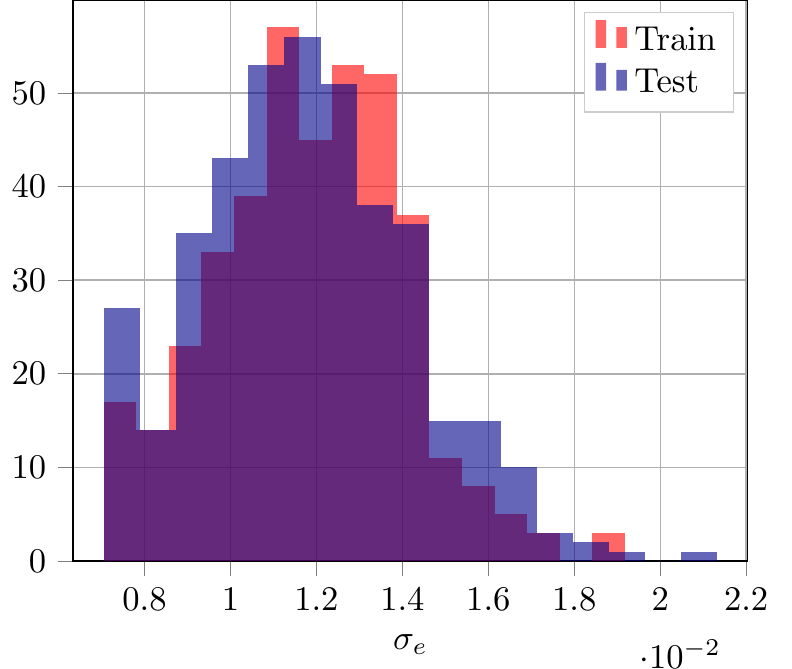}}%
	\subfigure[SNDCGAN celebA]{\includegraphics[width=4cm]{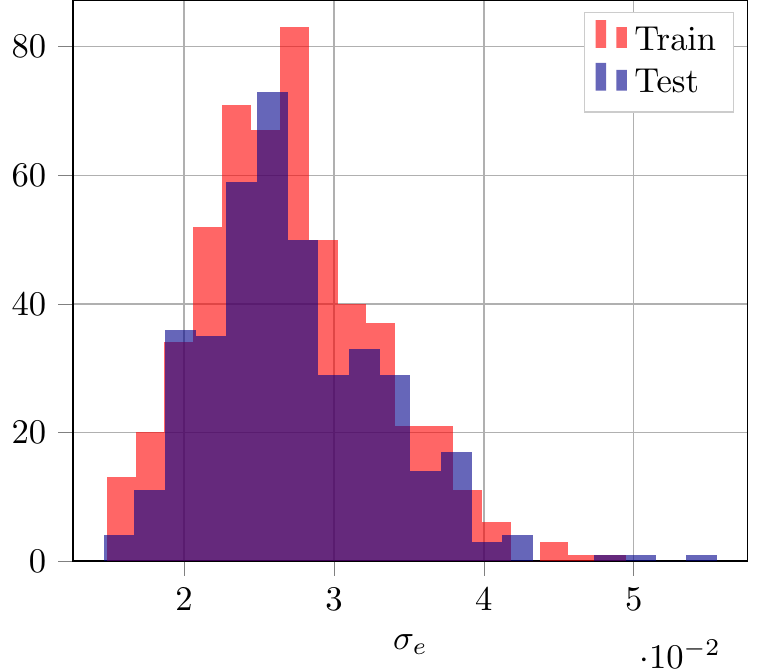}}%
}
	\\
	
	\hspace{-2.5cm}{
	\subfigure[WGANGP CIFAR10]{\includegraphics[width=4cm]{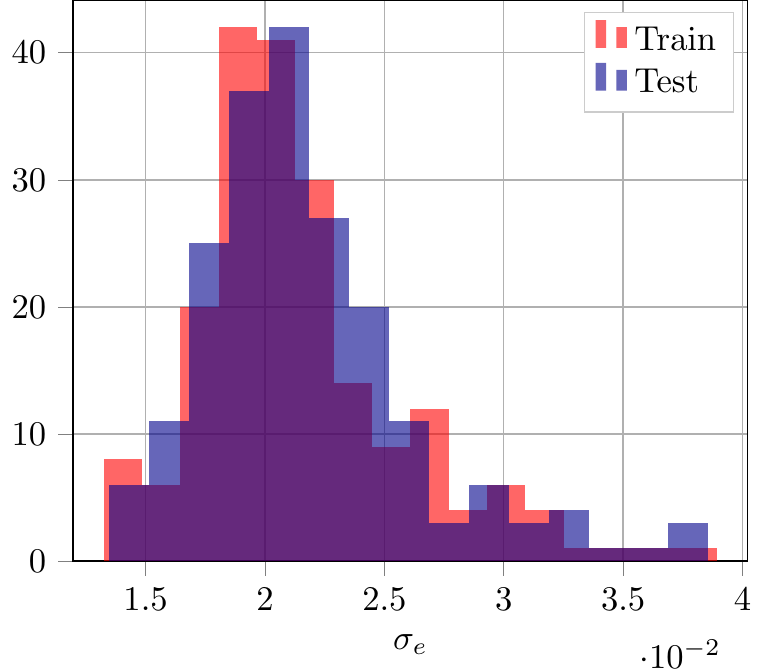}}%
	\subfigure[WGAN CIFAR10]{\includegraphics[width=4cm]{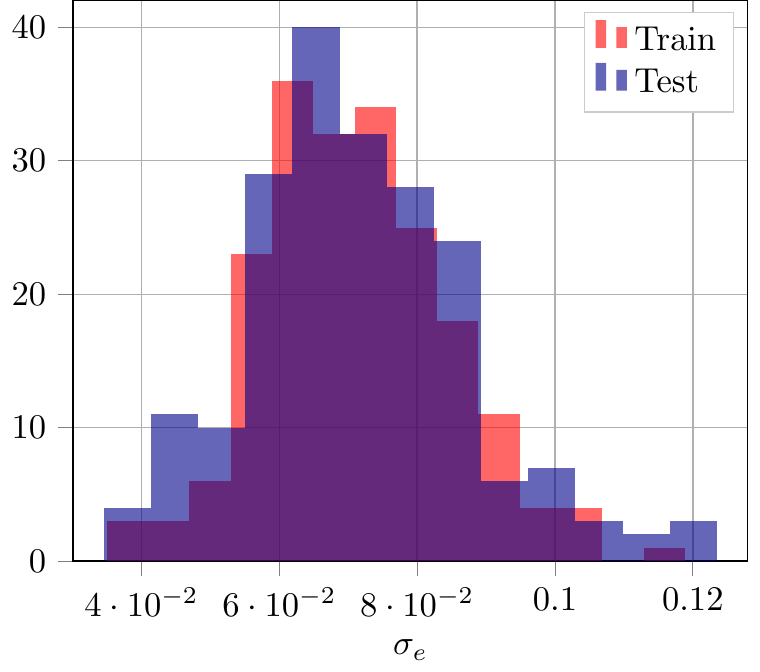}}%
	\subfigure[SNDCGAN CIFAR10]{\includegraphics[width=4cm]{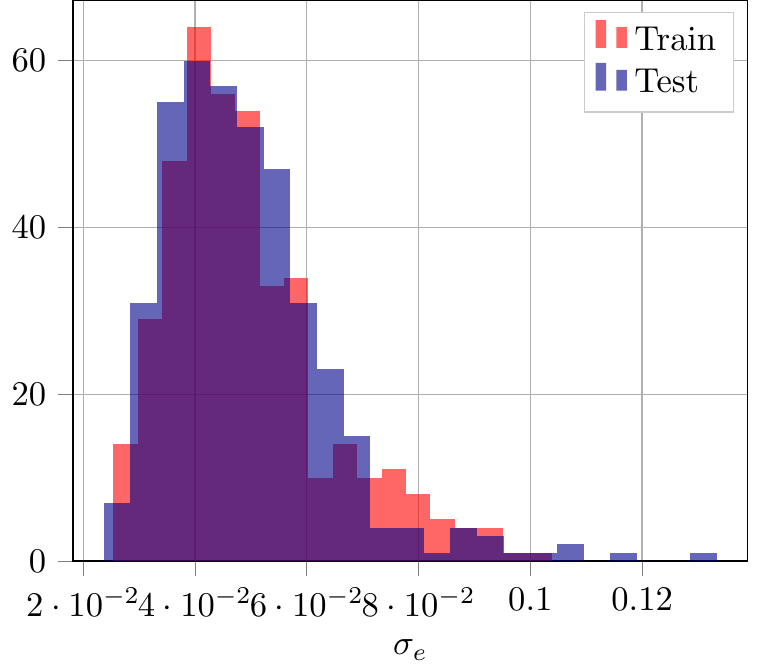}}%
	\subfigure[WGAN-GP celebA]{\includegraphics[width=4cm]{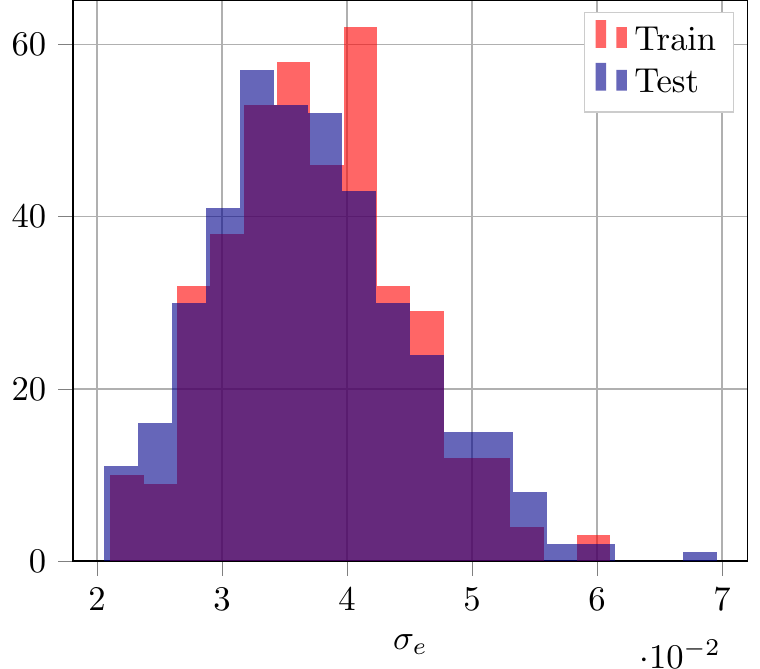}}%
	\subfigure[SNDCGAN celebA]{\includegraphics[width=4cm]{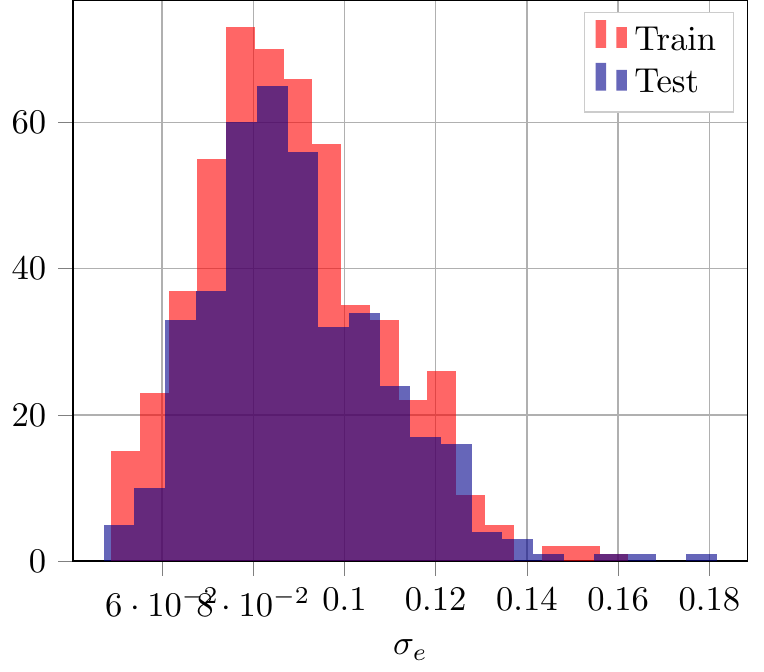}}%
}
	\caption{In the top row we show the histogram of the maximum value of $\sigma_\epsilon$ for which the average PSNR w.r.t. to $G(\z_c^*)$ is less than 40 dB. In the bottom row, we reproduce this experiment for a maximum  PSNR value of 30 dB.}%
	\label{fig:hist_PSNR}
\end{figure*}

\begin{figure*}[hb!]
\hspace{-2.5cm}{
	\subfigure[WGANGP CIFAR10]{\includegraphics[width=4cm]{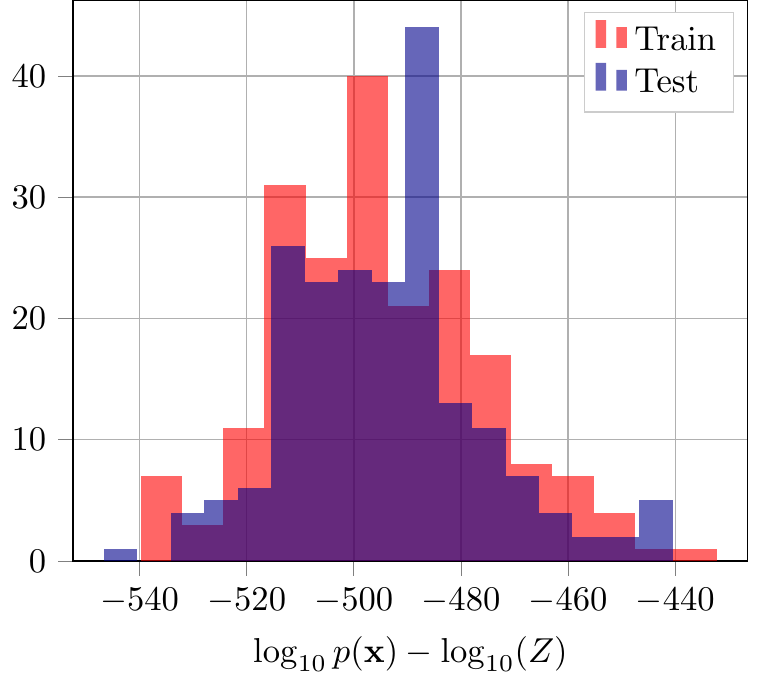}}%
	\subfigure[WGAN CIFAR10]{\includegraphics[width=4cm]{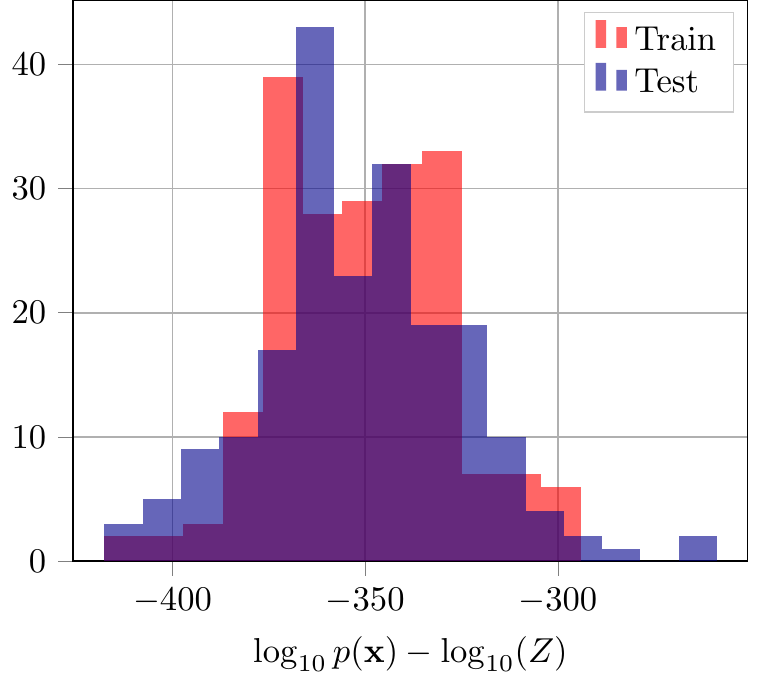}}%
	\subfigure[SNDCGAN CIFAR10]{\includegraphics[width=4cm]{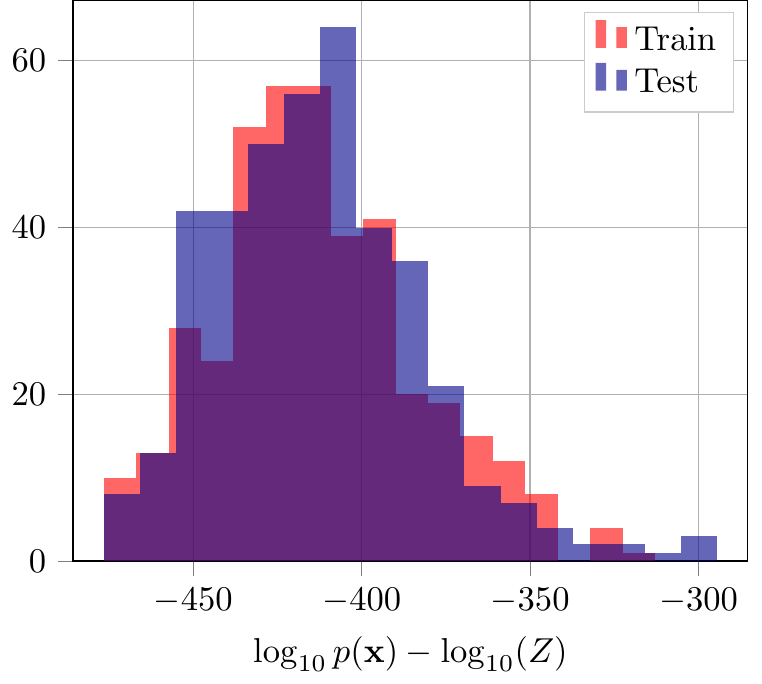}}%
	\subfigure[WGAN-GP celebA]{\includegraphics[width=4cm]{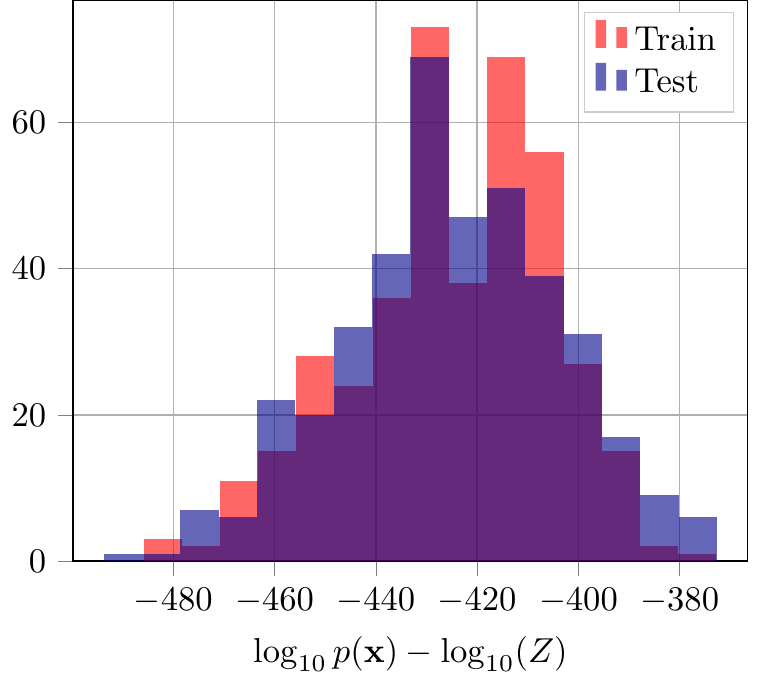}}%
	\subfigure[SNDCGAN celebA]{\includegraphics[width=4cm]{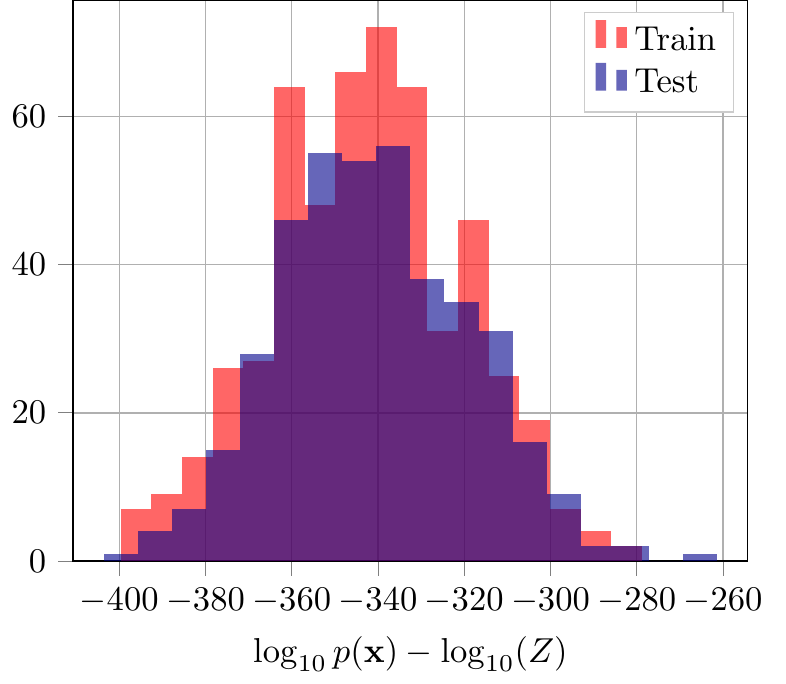}}%
	\caption{Histogram of the EvalGAN estimated log-probability (unnormalized) using 400 test_train images.}%
}
	\label{fig:non_iso_LogP2}
\end{figure*}

\begin{figure*}[h!]
	\centering
	\subfigure[WGANGP CIFAR10]{\includegraphics[width=7.5cm ]{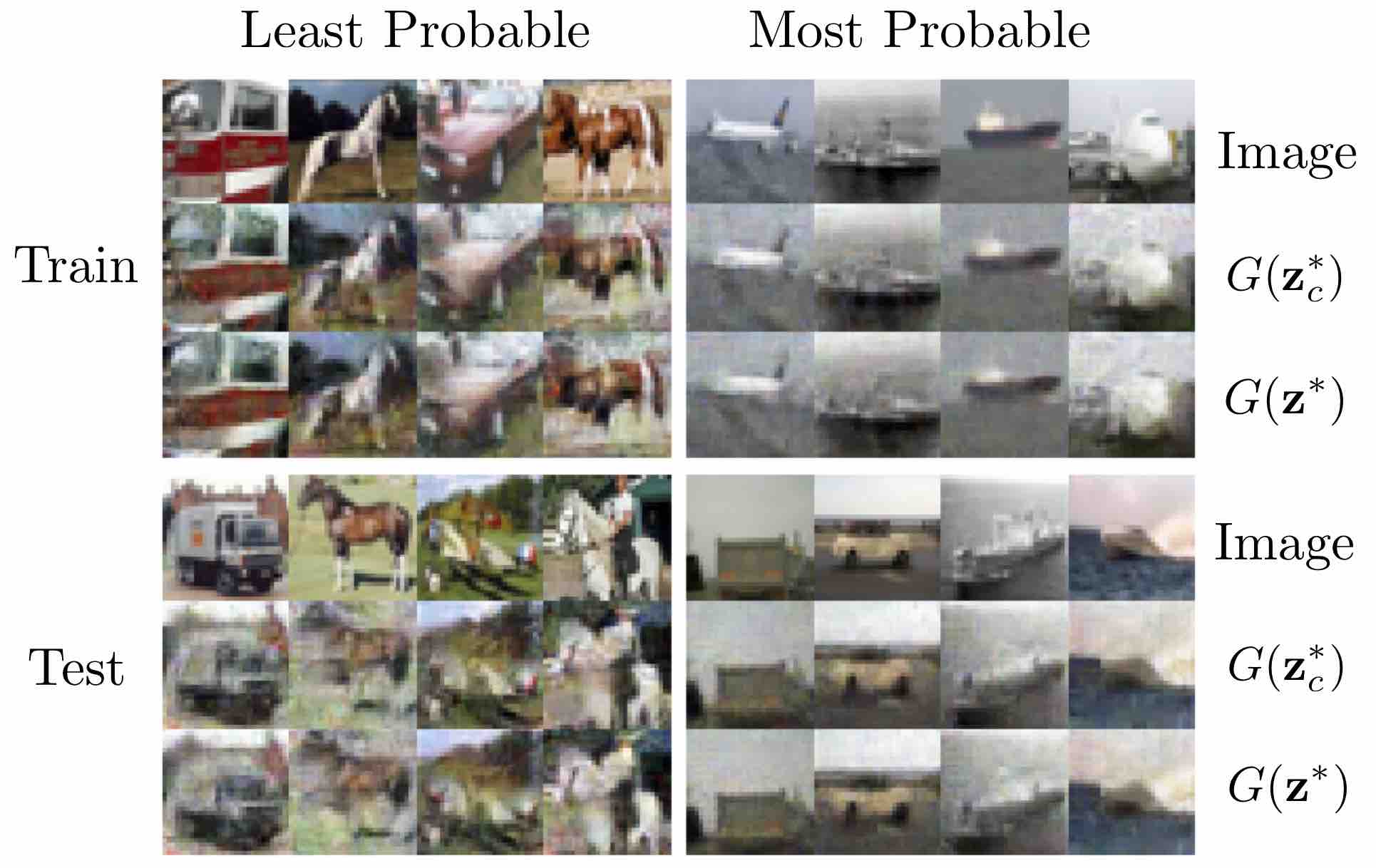}}\\%
	\subfigure[WGAN CIFAR10]{\includegraphics[width=7.5cm ]{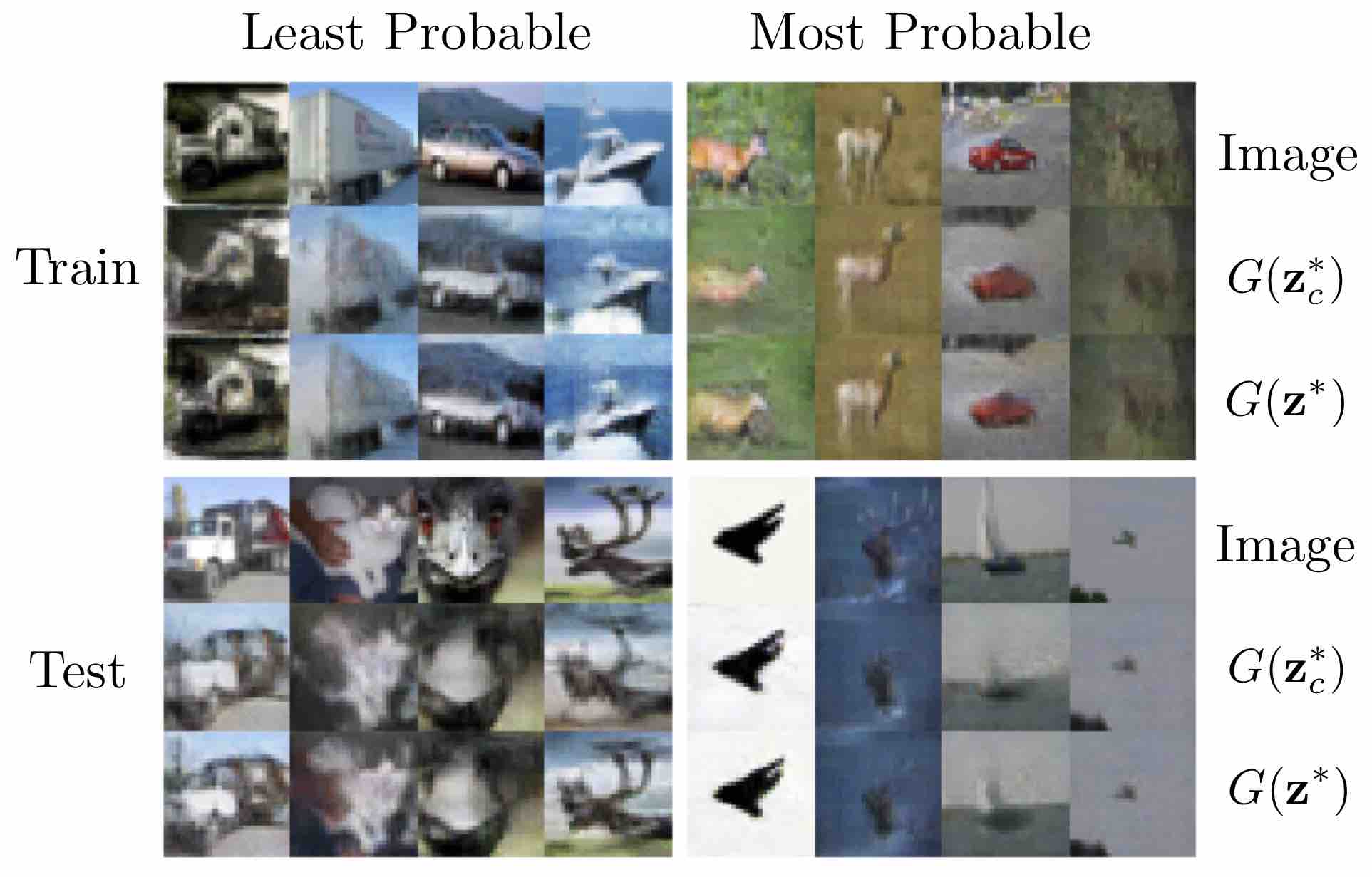}}%
	\subfigure[SNDCGAN CIFAR10]{~~~~~~\includegraphics[width=7.5cm ]{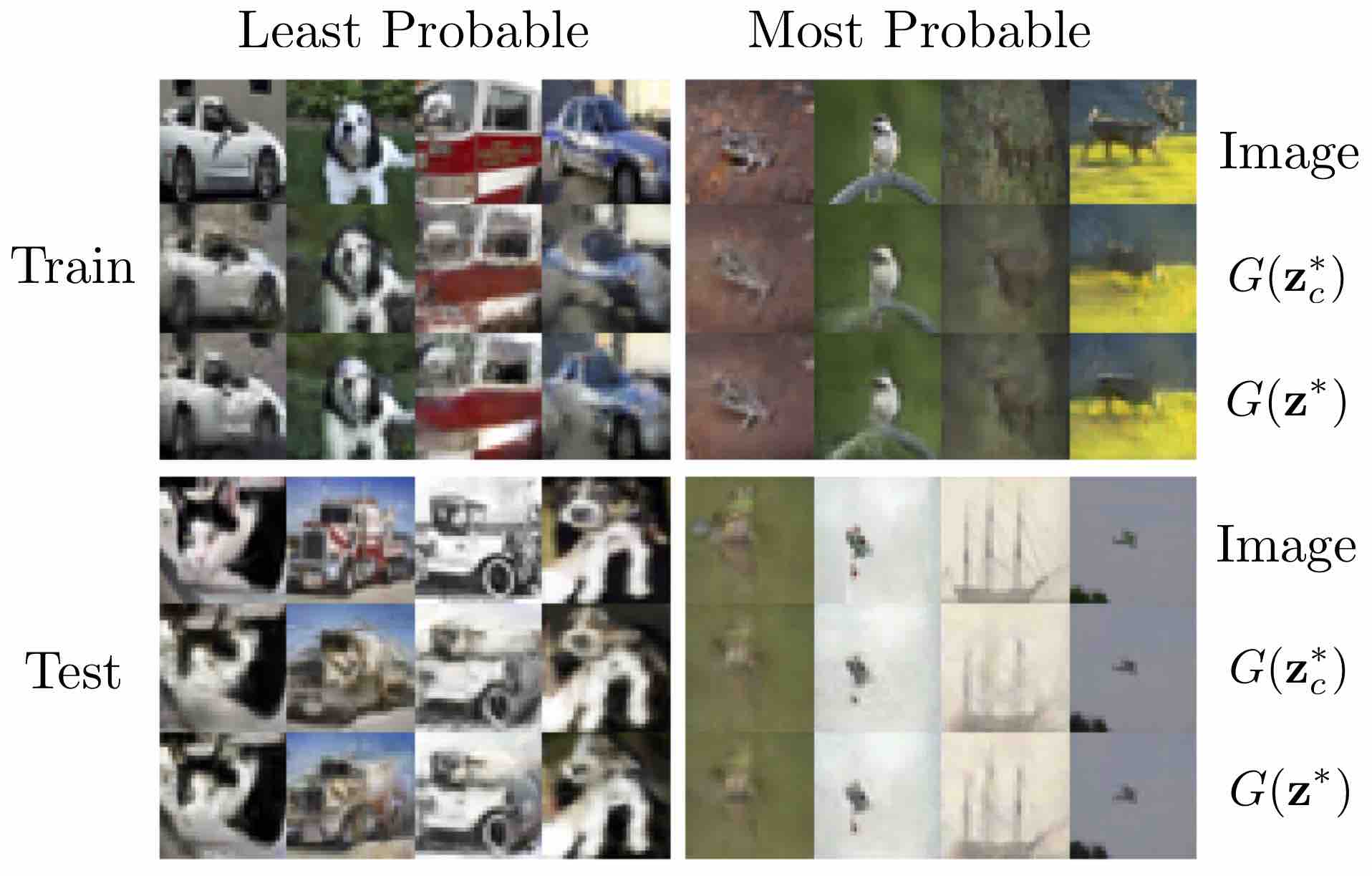}}%
	\\
	\subfigure[WGAN-GP celebA]{\includegraphics[width=7.5cm ]{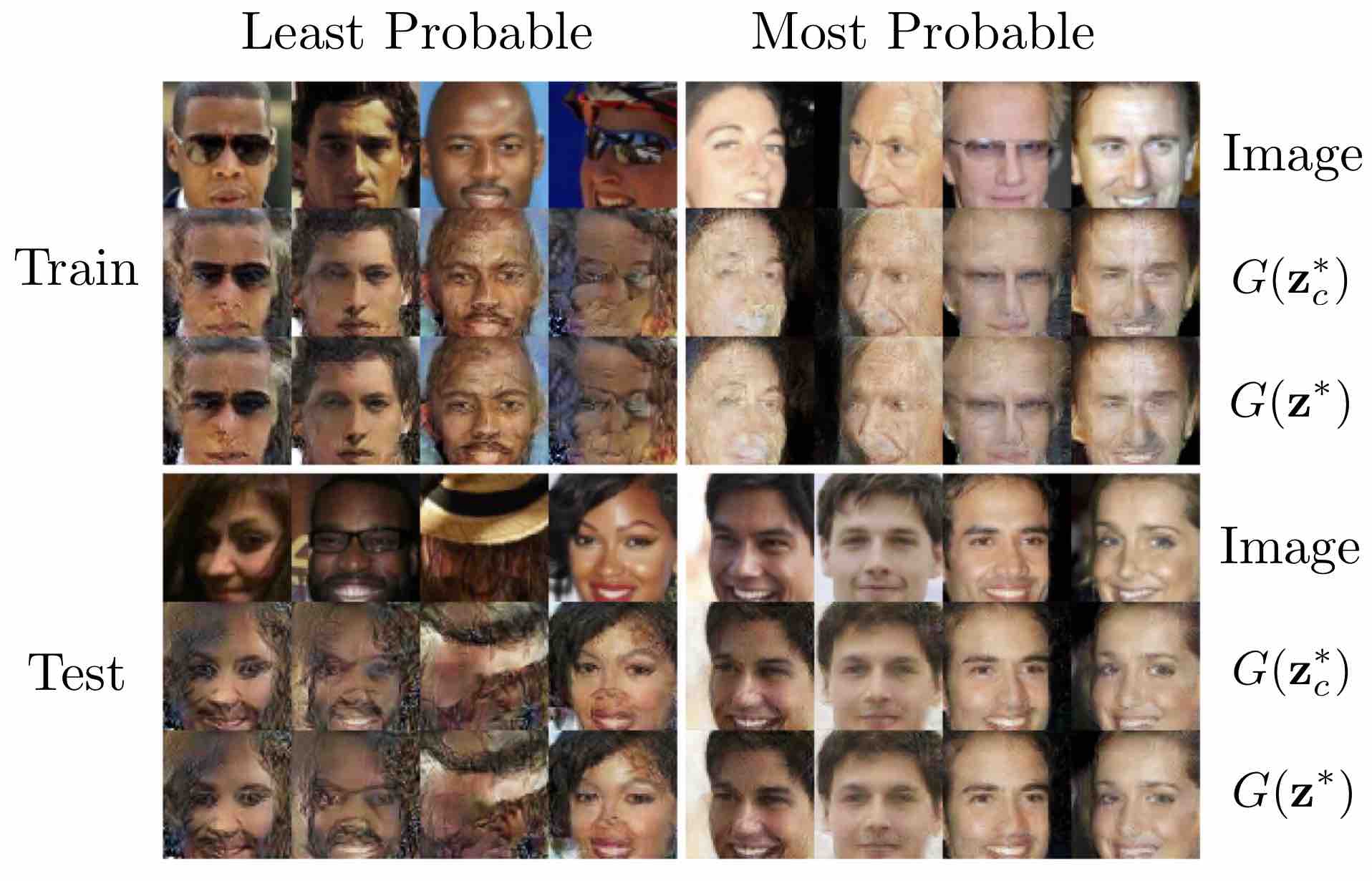}}%
	\subfigure[SNDCGAN celebA]{~~~~~~~~\includegraphics[width=7.5cm ]{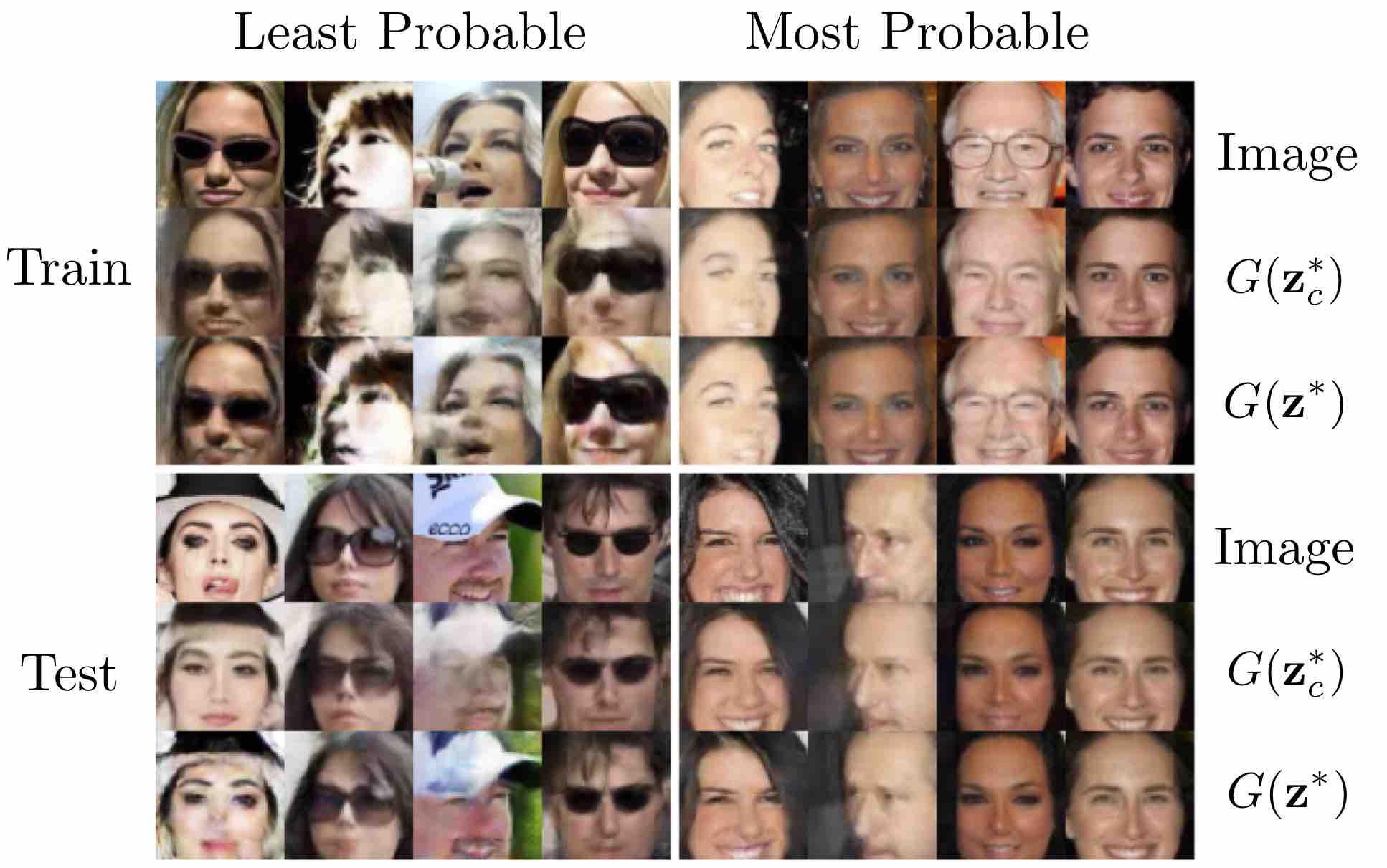}}%
	\caption{We plot the most and least probable images for each case according to the EvalGAN probability measure.}%
	\label{fig:non_iso_less_most_prob}
\end{figure*}

\begin{figure*}[h!]
	\centering
	\subfigure[WGANGP CIFAR10]{\includegraphics[width=7cm]{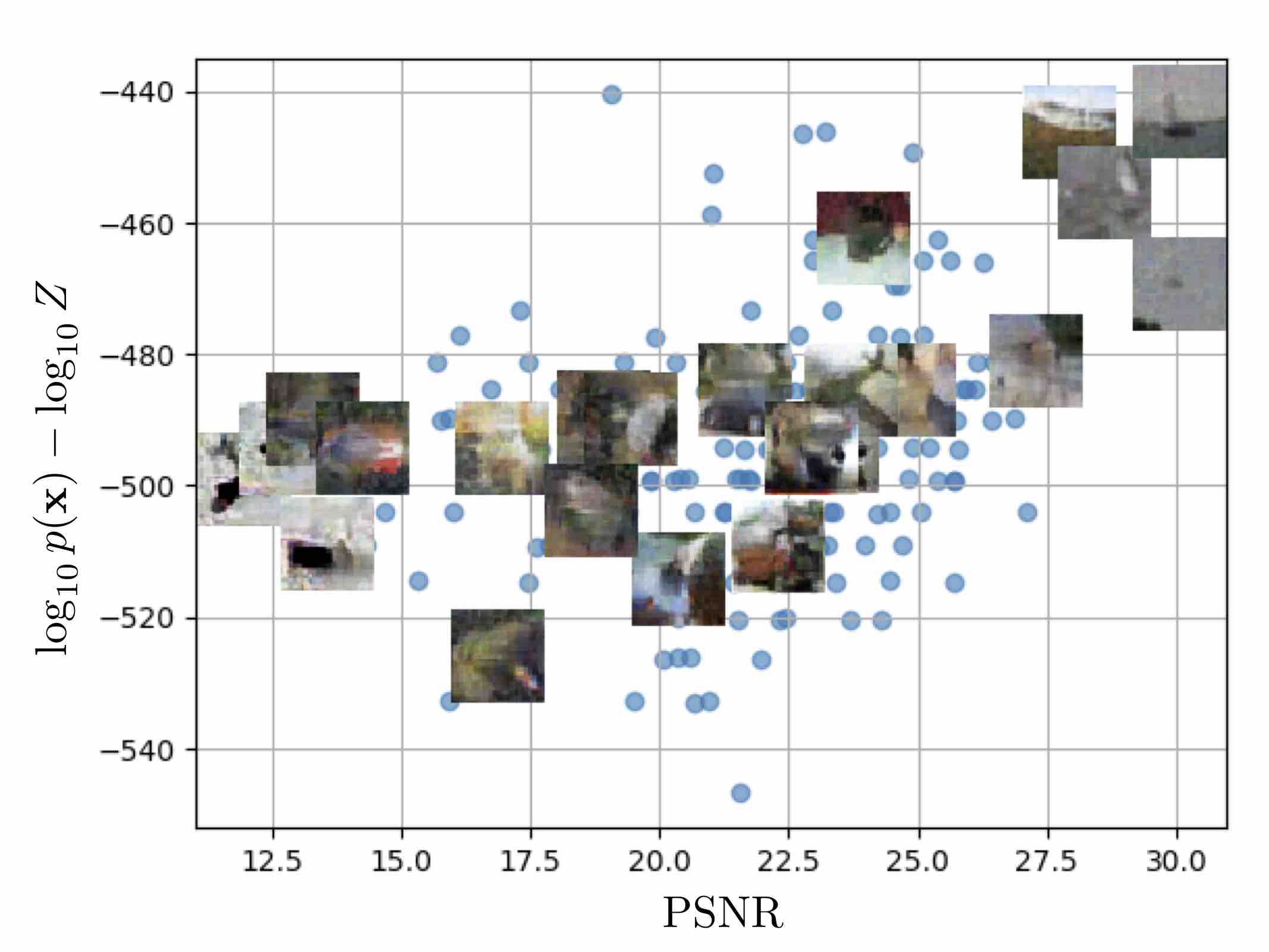}}%
	\\
	\subfigure[WGAN CIFAR10]{~~~\includegraphics[width=7cm]{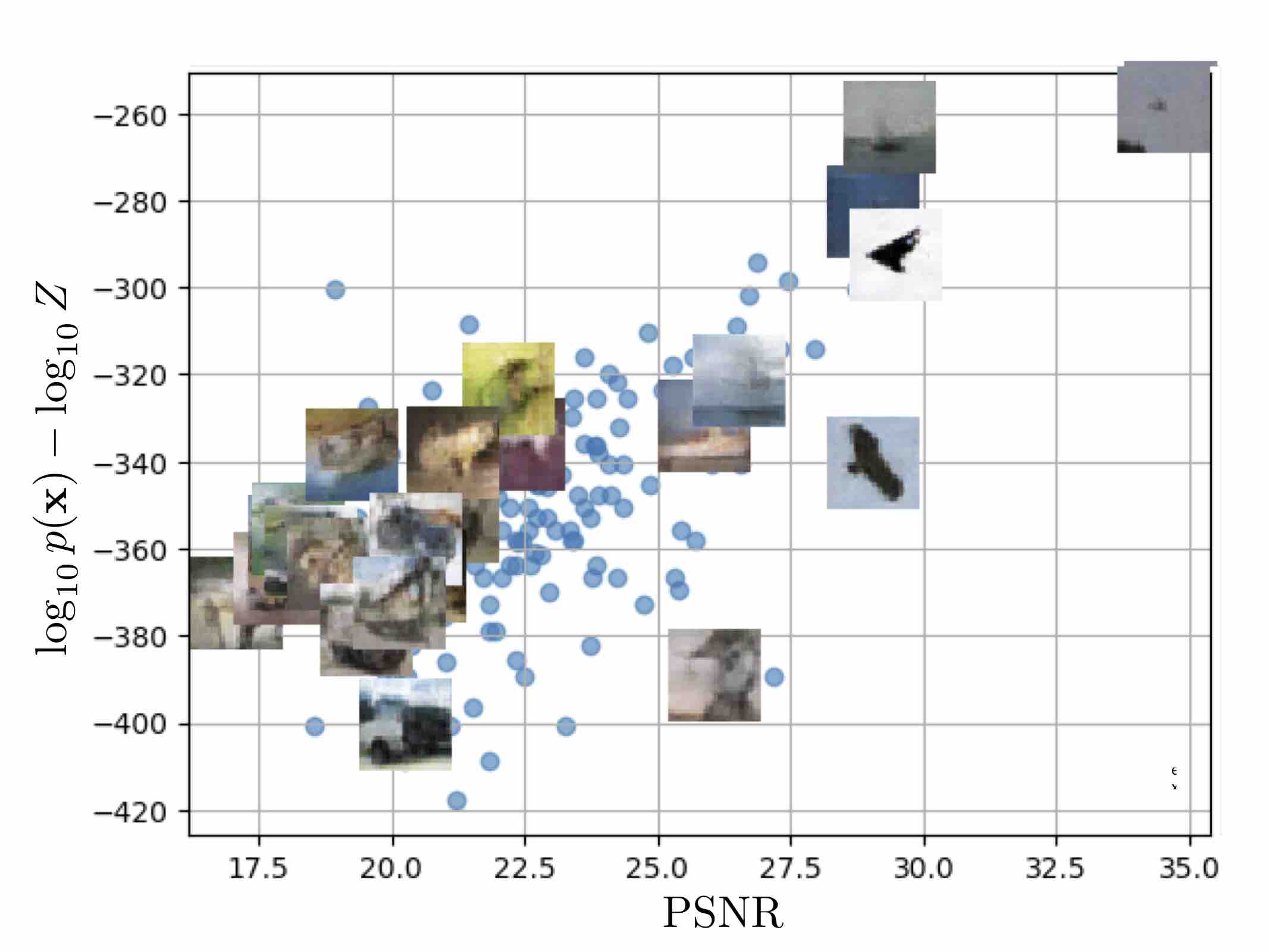}}%
	\subfigure[SNDCGAN CIFAR10]{\includegraphics[width=7cm]{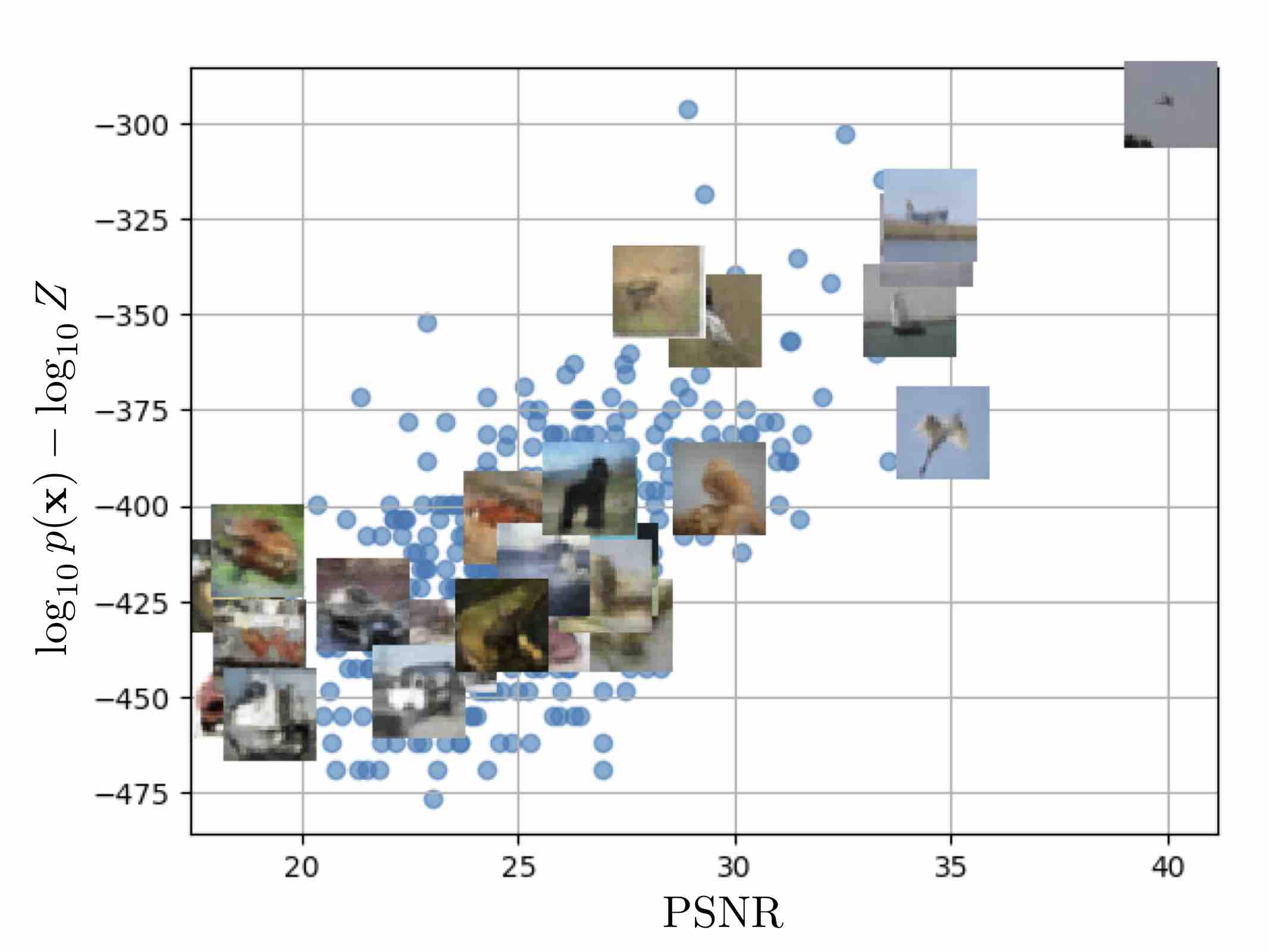}}%
	\\
	\subfigure[WGAN-GP celebA]{~~~\includegraphics[width=7cm]{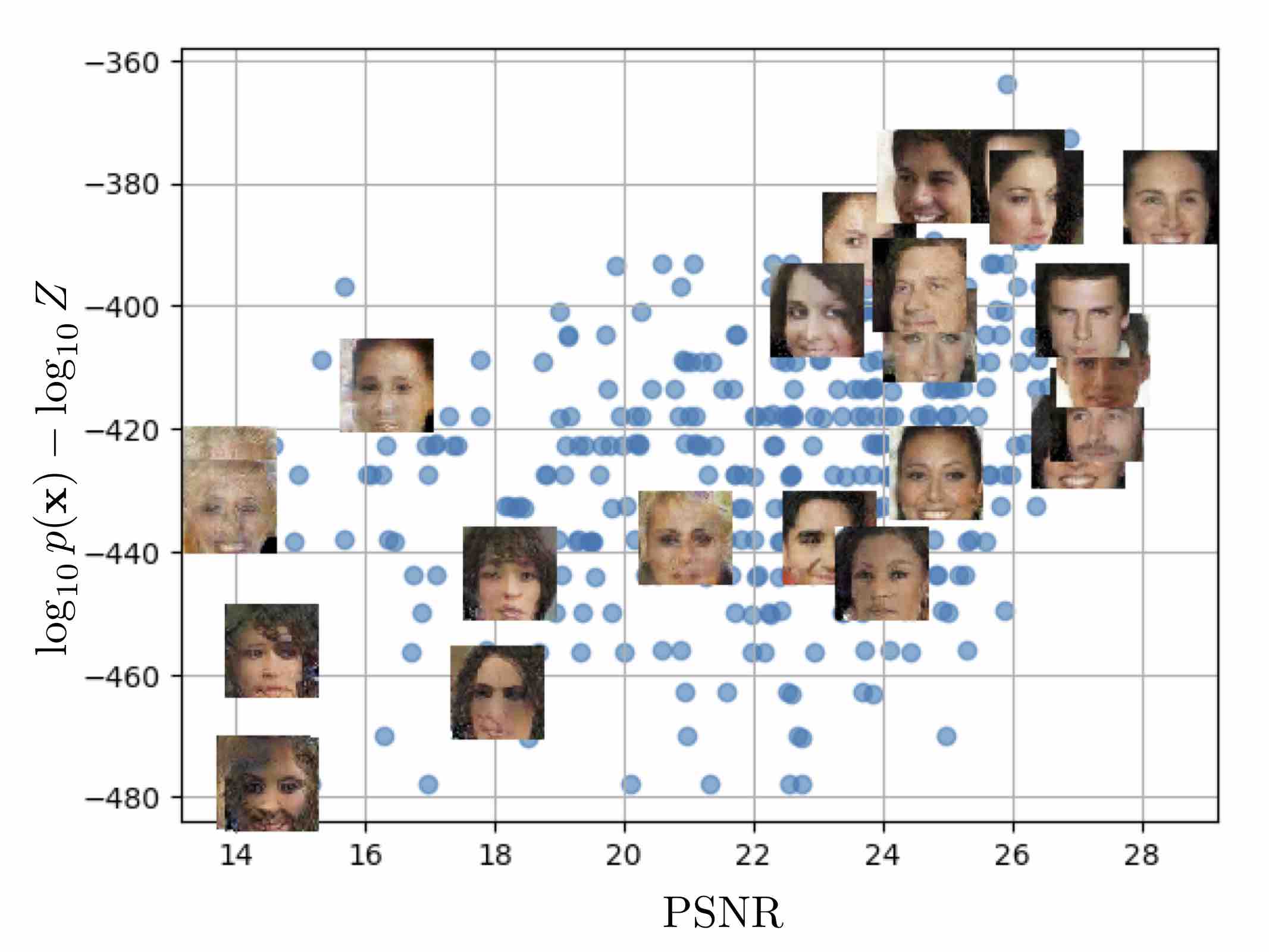}}%
	\subfigure[SNDCGAN celebA]{\includegraphics[width=7cm]{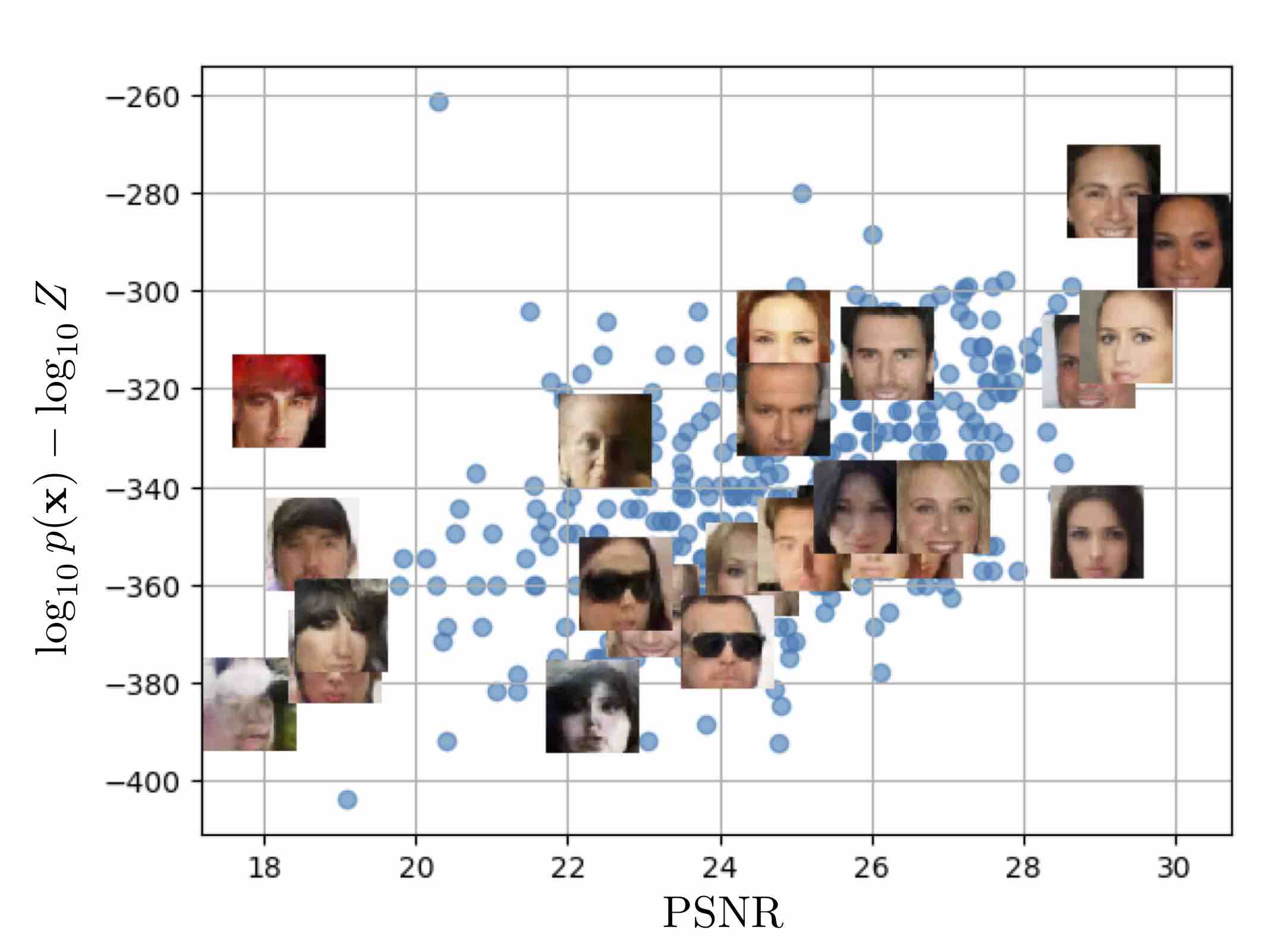}}
	\caption{Scatter plot of PSNR versus estimated loglikelihood obtained with EvalGAN. We also show some reconstructed images $G(\z^*_c)$ overlaying their corresponding location in the plot.}%
	\label{fig:scatter}
\end{figure*}

The proposed metric to estimate the marginal likelihood of generating a given sample $p(\x_{\text{test}})\propto \frac{N_c^*}{N}\bar{\sigma}_\epsilon^{\dim(\z)}$ is based on evaluating the distortion between the generator output with inputs $\z_c^*$ and $\z_c^*$ corrupted by additive Gaussian noise of a certain variance $\sigma^2_\epsilon$.  For 20 test and train images, in Figure \ref{fig:z_region_PSNR_norm_some} we plot the evolution of the average PSNR between $G(\z_c^*)$ and $G(\z_c^*+\epsilon_i)$  as $\sigma_\epsilon$ grows. In all cases $\text{dim}(\z)=256$. Observe that  there exists a significant variability in the degradation that each image suffers as samples are further apart from $\z_c^*$. This is better illustrated in Figure \ref{fig:hist_PSNR}, where in the top row we show the histogram of the maximum value of $\sigma_\epsilon$ for which the average PSNR w.r.t. to $G(\z_c^*)$ is less than 40 dB. In the bottom row, we reproduce this experiment for a maximum  PSNR value of 30 dB. In all cases we have used 400 test/train images. For the same set of images, in Figure \ref{fig:non_iso_LogP2} we show the unnormalized log marginal likelihood histogram using the. In all cases, results indicate an extreme overrepresentation of some samples in the test set, which corresponds to simple images with smooth textures and uniform backgrounds in CIFAR10 and plain smiling faces in SN-DCGAN, as it can be observed in Figure \ref{fig:non_iso_less_most_prob}. It is interesting to note that, particularly for SN-DCGAN with CelebA, reconstructed images using the solution $\z_c^*$ to the constrained problem tend to simplify the original image including features common in the set of most probable images, e.g. inserting soft smiles instead of more complicated gestures, or even removing the glasses.  This effect is less severe when we visualize the reconstructed image from the solution $\z^*$ to the unconstraint problem.
Figure \ref{fig:scatter} shows scatter plots comparing the PSNR w.r.t. the original image versus the estimated log marginal likelihood obtained using EvalGAN. Observe that simpler images tend to be in regions with higher marginal likelihoods and better reconstructions, according to PSNR. We believe this effect must be certainly introducing a bias during the training of the GANs, as we sample minibatches of images from the generator at every training step.

\begin{figure*}[h!]
	\centering
	\subfigure[CIFAR10]{\includegraphics[width=8cm]{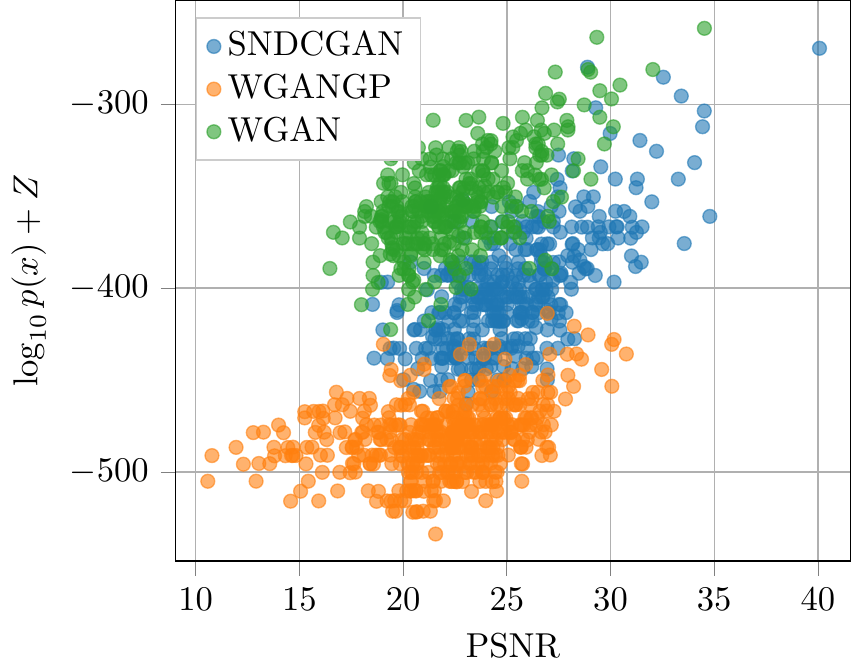}}%
	\subfigure[CelebA]{\includegraphics[width=8cm]{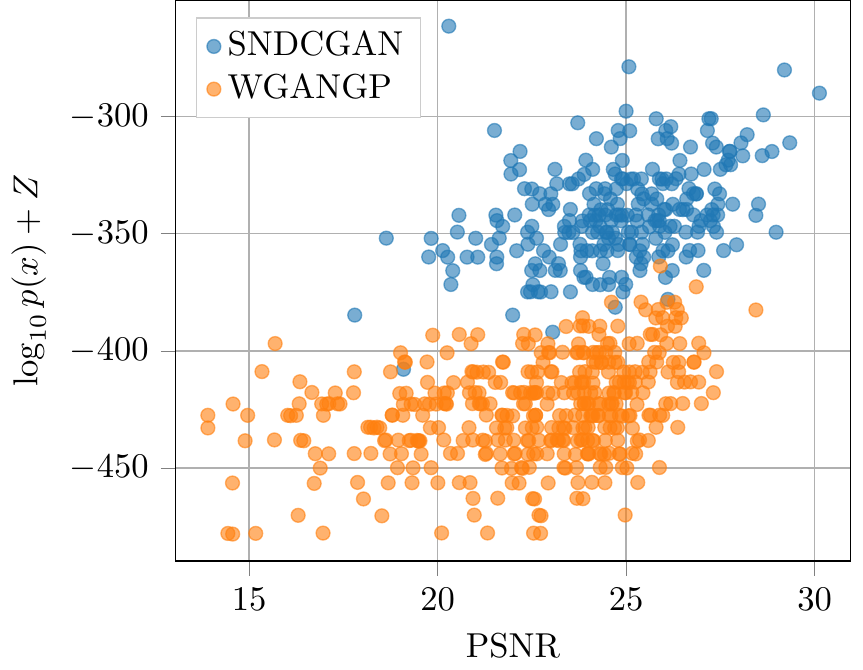}}%
	\caption{Comparison of different models using EvalGAN for CIFAR10 in (a) and CelebA in (b)}%
	\label{fig:scatter_comparison}
\end{figure*}

In Figure \ref{fig:scatter_comparison} we show a comparison of different GANs using EvalGAN. SN-DCGAN performs better than WGAN-GP both in terms of reconstruction capabilities and in sample marginal likelihood. Also,  WGAN on CIFAR10 provides much higher marginal likelihoods than SN-DCGAN, at the cost of worse average PSRN reconstruction quality.

\end{document}